\documentclass[journal]{IEEEtran}
\usepackage{cite}

\ifCLASSINFOpdf
\else
\fi
\usepackage{graphicx}
\usepackage{epstopdf}
\usepackage{enumerate}
\usepackage{hyperref}
\usepackage{color}
\usepackage{soul}
\usepackage{graphicx}
\usepackage{caption}
\usepackage{subcaption}
\usepackage[fleqn]{amsmath}
\usepackage{amssymb}
\usepackage{algorithm,algorithmic}
\usepackage{paralist}
\usepackage{multirow}
\DeclareMathOperator*{\argmax}{arg\,max}
\usepackage[inline]{enumitem}
\newcommand\siz{0.7in}
\newcommand\sizz{0.65in}
\newcommand\sizp{0.7in}

\begin{document}
%
\title{Curvature Integration in a 5D Kernel for Extracting Vessel Connections in Retinal Images}
%
%
%

\author{Samaneh~Abbasi-Sureshjani,~Marta~Favali,\\~Giovanna~Citti,~Alessandro~Sarti,~Bart~M.~ter~Haar~Romeny,~\IEEEmembership{Senior Member,~IEEE}
\thanks{This project has received funding from the European Union's Seventh Framework Programme, Marie Curie Actions-Initial Training Network, under grant agreement $n^o 607643$, ``Metric Analysis For Emergent Technologies (MAnET)". It was also supported by the H\'e Programme of Innovation, which is partly financed by the Netherlands Organization for Scientific research (NWO) under grant $n^o 629.001.003$.}
\thanks{S. Abbasi-Sureshjani  and B. M. ter Haar Romeny are with the Department of Biomedical Engineering, Eindhoven University of Technology, Eindhoven 5600MB, the Netherlands (e-mail: s.abbasi@tue.nl).}
\thanks{M. Favali and A. Sarti  are with Centre d'Analyse et de Math\'ematique Sociales, \'Ecole des Hautes \'Etudes en Sciences Sociales, 190-198, Avenue de France 75244 Paris Cedex 13  France (e-mail:\{marta.favali, alessandro.sarti\}@ehess.fr).}
\thanks{G. Citti is with the Department of Matematics, University of Bologna, Piazza di Porta S.Donato 5, Bologna, Italy (e-mail: giovanna.citti@unibo.it).}
\thanks{B. M. ter Haar Romeny is with the Department of Biomedical and Information Engineering, Northeastern University, 500 Zhihui Street, Shenyang, 110167 China (e-mail: bart.romeny@outlook.com).}
\thanks{S. Abbasi-Sureshjani and M. Favali contributed equally to this work.}
\thanks{Manuscript received March 3, 2017; revised ***.}
}

%
%

\markboth{Journal of \LaTeX\ Class Files,~Vol.~*, No.~*, March~2017}%
{Shell \MakeLowercase{\textit{et al.}}: Bare Demo of IEEEtran.cls for IEEE Journals}
%



\maketitle

\begin{abstract}
Tree-like structures such as retinal images are widely studied in computer-aided diagnosis systems for large-scale screening programs. Despite several segmentation and tracking methods proposed in the literature, there still exist several limitations specifically when two or more curvilinear structures cross or bifurcate, or in the presence of interrupted lines or highly curved blood vessels. 
In this paper, we propose a novel approach based on multi-orientation scores augmented with a contextual affinity matrix, which both are inspired by the geometry of the primary visual cortex (V1) and their contextual connections. The connectivity is described with a five-dimensional kernel obtained as the fundamental solution of the Fokker-Planck equation modelling the cortical connectivity in the lifted space of positions, orientations, curvatures and intensity. It is further used in a self-tuning spectral clustering step to identify the main perceptual units in the stimuli. The proposed method has been validated on several easy and challenging structures in a set of artificial images and actual retinal patches. Supported by quantitative and qualitative results, the method is capable of overcoming the limitations of current state-of-the-art techniques.

\end{abstract}

\begin{IEEEkeywords}
Contextual affinity matrix, Curvature, Perceptual grouping, Primary visual cortex, Retinal image analysis, Spectral clustering.
\end{IEEEkeywords}

\section{Introduction}
\label{sec:intro}
\sloppy Diabetes, as many other systematic, cardiovascular and ophthalmologic diseases, is widespread worldwide, especially in developing countries. Diabetic retinopathy is the progressive damage to the network of tiny blood vessels in the retina and it is due to the diabetes. It is the main cause of blindness and affecting the quality of life of many people in addition to raising healthcare and social costs substantially~\cite{viswanath2003diabetic,hu2011globalization}. Early diagnosis and treatment is essential to stop diseases progression and reduce the financial and emotional costs. Moreover, developing automated computer-aided systems facilitate the diagnostic process for a larger population of people in a shorter time and at lower costs.  

Based on several studies, the retinal vasculature is one of the easiest-to-access and highly informative sources of diagnostic information not only for diabetic retinopathy, but also glaucoma, hypertensive retinopathy, and other diseases, considering them being part of the brains vasculature \cite{hubbard1999methods,foracchia2001extraction}. 
Geometrical features such as change of vessel width, curvature, branching patterns and fractal dimension are all considered as biomarkers for clinical studies~\cite{Fan2015FD, Bekkers2015, chapman2002peripheral,habib2014association}. 
Quantitative measurement of these features is highly dependent on correct detection and analysis of the morphologic and geometric structure of the retinal vasculature i.e. blood vessels, bifurcations and their connections. 

Detection of blood vessels is often done via vessel segmentation, which is a well-studied topic in retinal image analysis. The segmentation methods basically differentiate between blood vessels and background pixels, but they do not separate individual vessels from each other. Therefore, they have been used often as an initial step in tracking approaches (e.g. in~\cite{xu2011vessel,de2014tracing,de2013automated,de2016graph,turetken2011automated,turetken2012automated,dashtbozorg2014automatic}). Tracking the vessels makes it possible to investigate the features along individual vessels separately. 
The vasculature network constructed during tracking not only needs to be a correct global model, but it also should model the local connections between individual vessel segments at crossings and bifurcations correctly.

Despite all the methods proposed in literature, tracking approaches are still facing some difficulties, which are either coming from imperfect segmentation and their corresponding skeletons, or they arise at crossovers and junction points, where two vessels belonging to separate trees meet or one vessel bifurcates. Any errors created during the segmentation or skeleton extraction steps propagate to the next levels and if the method is not able to resolve these issues, it results in having wrong paths. 
There have been several efforts in the literature for tackling these difficulties from different points of views. In addition to great performance improvements in state-of-the-art segmentation techniques (e.g.~\cite{azzopardi2015trainable}), the authors in \cite{de2013automated,de2016graph,turetken2011automated,turetken2012automated} have used tubularity measurement techniques as the initial step of their method. Moreover, to solve the centreline extraction problems, which is often done via morphological thinning, several modifications and rule-based post-processing methods have been proposed \cite{dashtbozorg2014automatic,de2016graph}. On the other hand, in order to find the right connections at junction points, the geometrical configuration of the junctions has been used in local and global costs followed by optimization procedures~\cite{hu2015automated,turetken2011automated,turetken2012automated}. In all these works, cost functions were designed in a way to avoid abrupt changes of geometrical properties (most importantly orientation) at junction points and allow a smooth transition from one vessel to the other. 
The method proposed by \cite{de2016graph} is a digraph-based label propagation technique using the matrix-forest theorem. The digraph weight matrix is designed in such a way that it does not allow the connected filaments (vertices) to bend too much. In the methods proposed by~\cite{de2013automated,cheng2014tracing} these constraints on the geometrical properties at junction points were used to find the right connections. 
A different research line, aimed to extract geometrical features such as orientation or curvature from natural images, is inspired by the structure of the visual cortex. 
The works by \cite{koenderink1987representation,hoffman1989visual,hoffman1966lie} introduced the mathematical modelling of its functional architecture in terms of differential geometry. The study of \cite{field1993contour} described the problem of context perception with psychophysical experiments, introducing the notion of association fields as the information integration along elements of images that satisfy the Gestalt law of good continuation~\cite{wagemans2012century,wertheimer1938laws}. Models which take into account the structure of the cortex were proposed by \cite{mumford1994elastica,williams1997stochastic,august2000curve}, where the association fields were described with the Fokker-Planck equations. In these models, different features as orientation and curvature were considered. 
Then \cite{citti2006cortical} proposed a model of the association fields based on the
functional organization of the primary visual cortex (V1), establishing a relation between neural mechanism and image completion further developed by \cite{sanguinetti2008image}. A similar approach was independently developed by \cite{duits2007image,duits2007invertible} who developed a Lie group theory for image analysis. This cortical method was later adapted to the problem of perceptual unit identification in \cite{sarti2015constitution}, based on a previous study on neural synchronization \cite{sarti2003neural}. In our previous work \cite{favali2016analysis}, we applied an instrument inspired by the geometry of the primary visual cortex (V1) and the approach of \cite{sarti2015constitution} to group and separate the blood vessels as individual perceptual units, even though there were  informations missing due to poor segmentations. This method could find the right connections between small vessels and their parents, which are usually missing in the literature. However, the bottleneck of this method is that it is not data-adaptive and it does not follow some of the highly curved vessels.

%

In this work we further develop our previous model in order to solve the limitations found during the tracking of blood vessels, adding several novelties. We develop an special geometrical setting, introducing new theoretical instruments of differential geometry of vector fields. Particularly we work in a 5D space of positions, orientations, intensity and curvatures, and develop a sub-Riemannian structure in this setting.  We introduce the feature of curvature as an important biomarker in retinal image analysis. We will see that detecting the feature of curvature in very noisy retinal images presents technical difficulties,  and we will need to extend the previous highly robust single-scale detection technique developed by \cite{Bekkers2015} to a multi-scale approach considering that the blood vessels in retinal images have varying widths. Moreover, we apply a self-tuning spectral clustering technique proposed by \cite{zelnik2004self} that allows to standardize the method to automatically detect the main groups in the data without any further parameter tuning. Summarizing, the main contributions of this article are:
\begin{enumerate}[label=\alph*)]
\item presenting a novel standardized and fully automatic application of the modeling of the visual cortex to the analysis of medical images;
\item  proposing a new sub-Riemannian structure in 5D space of position, curvature, orientation and intensity. This allows us to detect salient perceptual units in a self-tuning spectral clustering step, despite the presence of different types of variations including scale, orientation, curvature and intensity;
\item extending the curvature extraction technique to a multi-scale approach making it suitable for analysis of multi-scale blood vessels;
\item retrieving the challenging connections among bended, crossed, bifurcated or interrupted vessels in noisy, low quality or diseased retinal images;
\item demonstrating potential extensions of the method by applying it successfully on several curvilinear synthetic images with various complexities.
\end{enumerate}

The structure of the paper is the following. We will start in section \ref{sec:theory} with introducing the five-dimensional feature space, the new model of the cortical connectivity in this space, and the steps for lifting 2D retinal images. The qualitative and quantitative results of proposed experiments on two different sets are discussed in section \ref{sec:experiments}. Finally, the article is concluded in section \ref{sec:conclusion}.

\section{Theory and Methodology}
\label{sec:theory}

In this section, we present: a new algorithm based on a higher dimensional feature space, the model of cortical connectivity in this integrated 5D space and its application in the definition of an affinity matrix. Finally an automatic clustering algorithm, which takes as input our affinity matrix, and a new multi-scale curvature extraction technique for retinal images are presented.
Additionally, we explain the required preprocessing steps for preparing and lifting the retinal images in 5D space.

\subsection{Lifting the Image to the Orientation Cortical Space}A regular curve in the two-dimensional plane can be represented by $\gamma_{2D}(t) = (x(t),y(t))$. The tangent vector to the curve may be indicated as:
\begin{equation}
(\dot x(t), \dot y(t)) = (cos(\theta(t)),sin(\theta(t))).
\label{eq:tangent2D}
\end{equation}
Based on previous studies~\cite{citti2014neuromathematics,duits2007invertible}, the two-dimensional curves can be lifted to the ($\mathbb{R}^2\times S^1$) space of positions and orientations ($SE(2)$) using the direction of the tangent vector as:
\begin{equation}
\gamma_{2D} = (x(t),y(t)) \rightarrow
\gamma(t) = (x(t),y(t),\theta(t))
\end{equation}
where $\theta = arctan(\frac{\dot y}{\dot x})$. Therefore, we can write the following:
\begin{equation}
\dot\gamma(t) = (\dot x(t),\dot y(t),\dot \theta(t)) = \vec{X_1}(t) +\dot\theta(t) \vec{X_2}(t)
\label{eq:2Ddot}
\end{equation}
where $\vec{X_1} = (\cos(\vec\theta),\sin(\vec\theta),0)$, $\vec{X_2} = (0,0,1)$ are left-invariant vector fields with respect to the group law of $SE(2)$. The neural interpretation of this is that the lifted curves in the cortical space are the integral curves of the two vector fields $\{\vec{X_1},\vec{X_2}\}$. The curvature term $\dot\theta(t)=\kappa(t)$ indicates the rate of change of the orientation and determines the shape of the curve. The Lie algebra is generated by $\vec{X_1}$ and $\vec{X_2}$ coincides with $Span\{\vec{X_1},\vec{X_2},\vec{X_3}\}$, where $\vec{X_3} =-[\vec{X_1}, \vec{X_2}] = 
-\sin(\theta) \partial_x + \cos(\theta) \partial_y$. 


\subsection{Integrated Model of Position, Orientation, Curvature and Intensity}
\label{sec:corticalConn}
In this section, we describe a new model of the cortical connectivity in a 5D feature space of position, orientation, intensity and curvature. In our previous model~\cite{favali2016analysis} curvature was not present. We will see that this new feature will significantly change the structure of the space, since there is a strong interplay between orientation and curvature. In this way the model will allow strongly curved connectivity patterns. 
 
We assume that the two-dimensional curve in cortical plane ($\mathbb{R}^2$) is lifted to a 5D space of positions, orientations, intensity and curvature ($\mathbb{R}^2\times S^1\times \mathbb{R}^+\times \mathbb{R}$). Thus, the lifted curve may be written as:
\begin{equation}
\gamma_{2D} = (x(t),y(t)) \rightarrow
\gamma(t) = (x(t),y(t),\theta(t),f(t), \kappa(t)).
\end{equation}
Similar to (\ref{eq:2Ddot}), we will have:
\begin{equation}
\dot\gamma(t) = (\dot x(t),\dot y(t),\dot \theta(t),\dot f(t),  \dot\kappa(t)) 
\end{equation}
where $(\dot x(t),\dot y(t))= (\cos(\theta), \sin(\theta))$ 
and $\dot\theta(t) = \kappa(t)$. 
By defining new vectors in the 5D space as:
\begin{equation}
\begin{split}
\vec{Y}_1 = & (\cos(\vec\theta),\sin(\vec\theta),\kappa,0, 0) \\
\vec{Y}_5 =& (0,0,0,0,1)\qquad
\vec{Y}_4 = (0,0,0,1,0)\qquad 
\end{split}
\end{equation}
we are able to write $\dot\gamma(t)$ in terms of these vectors:
\begin{equation}
\dot\gamma(t) = \vec{Y}_1(t) + \dot f(t)  \vec{Y}_4 + \dot\kappa(t)\vec{Y}_5(t).
\label{eq:gammaDot4D}
\end{equation}
In general, the solution of the following differential equation represents the curves in this lifted domain: 
\begin{align}
\begin{split}
\dot\gamma(t) = (\alpha_1(t)\vec{Y}_1(t)+ \alpha_5(t)\vec{Y}_5(t)+ \alpha_4(t)\vec{Y}_4(t) )(\gamma(t))\\
\gamma(0) = (x_0,y_0,\theta_0,f_0, \kappa_0), \gamma(1) = (x_1,y_1,\theta_1,f_1,\kappa_1).
\end{split}
\end{align}
The horizontal distribution of planes is $\text{Span}\{\vec{Y}_1,\vec{Y}_5, \vec{Y}_4\}$. Denoting $Y_i$ the directional derivatives in the 
direction of the vectors $\vec{Y}_i$, the commutators of these vectors are as follows:

\begin{equation}
\begin{split}
[Y_1,Y_5]=-\partial_\theta=- X_2=- Y_2 \\
[[Y_1,Y_5],Y_1]=\sin(\theta)\partial_x - \cos(\theta)\partial_y =- X_3=- Y_3\\
[[Y_1,Y_5],Y_5] = 0.\\
[Y_i,Y_4]=0 \text{ for every }i.
\end{split}
\end{equation}
Then the Lie  algebra generated by $\text{Span}\{\vec{Y}_1,\vec{Y}_5, \vec{Y}_4\}$ is the whole space at every point.


The cortical connectivity can be modeled by a stochastic counterpart of (\ref{eq:gammaDot4D}). The Markov process that results from 
the Brownian motion with randomly curved paths has been introduced by~\cite{august2003sketches}. The process is represented by the following differential equations: 
\begin{equation}\label{eq:system}
\gamma' = \vec Y_1 + N(0,\sigma_1^2)\vec Y_4 + N(0,\sigma_2^2)\vec Y_5
\end{equation}
where $N(0,\sigma_i{^2})$ is a normally distributed variable with zero mean and variance equal to $\sigma_i^2$.  If $p_5$ denotes the probability density to find a particle at a point $(x,y)$ with a certain direction $\theta$, intensity $f$  and curvature $\kappa$, at a specific time $t$, then the Fokker-Planck equation describing the diffusion of the particle density will be:
\begin{equation}
\begin{split}
\partial_t p_5=  \frac{ \sigma_5^2}{2} Y_{55}p_5 + \frac{ \sigma_4^2}{2} Y_{44}p_5- Y_1 p_5
\end{split}
\label{eq:FP4D}
\end{equation}
so that $Y_{44} = {\partial^2}/{\partial f^2}$ and $Y_{55} = {\partial^2}/{\partial\kappa^2}$.  
This partial differential equation means that a particle at a point $(x,y,\theta,f,\kappa)$ transports in the direction of $\vec{Y_1} = (\cos(\theta),\sin(\theta),\kappa,0,0)$ in the 5D space. There is no  transport in the $f$ or $\kappa$ direction, but the diffusion in the $\kappa$ direction indicates the rate of transport in the $\theta$ direction. The diffusion in direction  $f$ is independent from the other variables.
The non-negative fundamental solution of  (\ref{eq:FP4D}) satisfies the following equation:

\begin{equation}
	\begin{split}
 \frac{ \sigma_4^2}{2}  Y_{44}\Gamma_5((x,y,\theta,f, \kappa),(x',y',\theta',f', \kappa'))+ \\+   \frac{ \sigma_5^2}{2}Y_{55} \Gamma_5((x,y,\theta,f, \kappa),(x',y',\theta',f', \kappa')) + \\ -Y_1 \Gamma_5((x,y,\theta,f, \kappa),(x',y',\theta',f', \kappa')) = 
\delta(x,y,\theta,f, \kappa).
	\end{split}
\label{eq:gamma}
\end{equation}
A good estimate of this solution is a section of fundamental solutions with $\kappa$ fixed ($\Gamma'_\kappa$ as a 3D kernel) symmetrized and multiplied to
an exponential term which considers the closeness between two points located in the different intensity planes, times 
an exponential term in the curvature planes. Hence, this new connectivity kernel is presented as:
\begin{equation}
	\begin{split}
	 w_5((x,y,\theta,f, \kappa),(x',y',\theta',f', \kappa')) =\\ e^ {- \frac{(\kappa - \kappa')^2} {\sigma_\kappa^2}} \times e^ {- \frac{(f - f')^2}{\sigma_{int}^2}} \times \\
	 \frac{1}{2} \bigg(  \Gamma'_\kappa((x,y,\theta),(x',y',\theta'))  + \Gamma'_{\kappa'}((x',y',\theta'),(x,y,\theta))   \bigg)
	\end{split}
	\label{eq:connectKernel}
\end{equation}


%

 In previous works \cite{favali2015local,favali2016analysis,sarti2015constitution} the connectivity kernel has been equivalently used as an affinity matrix describing the probability of existence of a connection between two points in the lifted domain. Similarly, the new kernel can be used for defining the affinity matrix as:
 \begin{equation}
 	A_{i,j} = w_5((x_i,y_i,\theta_i,f_i,\kappa_i),(x'_j,y'_j,\theta'_j,f'_j,\kappa'_j)).
 	\label{eq:affinity}
 \end{equation}
 
 By this definition, considering the fact that this matrix includes information about the correct grouping, this problem has been presented in terms of dimensionality reduction of this matrix \cite{favali2015local,sarti2015constitution}, often done by eigensystem analysis, which we will explain in more details in section \ref{sec:sepctral}.

\subsection{Numerical Approximation of the 5D Kernel}
\label{4Dapprox}
The kernel is numerically estimated using the general Markov Chain Monte Carlo method technique. We say a few words on how to adapt it to the 5D case for reader convenience. 
The system in equation \eqref{eq:system} can be approximated by:
\begin{equation}
\begin{cases} 
x_{s+\Delta{s}} - x_s = \Delta{s} \cos(\theta) \\ 
y_{s+\Delta{s}} - y_s = \Delta{s} \sin(\theta) , \mbox{\quad} s\in{1,...,H} \\
\theta_{s+\Delta{s}} - \theta_s = \Delta{s} \kappa\\
\kappa_{s+\Delta{s}} - \kappa_s = \Delta{s} N(0,\sigma_\kappa)
\end{cases}
\label{system_FP}
\end{equation}
where $H$ is the number of steps of the random path and $N(0,\sigma_\kappa)$ is a generator of numbers taken from a normal distribution with mean 0 and standard deviation of $\sigma_\kappa$. 
The stochastic path is obtained  from the estimate of the kernel as the average of their passages over discrete volume elements, solving this finite difference equation $n$ times \cite{sarti2015constitution}. The affinity matrix described in  (\ref{eq:affinity}) is evaluated from this kernel.

\begin{figure}
    \centering
    \begin{subfigure}[b]{0.45\columnwidth}
       \includegraphics[width=\textwidth]{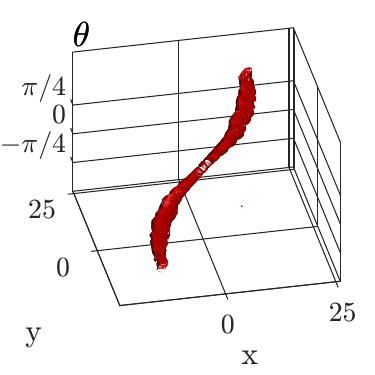} \caption{} \label{fig:isosurf-theta}
    \end{subfigure}
    ~ 
    \begin{subfigure}[b]{0.45\columnwidth}
	 	\includegraphics[width=\textwidth]{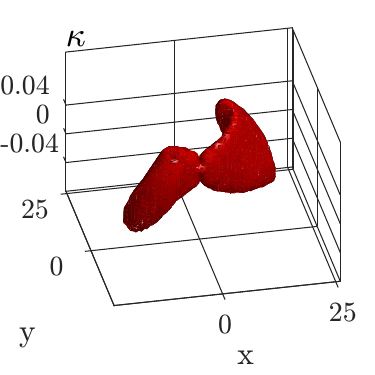} \caption{} \label{fig:isosurf-kappa}
    \end{subfigure}
    ~ 
    \begin{subfigure}[b]{\columnwidth}
 		\includegraphics[width=0.96\textwidth]{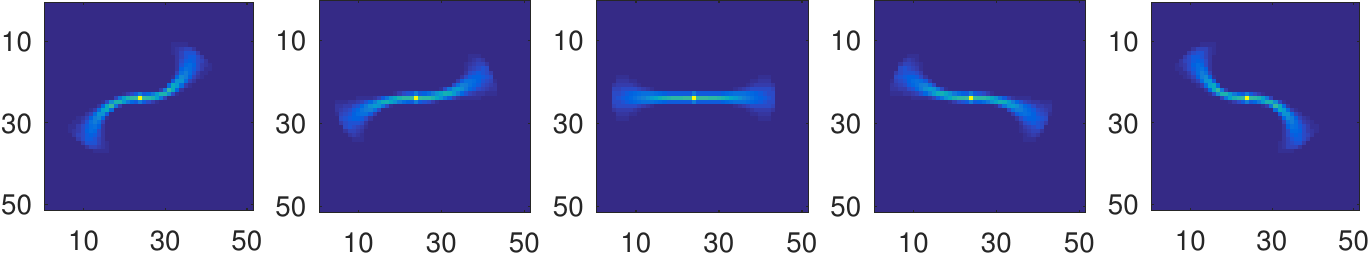}  \caption{} \label{fig:2DprojectionKernels}
    \end{subfigure}
    \caption{Visualizations of the 5D stochastic kernels in 3D and 2D. \subref{fig:isosurf-theta}) and \subref{fig:isosurf-kappa}) the iso-surfaces of the kernel while keeping $\kappa$ and $\theta$ fixed respectively; \subref{fig:2DprojectionKernels}) the 2D projection of the kernel over all orientations for several curvature values: $\kappa=\{-0.08,-0.04,0,0.04,0.08\}$ from left to right. Intensity ($f$) is constant for all figures.}\label{fig:5Dkernel}
\end{figure}

Fig.~\ref{fig:5Dkernel} represents two level sets of the 5D kernel by fixing $\kappa$ (Fig.~\ref{fig:isosurf-theta}) and $\theta$ (Fig.~\ref{fig:isosurf-kappa}) dimensions. The 2D projection on $\mathbb{R}^2$ by summation over all orientations of five different 4D stochastic kernels having different curvature ($\kappa$) values is also presented by Fig.~\ref{fig:2DprojectionKernels}. The intensity term is kept constant for all figures. As seen in these figures, by increasing the absolute value of curvature, the shape of the kernel also changes and it deviates from the elongated shape.

\subsection{Lifting Retinal Images in the 5D Space}
\label{sec:liftRetina}
In this section, we explain how we lift the 2D retinal images to our proposed feature space of positions, orientations, curvatures and intensity in practice.
\subsubsection{Orientation Score Transform}
\label{sec:OST}
Being inspired and supported by several electrophysiological experiments, the receptive profiles of simple cells in the primary visual cortex are often interpreted as oriented Gabor filters or directional derivatives of the Gaussian  filters~\cite{daugman1985uncertainty,florack1992scale,lee1996image,citti2014neuromathematics}. Moreover, ~\cite{duits2007invertible,duits2007image} introduced the invertible orientation score (OS) transformation for lifting the image to the $SE(2)$ space. The invertibility property prevents information loss during the transformation, which is guaranteed by certain requirements of the kernel used for the transformation. The cake wavelets introduced by~\cite{duits2005perceptual,duits2009line} are proper wavelets, which satisfy these criteria. Similar to the Gabor wavelets, they are quadratic anisotropic filters, but unlike these, their summed Fourier transformations cover the entire frequency domain, making them spatially scale-independent. Therefore, it is possible to disentangle the crossing elongated structures in a 2D image from each other regardless of their scale and without loss of image evidence. In order to construct the orientation score $U_I(\mathbf{x},\theta)$, the image ($I$) is convolved with rotated and translated versions of a mother wavelet ($\psi$) as
\begin{equation} 
 \begin{split}
U_I(\mathbf{x},\theta)=(\overline{R_\theta(\psi)} \star I)(\mathbf{x})= \\ \int_{\mathbb{R}^2} \overline{\psi(R^{-1}_\theta (\mathbf{y}-\mathbf{x}))}I(\mathbf{y})\mathrm{d}\mathbf{y}
 \end{split}
 \label{eq:OST}
 \end{equation}
 where $R_\theta$ is the 2D counter-clockwise rotation matrix, the overline denotes the complex conjugate and $\star$ denotes the convolution~\cite{duits2007invertible,duits2007image}. 

Applying this tool for lifting the retinal images, considering the Gaussian profile of the blood vessels and the fact that the blood vessels have a darker intensity compared to the background, the real part of the negative OS response ($-U_I(x,y,\theta)$) is used for obtaining the orientation in these images. Because of the quadratic property of the cake wavelets, the real symmetric part of the transformation performs as a vessel ridge detector. Moreover, for each location the response  may be different from zero in more than one orientation; therefore, the orientation at which the maximum response is obtained at location $\mathbf{x}$ is the dominant orientation $\theta_d$ at that location:
\begin{equation}
\theta_d(\mathbf{x}) =  \argmax_{\theta\in\{-\pi/2,\pi/2-\pi/n_{\theta}\}}Re(-U_I(\mathbf{x},\theta)),
\label{ed:dominantOrientation}
\end{equation}
so that $\theta$ only takes $n_\theta$ discrete values.
\subsubsection{Curvature Extraction Technique}
\label{sec:curvature}
As mentioned before, the curvature determines the rate of orientation change ($\dot\theta(t)=\kappa(t)$). 
If the parametric representation of the curves $\gamma_{2D} = (x(t),y(t))$ in the image is available, then by differentiating $x$ and $y$ once more in (\ref{eq:tangent2D}), we will have:
\begin{equation}
(\ddot x,\ddot y) = (-\sin(\theta(t))\dot\theta(t),\cos(\theta(t))\dot\theta(t))
\end{equation}
where $t$ is the arc-length of the curve. 
So the curvature can be computed as:
\begin{equation}
\kappa =\dot\theta = \frac{\dot x \ddot y- \ddot x\dot y}{(\dot x^2+ \dot y^2)^{3/2}}.
\label{eq:kParametric}
\end{equation}
In computer vision and more specifically in retinal image analysis, several methods have been introduced for measuring the curvatures of curvilinear structures (i.e. blood vessels)~\cite{kalitzeos2013retinal}. The classical methods that measure the curvatures locally need an initial segmentation, centreline extraction, and separation of segments located between junctions. It is then followed by fitting curves to the segments and by curvature measurement using (\ref{eq:kParametric}) (e.g.~\cite{annunziata2016fully}). The drawback of all these methods is their dependency to initial preprocessing and segmentation steps, which may contain errors and missing information. More importantly, the curvature information is not available for junction points because it is not possible to fit a curve to these points where more than one elongated structure meet. 

To solve these problems,~\cite{Bekkers2015} proposed a local curvature measurement technique by locally fitting exponential curves~\cite{duits2016locally} to the lifted image in $SE(2)$. The exponential curves in $SE(2)$ are interpreted as straight lines considering the curved geometry of $SE(2)$ and they have constant tangent vectors relative to the rotating frame $\{\vec{X_1},\vec{X_2},\vec{X_3}\}$. The tangent vectors of the exponential curve that best fit the data in the lifted image are obtained by eigensystem analysis of the Gaussian Hessian (expressed in the rotating frame). Then they directly define the curvature value of their spatial projections \cite{franken2009crossing}. This approach makes it possible to assign to each location and orientation in the lifted image a curvature value, without needing explicit curve parameterizations. Such curvature maps (on SE(2)) can be projected on the plane whereby only one value of curvature value is assigned to each spatial location in the image. Finally, these 2D curvature maps can be filtered in a later stage by a vessel confidence map such as $s_\sigma(\mathbf{x},\theta)$, as a Laplacian ridge detector at a single scale ($\sigma$) (See ch. 5.3,~\cite{franken2009crossing} for detailed explanations). 

In order to obtain the curvature value for each location ($\kappa_{map}(\mathbf{x,\theta})$) in our 5D lifted retinal images an altered version of the above-described method is used. The method performs the best when the spatial scale ($\sigma$) of the Gaussian derivatives used to calculate the Gaussian Hessian matrix matches the vessel width. By using a single scale, the curvature values are accurate only for the vessels which their width match the used scale. Since the blood vessels in retinal images have different widths, using a multi-scale approach helps in covering various vessel  widths available in the image and getting more accurate results. Therefore, we altered the method to a multi-scale approach. To do so, confidence and curvature maps have been obtained for several scales ($s_{\sigma_i}(\mathbf{x},\theta)$ and $\kappa_{\sigma_i}(\mathbf{x},\theta)$, $i=1,\dots,n$). Then the final curvature map is constructed by assigning to each pixel the curvature value that corresponds to the scale at which the largest confidence value has been obtained, i.e. $\forall(x,y,\theta)\in \mathbb{R}^2\times S^1$:
\begin{equation}
\begin{split}
\kappa_{map}(\mathbf{x},\theta) = \\ 
\{\kappa_{\sigma_{max}}(\mathbf{x},\theta) |
\sigma_{max} = \argmax_{\sigma_i\in\{\sigma_1.\dots \sigma_n\}}s_{\sigma_i}(\mathbf{x},\theta)\}.
\end{split}
\label{eq:curv}
\end{equation}
\subsubsection{The Intensity Information}
For a retinal image $I_\circ\in\mathbb{R}^2$, an appropriate  pre-processing step is needed to normalize the luminosity and contrast all over the image, reduce noise and to enhance the blood vessels. We use the same pre-processing technique proposed by~\cite{abbasi2015biologically} on the green and red channels of the original image separately and combine them as $I(\mathbf{x}) = \sqrt{I_{g}^2(\mathbf{x}) + I_{r}^2(\mathbf{x})}$, where $I_{g}$ and $I_{r}$ are preprocessed red and green channels. This step is essential as it helps in increasing the difference between the intensity of arteries and veins and helps in better discrimination of blood vessels crossing each other. The final value of intensity at position $\mathbf{x}\in\mathbb{R}^2$ of image $I$ is considered as $f_{map}(\mathbf{x}) = I(\mathbf{x})$.
\subsubsection{Set of Interest Positions}
As mentioned in section~\ref{sec:intro}, vessel segmented images may contain vessel disconnections or wrongly detected vessel pixels. However, we use the segments to get an initial estimation of the location of the blood vessels and only focus on those locations. This helps in reducing the computation complexity as the size of affinity matrix (\ref{eq:affinity}) reduces tremendously. The outcome of segmentation can be either a probability map showing the probability for each pixel to be part of a vessel or it can be a deterministic binary map, in which each pixel is labeled as a vessel (label $1$) or background (label $0$). In this work, we use the deterministic binary segmentations obtained by the method proposed by~\cite{abbasi2015biologically} for both RGB and SLO retinal images. 
If the vessel segmented image is called as $I_{seg}$ then the vessel locations are found as:
\begin{equation}
v=\{\mathbf{x} | I_{seg}(\mathbf{x}) = 1,\mathbf{x}\in\mathbb{R}^2\}.
\label{eq:interestPoints}
\end{equation}
This set provides a set of interest points as we only consider the information at these locations for our connectivity analysis. 

To summarize this section, for an image $I_{\circ}$, after preprocessing the image and obtaining $I$, we use the orientation score transform  and the segmentation technique on $I$ to get $U_I$ and $I_{seg}$ respectively. Then by finding the set of interest points using (\ref{eq:interestPoints}), we create a 5D feature map called $U_{I,5D}\in\mathbb{R}^2\times S^1\times \mathbb{R}^+ \times \mathbb{R}$ as:
\begin{equation}
\begin{split}
U_{I,5D}(\mathbf{x})=  \\
 \{(\mathbf{x},\theta,f,\kappa)~|~\theta = \theta_d(\mathbf{x}),f=f_{map}(\mathbf{x}),\kappa=\kappa_{map}(\mathbf{x})\}, \\ 
\forall \mathbf{x} \in v_I 
\end{split}
\label{eq:5Dretina}
\end{equation}

\subsection{An Efficient Cortically Inspired Spectral Clustering}
\label{sec:sepctral}
%
%

In modern data analysis clustering is an important research topic and many algorithms have been proposed in this area (e.g.~\cite{shi2000normalized,perona1998factorization,von2007tutorial}) where the eigenvectors of the graph Laplacian are used to analyse and reveal the cluster structure of the data. 
In our previous work \cite{favali2016analysis}, the problem of grouping the blood vessels in retinal images (interpreted as perceptual units) was solved by proposing a semi-automatic approach based on spectral analysis of the affinity matrix. So the salient groups were obtained as the first eigenvectors corresponding to the largest eigenvalues. However a fixed threshold was defined for selecting the largest eigenvalues. In order to avoid the manual intervention, we follow the method proposed by \cite{zelnik2004self} in which this problem is solved. 

Unlike more basic clustering methods, as k-means, where the number of clusters and interaction distances of the nodes in the dataset must be assumed a priori, this algorithm automatically and optimally tunes these parameters. In this algorithm,
the structure of the eigenvectors is used to determine the number of groups. A cost function is defined and evaluated from the alignment of the eigenvectors. Consecutively, the best number of clusters is considered as the one which minimizes the cost function. Correspondingly, the best clustering quality $Q_{clust}$ that has a reverse relation to the alignment cost is obtained (see~\cite{zelnik2004self} for detailed explanations). In the final step, the noisy elements that construct small sized groups are removed. From a neural approach, the fact that in the presence of a visual stimulus the perceptual units are defined by the emerging eigenvectors is described by \cite{sarti2015constitution}.

The full pipeline is presented in Algorithm~\ref{alg:ConnectAnal}. 
We assume an image $I_\circ\in R^2$ has been given as input. The first step is to lift the image to the 5D feature space (as explained in section~\ref{sec:liftRetina}). Then we need an affinity matrix which is created after modelling the cortical connectivity in the 5D space as we proposed. Finally, the self-tuning spectral clustering algorithm returns the final perceptual units in that image $C = \{c_1,\dots,c_{K}\}$, where $K$ is the final number of clusters and $c_i$ includes set of points in image $I$ which belong to cluster $i$. 
\begin{algorithm}
 	\caption{Proposed perceptual grouping technique for a given image $I$$\in \mathbb{R}^2$} 
 		\begin{algorithmic} [1]
 		\label{alg:ConnectAnal}
 			\STATE Lift the image $I(x,y)\in\mathbb{R}^2$ to the 5D space of $U_{I,5D}(x,y,\theta,f,\kappa)\in\mathbb{R}^2\times S^1\times \mathbb{R} \times \mathbb{R}$ (\ref{eq:5Dretina}). \label{st:OST}
			\STATE Calculate all the fundamental solutions of (\ref{eq:gamma}), $\Gamma'_\kappa((x,y,\theta),(x',y',\theta'))$ and  $\Gamma'_{\kappa'}((x',y',\theta'),(x,y,\theta))$ stochastically, for all $\{(x,y,\theta,\kappa),(x',y',\theta',\kappa')\}$ in $U_{I,5D}$. \label{st:fundSol}
			\STATE Calculate the final connectivity kernel $w_5((x,y,\theta,f, \kappa),(x',y',\theta',f', \kappa'))$ for all pairs of points in $U_{I,5D}$ (\ref{eq:connectKernel}).
			\STATE Create the affinity matrix (\ref{eq:affinity}).
			\STATE Apply the automatic spectral clustering technique for detecting the $K$ perceptual groups in image: $C = \{c_1,\dots,c_{K}\}$. 
 		\end{algorithmic}		
\end{algorithm}

\begin{figure*}[htbp]
  \centering 
  \begin{subfigure}[b]{1.15in}
  	\includegraphics{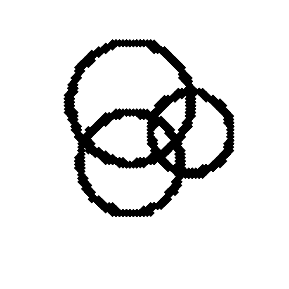}
   	\includegraphics[trim={0.1in  0.09in 0.2in 0.1in},clip]{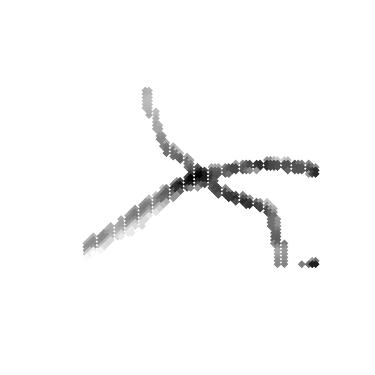} 
	\caption{} \label{fig:Orig}
   \end{subfigure}
    \begin{subfigure}[b]{1.5in}
  	\includegraphics{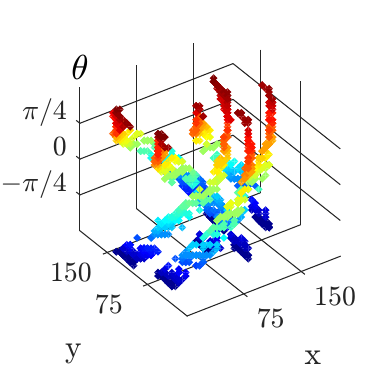}
   	\includegraphics{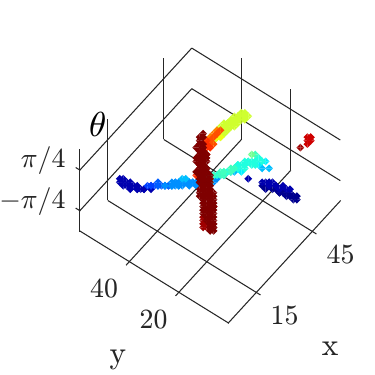} 
	\caption{} \label{fig:osmap3D}
   \end{subfigure}
    \begin{subfigure}[b]{1.5in}
  	\includegraphics{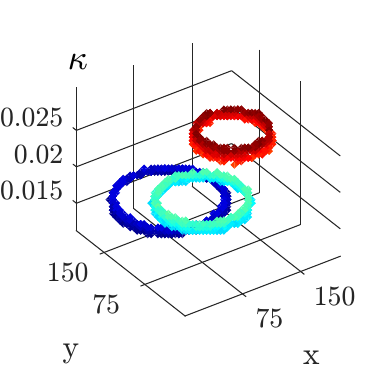}
   	\includegraphics{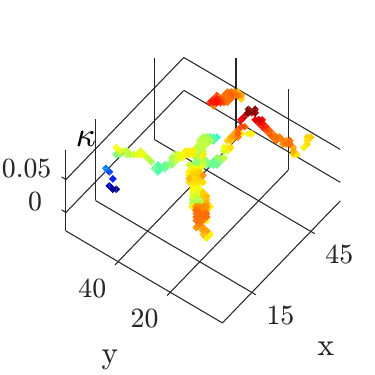} 
	\caption{} \label{fig:CurvMap3D}
   \end{subfigure}
    \begin{subfigure}[b]{1.5in}
  	\includegraphics{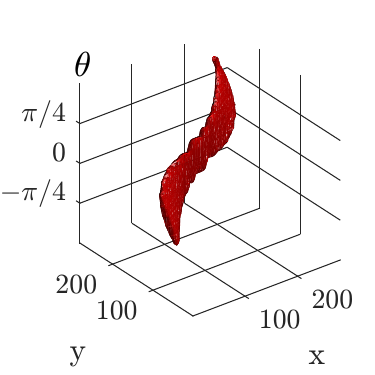}
   	\includegraphics{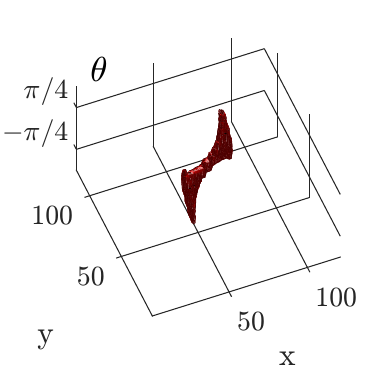} 
	\caption{} \label{fig:Isosurf}
   \end{subfigure}
    \begin{subfigure}[b]{1.15in}
  	\includegraphics{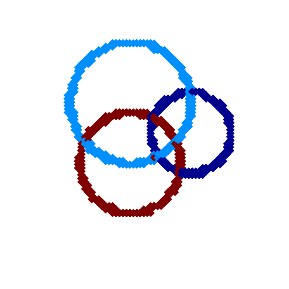}
   	 \includegraphics[trim={0.1in  0.09in 0.2in 0.1in},clip]{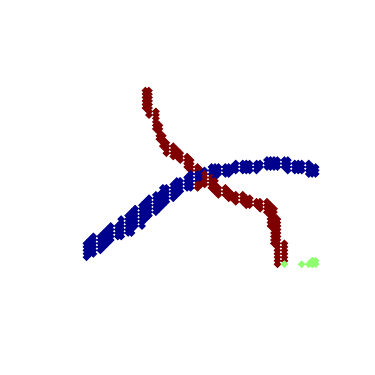} 
	 \caption{} \label{fig:Results}
   \end{subfigure}
   
	
  \caption{A synthetic image consisted of three crossing circles (first row) and a sample retinal image patch including two crossing vessels (the second row). The columns represent the original image, the orientation map ($\theta_d(\mathbf{x})$), the 3D curvature maps ($\kappa_{map}(\mathbf{x},\theta_d)$), a level set of the 5D kernel while keeping two dimensions fixed and the final detected clusters in different colors, respectively.}
  \label{fig:samplePhantomImage}
\end{figure*}

Fig.~\ref{fig:samplePhantomImage} depicts a sample application of the proposed method for clustering the perceptual units in both an artificial image (the first row) and a small patch of retinal images (the second row). The synthetic image includes three crossing circles with different radii and corresponding curvatures, and the retinal patch includes two crossing vessels.
For each case, the orientation and curvature of the lifted images are shown in the column (\subref{fig:osmap3D}) and (\subref{fig:CurvMap3D}) respectively. The intensity value for the synthetic image is constant all over the circles and for the retinal patch, it is color-coded in~(\subref{fig:Orig}). A sample level set of corresponding 5D kernels (while keeping two dimensions ($\kappa$ and $f$) fixed) has been shown in each row (\subref{fig:Isosurf}).
 Finally, (\subref{fig:Results}) shows the three detected groups in the synthetic image and three vessels in the retinal image in different colors. 
\section{Experiments}
\label{sec:experiments}
In this section, we present a potential application of the proposed connectivity analysis for solving the aforementioned problems in curvilinear structure tracking methods for retinal vasculature analysis (see section ~\ref{sec:intro}). After explaining the material used for validating the method, the details of the numerical simulation are described. Then the quantitative and qualitative results of the proposed technique are presented and discussed in detail. 
\subsection{Material}
Two datasets have been used for validating the method. The specifications and the preparation steps of each dataset are explained in detail as follows.
\subsubsection{Phantom images}
A set of synthetic images ($201\times201$) with known orientations and curvature values and constant intensities has been generated to include various rotated, curved and interrupted vessel-like structures. 
Five different groups are created to mimic possible structures that could be present in retinal images, similar to the categories proposed in our previous work~\cite{favali2016analysis}. These categories are
\begin{enumerate*}[label={(\Alph*)}]
\item crossings; 
\item bifurcations;
\item  close parallel vessels;
\item bifurcations and crossings; and 
\item vessels with multiple nearby bifurcations
\end{enumerate*}.
Each of these categories may also include challenging structures. For instance, they may be interrupted or highly curved or include small junction (crossing/ bifurcation) angles. In order to differentiate between the simple and the challenging cases for each category, we name group $X$ as $X1$ if it is challenging.  Fig.~\ref{fig:phantomResults} depicts ten different phantom images in each row of the first column, two per category, together with their color-coded orientation and curvature maps (second and third columns). 

The basic element used for creating these phantom images is a sine wave-like structure, which is generated with several frequencies and amplitudes and it is rotated and located at different positions depending on the target shape. By adjusting the frequency and amplitude of the waves, different curvature values can be created. In addition to the vessel-like structures, other challenging structures such as dashing and the Euler spiral have been also used to examine the strength of the method in grouping these curved structures (e.g. Fig.~\ref{fig:phantomResults}, A1). 

\subsubsection{Retinal images}
The public IOSTAR\footnote{Available at: \hyperref[label_name]{http://www.retinacheck.org/datasets}} dataset contains images captured using scanning laser ophthalmoscope (SLO) technology\footnote{i-Optics BV, the Netherlands}. These high contrast images have a resolution of $1024\times1024$ with 45$^{\circ}$  FOV. The blood vessels, junctions and artery/vein labels have been annotated  for 24 images and corrected by two different experts in order to decrease the inter-user variability. 
To reduce the computation complexity these images and their annotations are downsampled to half size $512\times512$. Our main purpose is to emphasize the strength of the method in individuation of vessels, specifically at junction points or challenging cases, where most of the proposed methods in the literature face difficulties. Since the images in this set are very noisy and challenging for segmentations, they are good cases for examining our proposed method in detecting missing connections. 

\begin{figure*}[htbp]
  \centering 
 	 \begin{subfigure}[b]{1.4in}
        		 \includegraphics[width=1\textwidth]{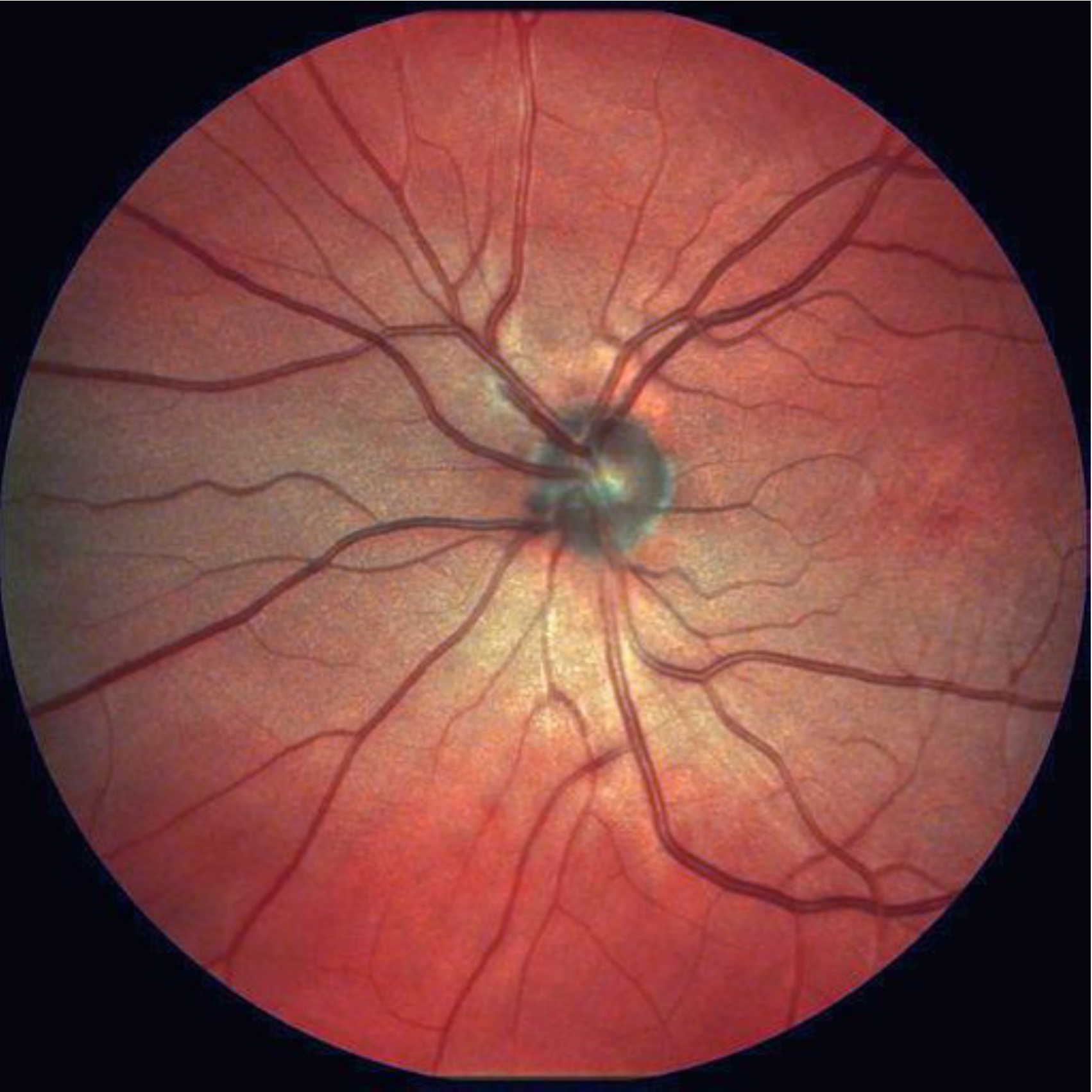}  \caption{} \label{fig:colored}
        	\end{subfigure}
	\begin{subfigure}[b]{1.4in}
        		 \includegraphics[width= 1\textwidth]{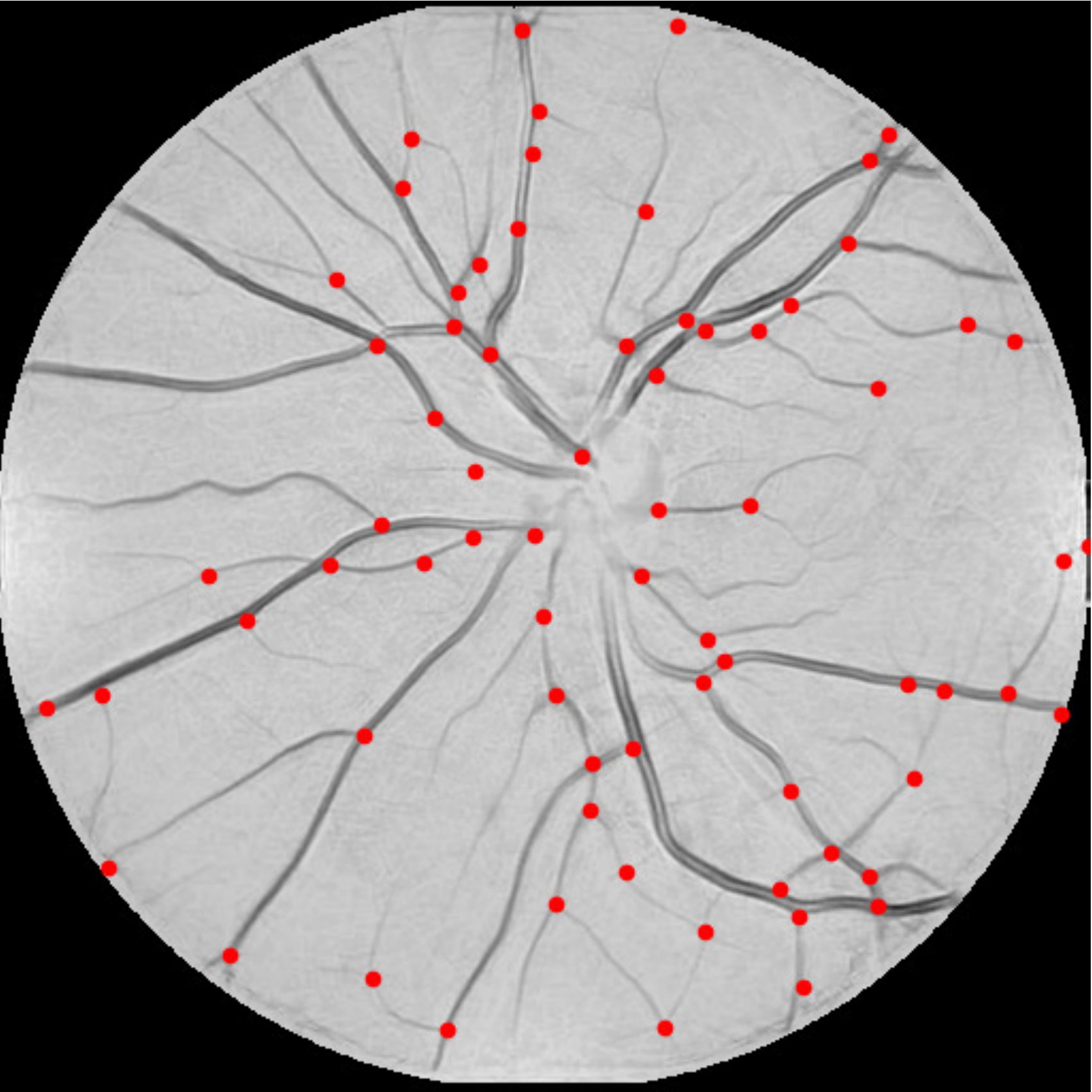} \caption{}  \label{fig:Junctions}
        	\end{subfigure}
	\begin{subfigure}[b]{1.4in}
        		 \includegraphics[width=1 \textwidth]{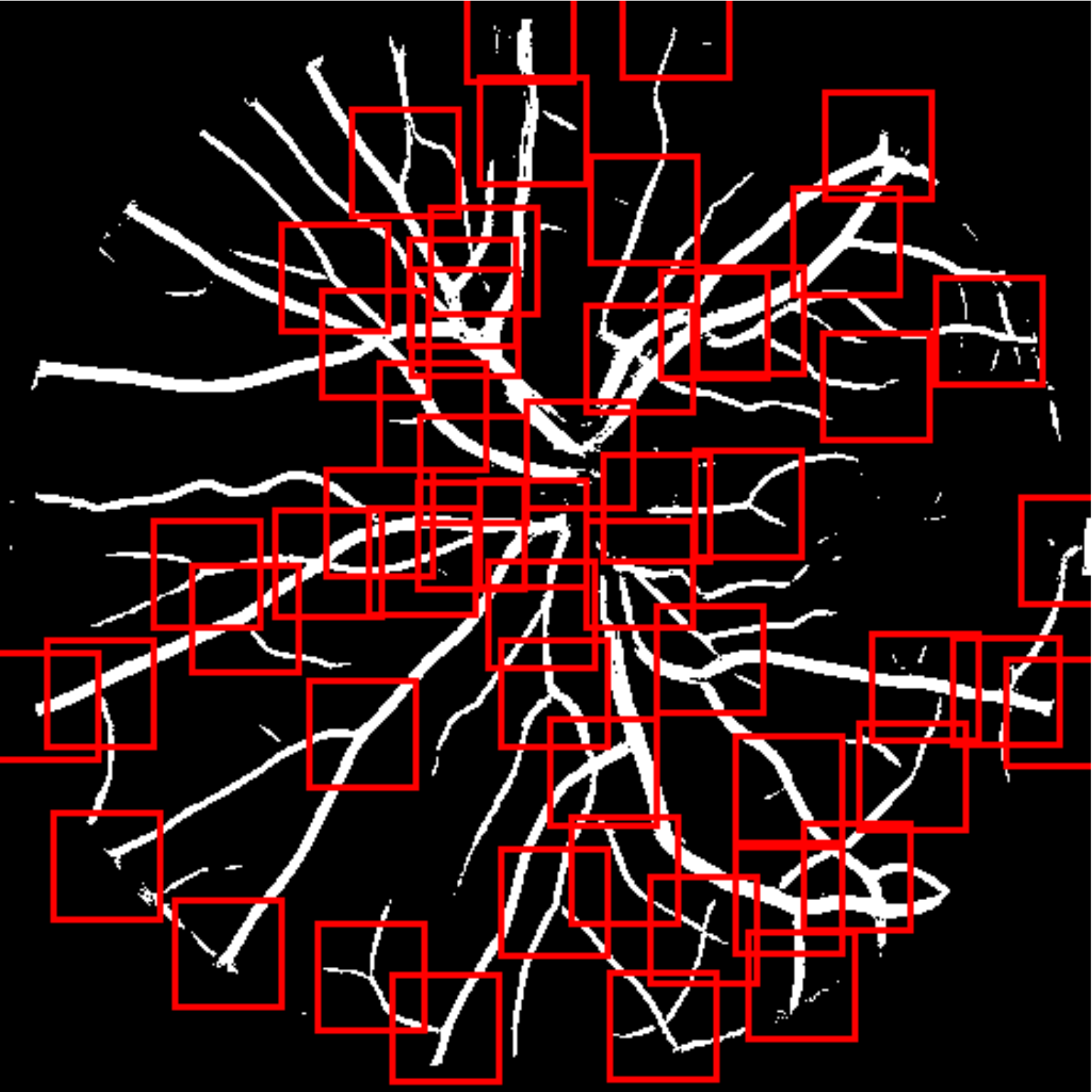} \caption{} \label{fig:Patches}
        	\end{subfigure}
	\begin{subfigure}[b]{1.4in}
        		 \includegraphics[width= 1\textwidth]{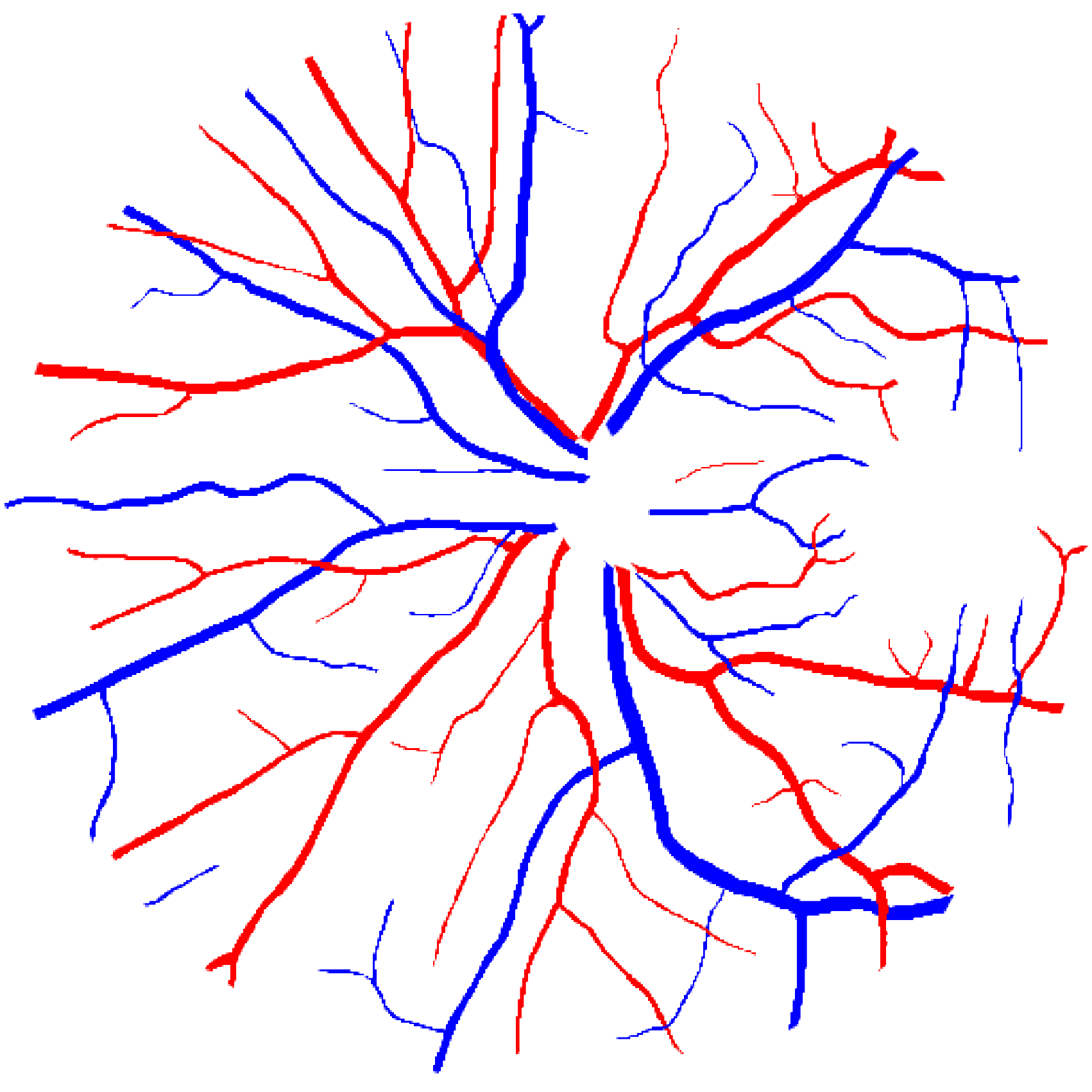} \caption{}  \label{fig:vesselGT}
        	\end{subfigure}
	 \begin{subfigure}[b]{1.8in}
        		 \includegraphics[width=\textwidth]{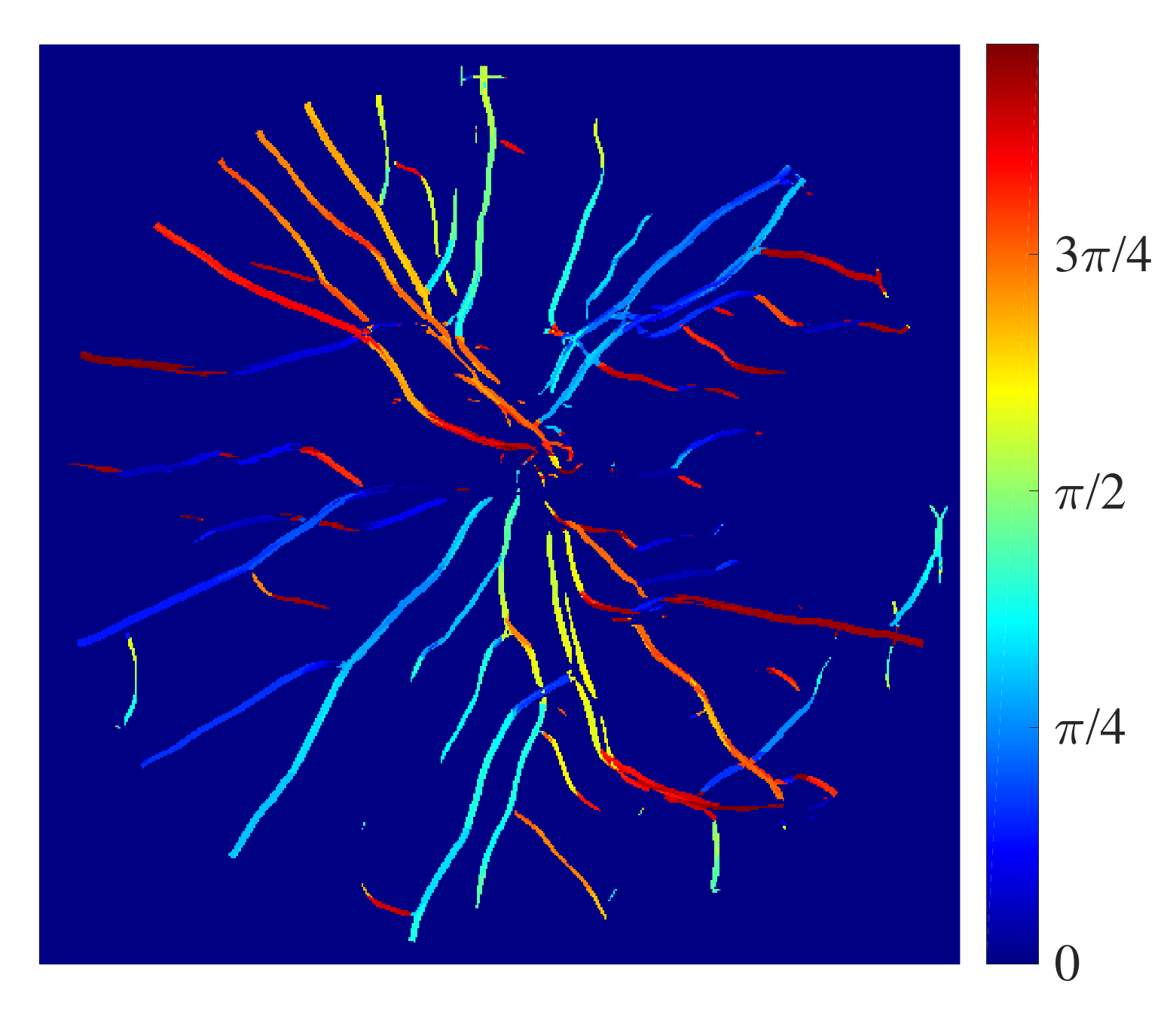}  \caption{} \label{fig:sloOSMap}
        	\end{subfigure}
	\begin{subfigure}[b]{1.8in}
        		 \includegraphics[width= 1\textwidth]{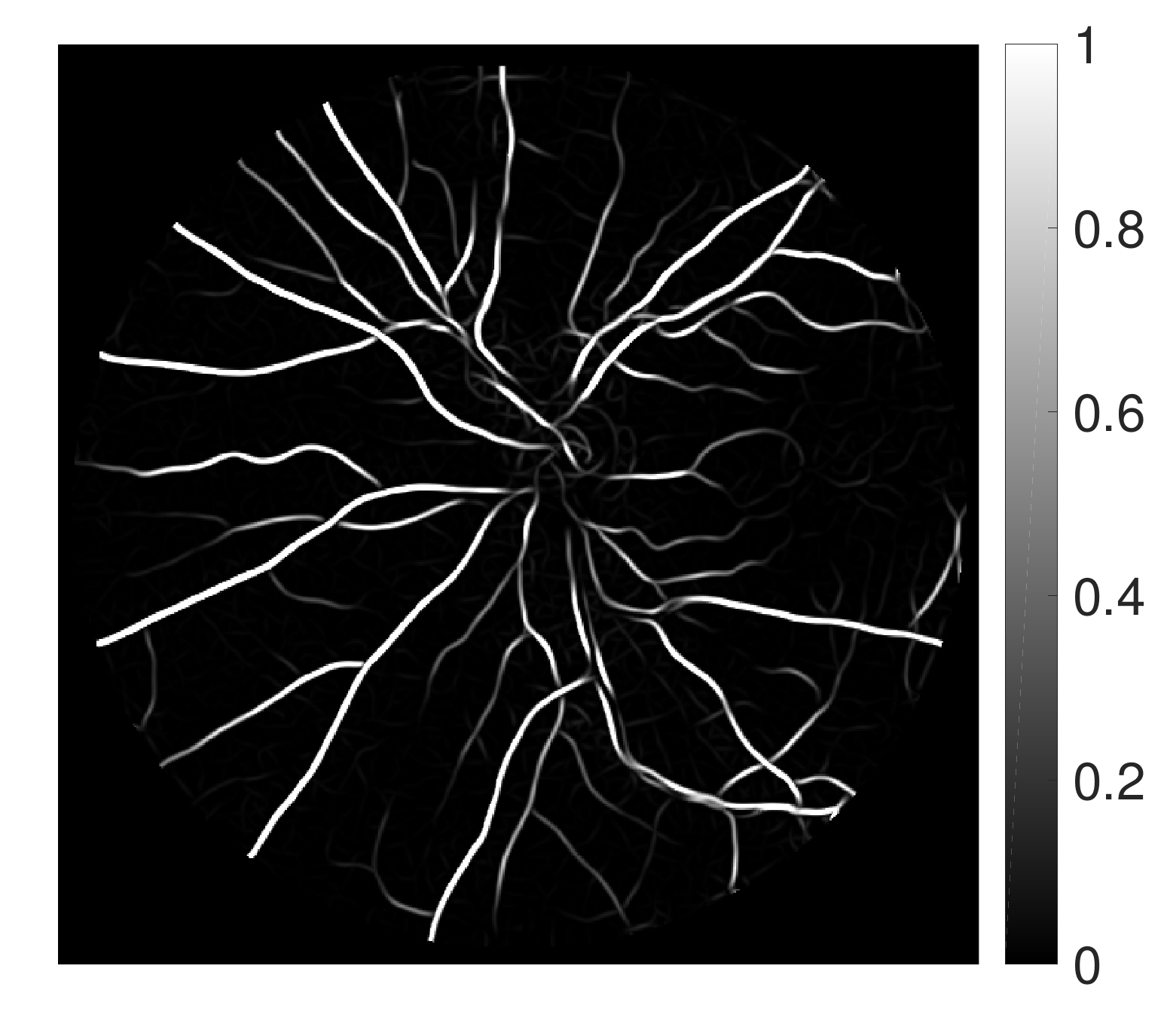} \caption{}  \label{fig:sloConfMap}
        	\end{subfigure}
	\begin{subfigure}[b]{1.8in}
        		 \includegraphics[width=1 \textwidth]{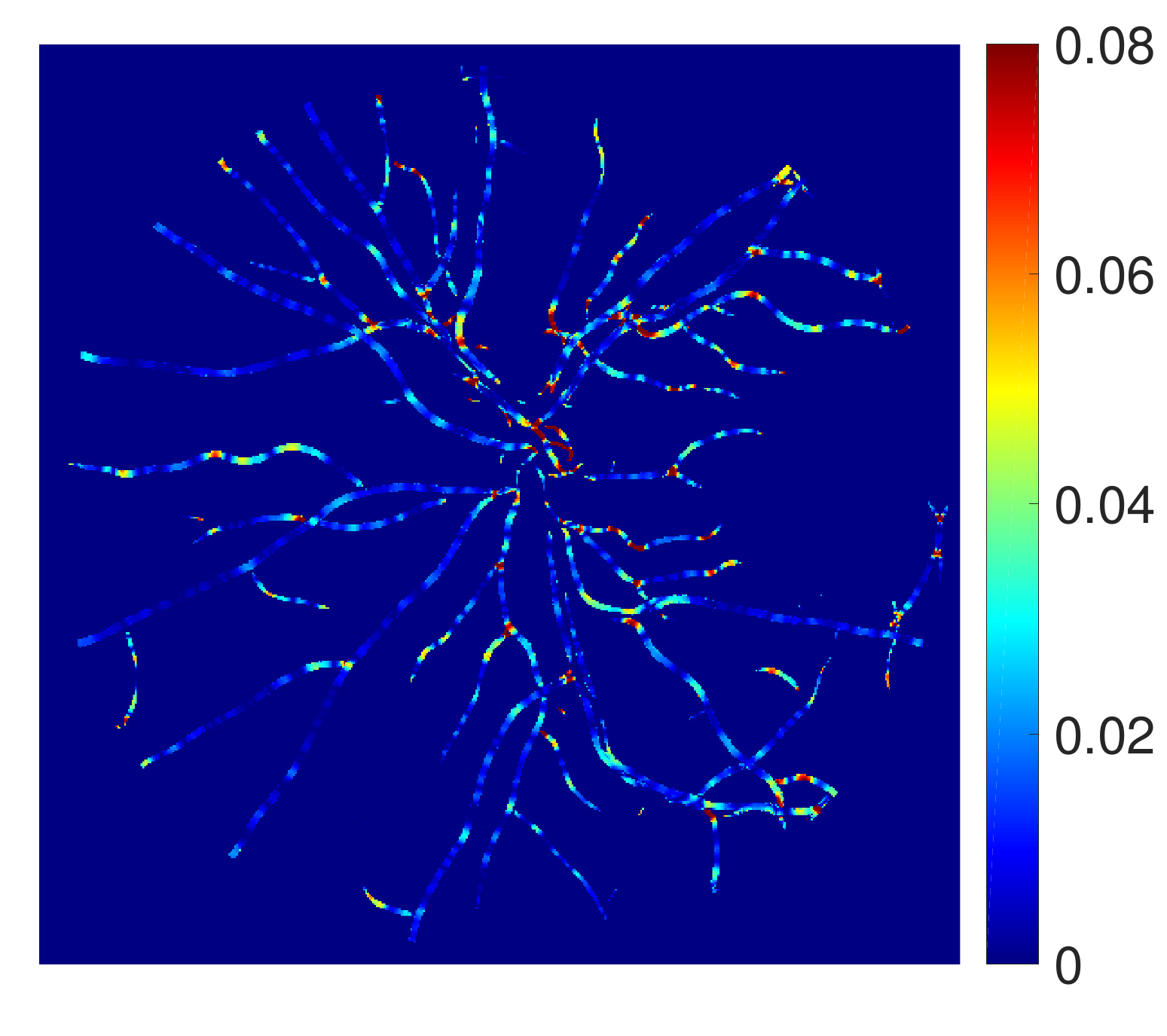} \caption{} \label{fig:sloCurvMap}
        	\end{subfigure}
  \caption{A sample SLO image, (\subref{fig:colored}) original image ($I_o$), (\subref{fig:Junctions}) detected junctions overlaid on the enhanced image ($I$), (\subref{fig:Patches}) selected patches overlaid on the vessel segmentation ($I_{seg}$), (\subref{fig:vesselGT}) the artery/vein ground truth, the color-coded (\subref{fig:sloOSMap}) orientation, (\subref{fig:sloConfMap}) confidence,  and (\subref{fig:sloCurvMap}) absolute curvature maps of this SLO image.}
  \label{fig:SLOimagePatchSelection}
\end{figure*}

Following the procedure described in section ~\ref{sec:liftRetina}, we lift each image to the 5D space after preprocessing and segmenting the images. The scales used for obtaining the curvature map for these SLO images are $\{1.5,2.5,3.5\}$ in pixels and $n_\theta=18$. Even though we lift full images, we only consider the information around junction points. In order to crop small patches, we detect the junction points in each image using the hybrid method proposed in~\cite{abbasi2016automatic}, and call the set of junction locations as $\epsilon_i = \{x_i,y_i\},~i=1,\dots,M$, where $M$ is the number of detected junctions in image $I$. If we assume that image $I$ is lifted to $U_{I,5D}(x,y,\theta,f,\kappa)\in\mathbb{R}^2\times S^1\times \mathbb{R}^+\times\mathbb{R}$, then the cropped patch at location $\epsilon_i$ ($\forall i=1,\dots,M$) is defined as:
\begin{equation}
\begin{split}
U^{i}_{I,5D}(x,y,\theta,f,\kappa) = \\
 \{U_{I,5D}(x,y,\theta,f,\kappa)|  x_i-s_\circ \leq  x \leq x_i+s_\circ, \\
 y_i-s_\circ \leq  y \leq y_i+s_\circ \},
\end{split}
\label{eq:CrpLiftImg}
\end{equation}
where $s_o$ determines the size of the neighborhood considered around each junction. We fix $s_o = 25$ pixels for all patches. At the end, the size of each 5D patch is $(2s_0+1)\times(2s_0+1)\times 5$. In total 272 random patches around junctions points, with resolution of $51\times 51\times 5$ have been selected from this set. Similar to the phantom images, these patches are also categorized in five different groups depending on their structure and complexity.
Fig.~\ref{fig:SLOimagePatchSelection} depicts a sample SLO image of the IOSTAR dataset (\subref{fig:colored}). It also represents the color-coded orientation (\subref{fig:sloOSMap}), confidence (\subref{fig:sloConfMap}), curvature (\subref{fig:sloCurvMap}) maps, detected junction points overlaid on the preprocessed image (\subref{fig:Junctions}), the corresponding neighborhood around each junction location (\subref{fig:Patches}) and the artery-vein labels of the vessels (\subref{fig:vesselGT}), which are later used for validation. Note that the depicted confidence and curvature maps are related to one single scale $\sigma = 1.5$, and the absolute curvature value is shown. 
\subsection{Validation}
\subsubsection{Phantom images}
\begin{figure*}[htbp]
  \centering
 	\begin{subfigure}[b]{6.8in}
 				\hspace{1.8in}
	  			\includegraphics[width=1in]{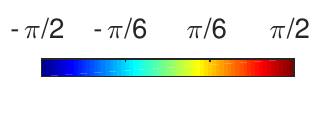}\hspace{0.18in}
	  			\includegraphics[width=1in]{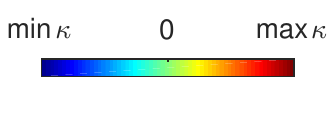} 
	\end{subfigure}
	\begin{subfigure}[b]{6.8in}
	\centering
 			 \makebox[0pt][r]{\makebox[15pt]{\raisebox{25pt}{\rotatebox[origin=c]{0}{A1}}}}  \qquad
 			 \includegraphics[width=\sizp,height=\sizp]{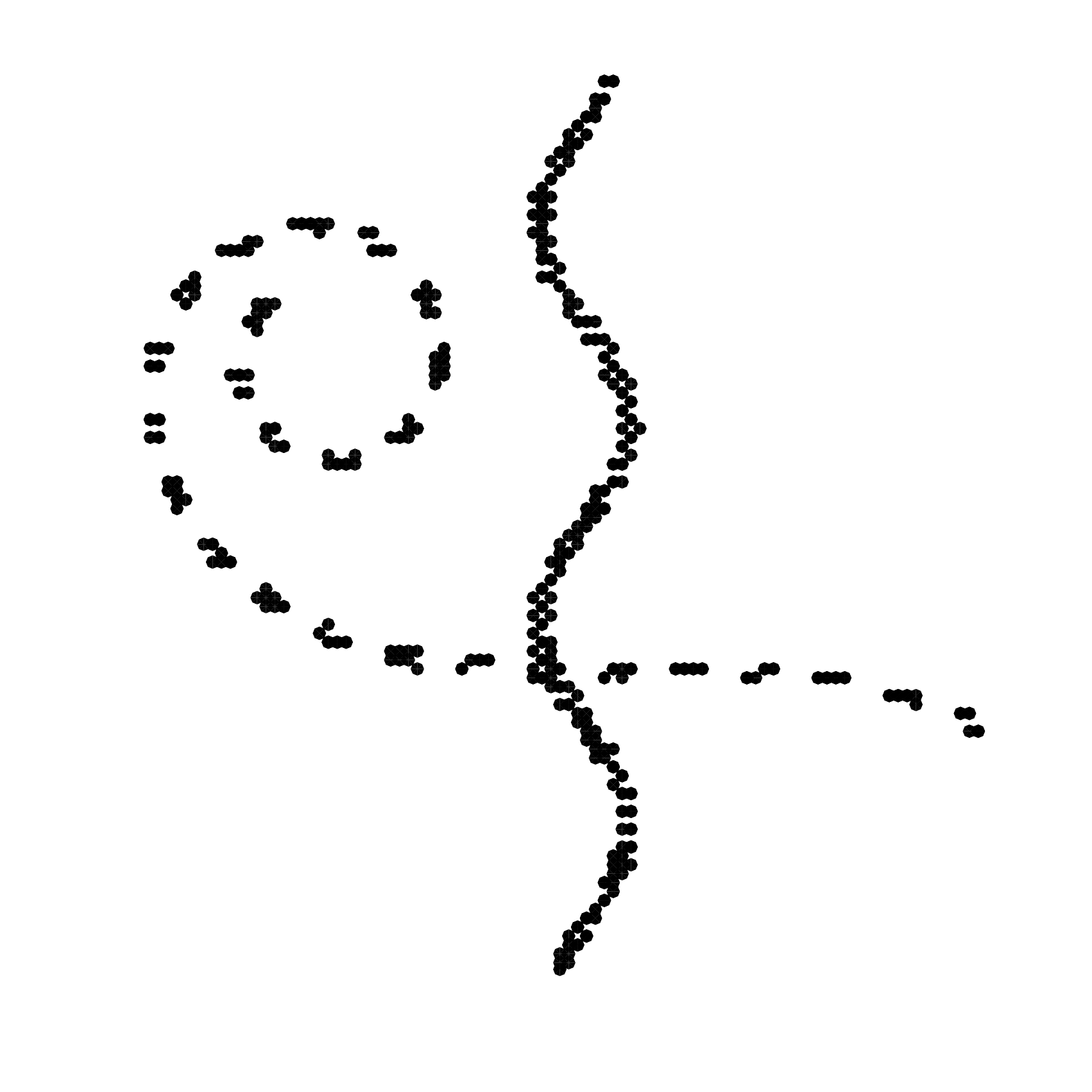} \qquad
			 	\includegraphics[width=\sizp,height=\sizp]{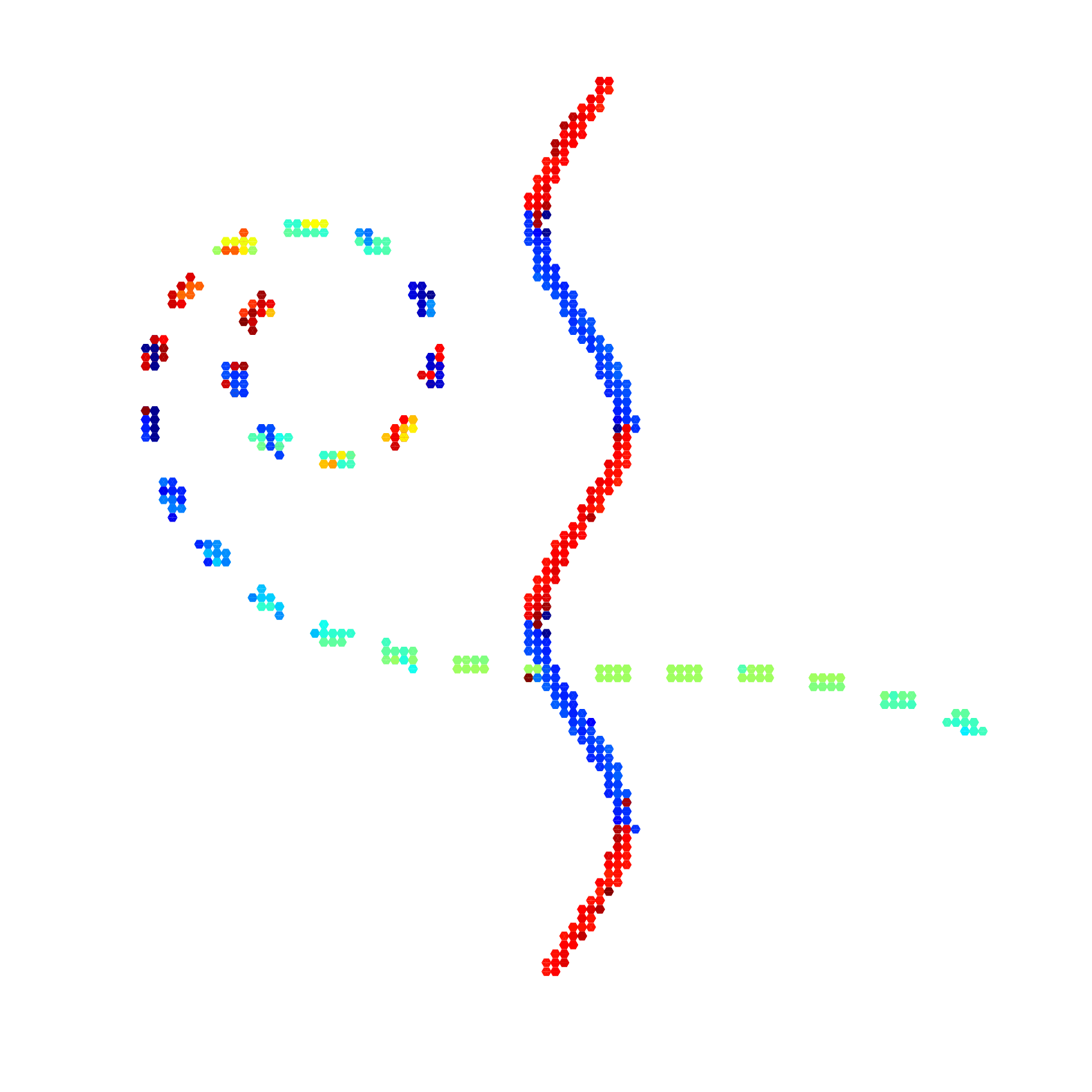} \qquad
	  			\includegraphics[width=\sizp,height=\sizp]{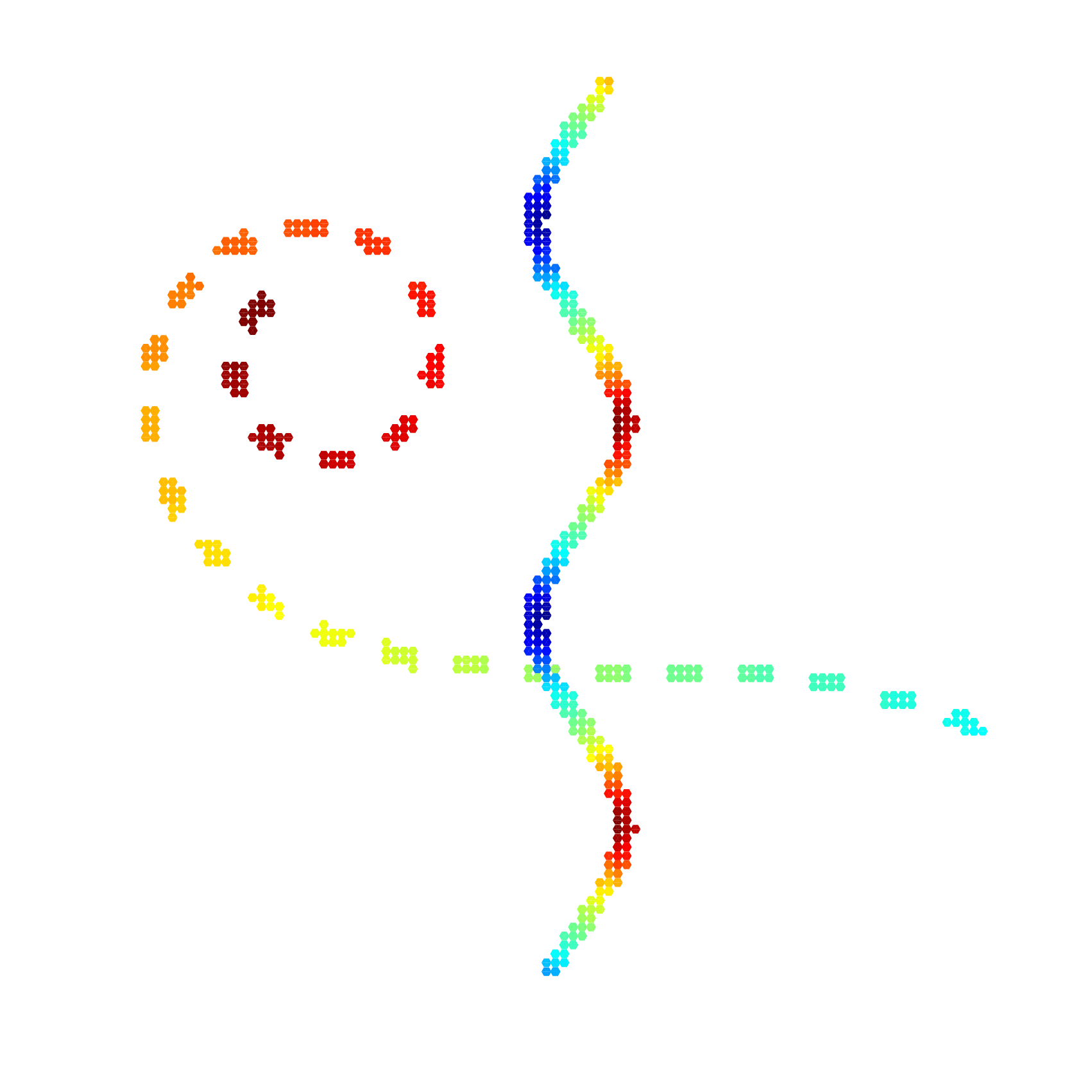}\qquad
	  			\includegraphics[width=\sizp,height=\sizp]{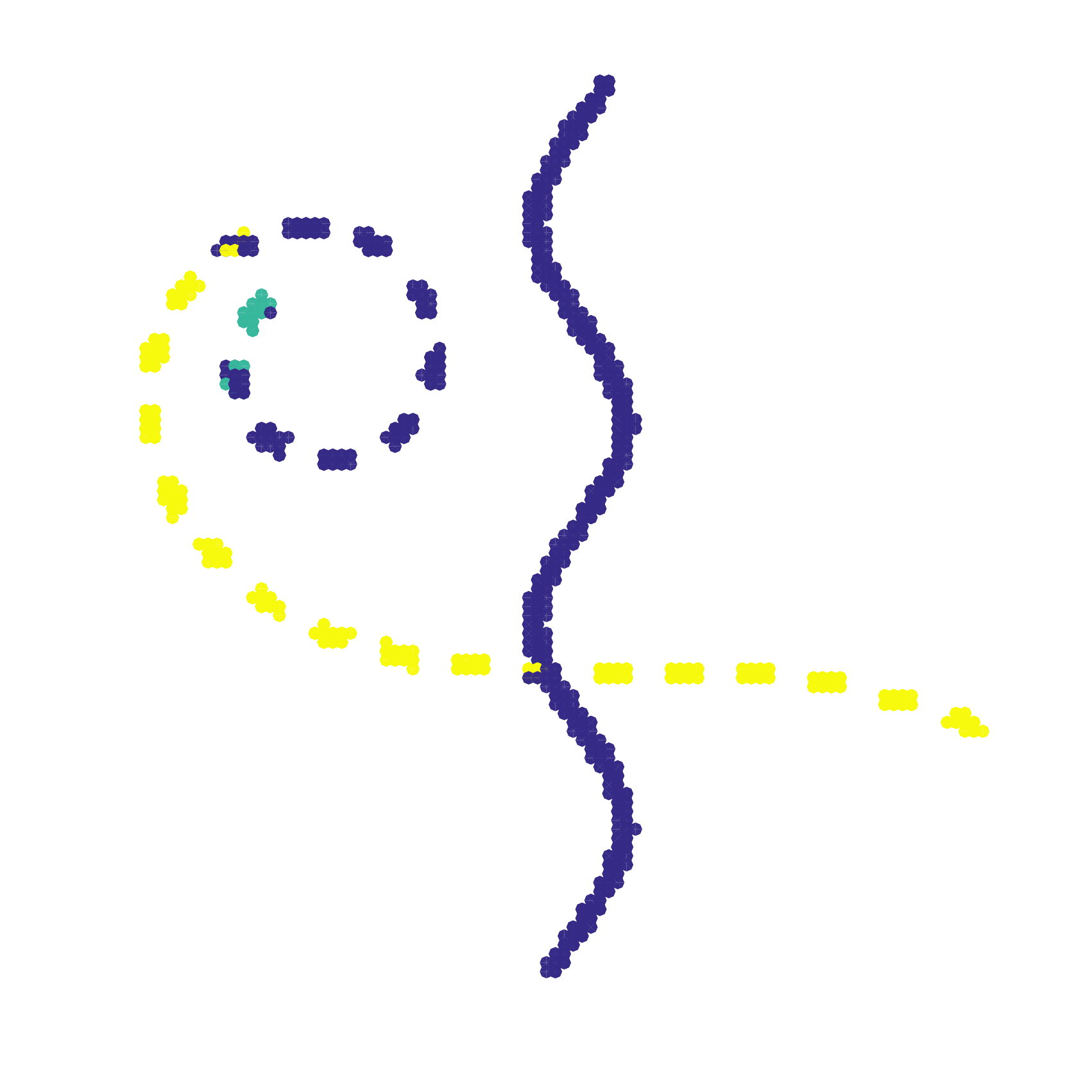} \qquad
	  			\includegraphics[width=\sizp,height=\sizp]{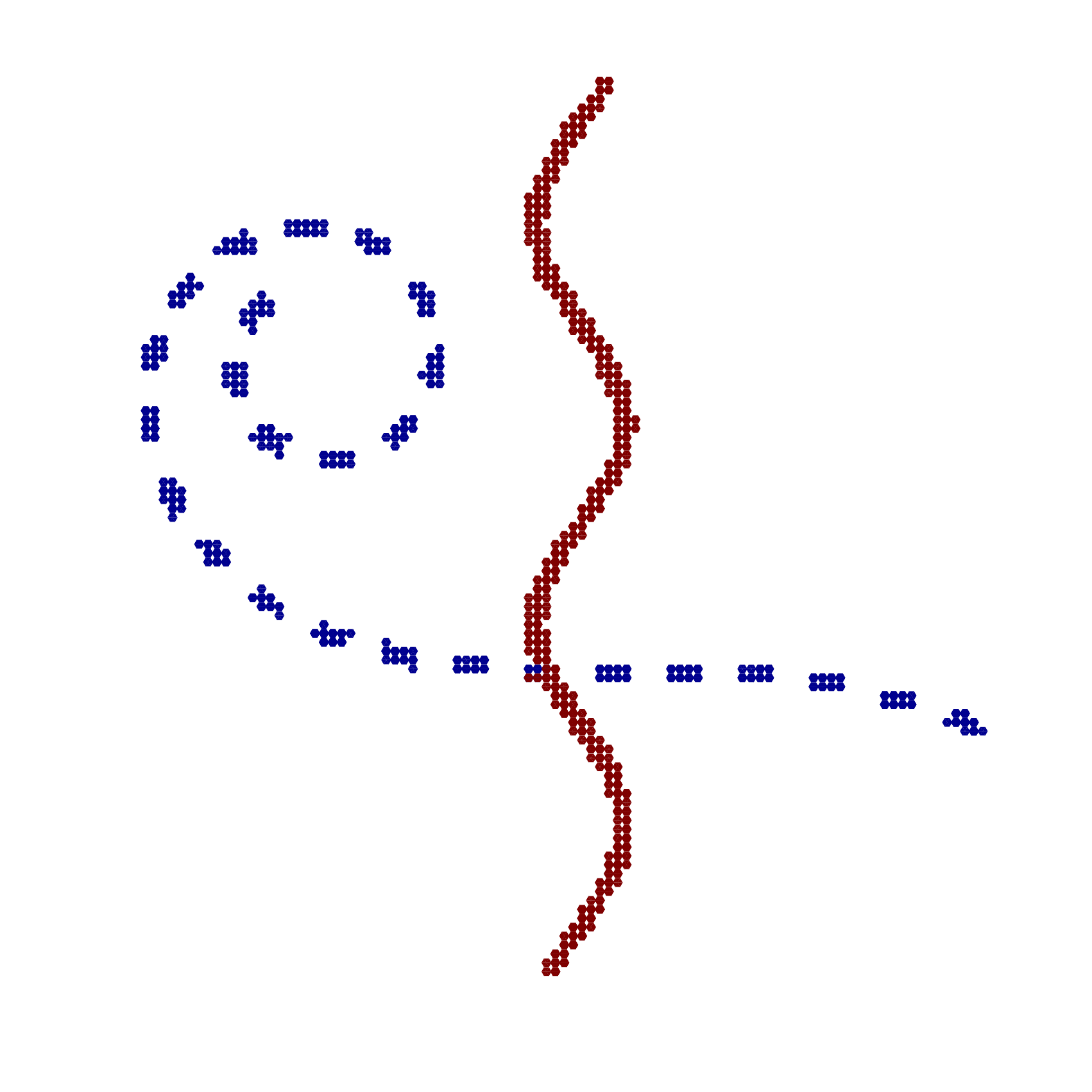} 
	\label{fig:phantomResultsA1}
	\end{subfigure}
	\begin{subfigure}[b]{6.8in}
	\centering
 			 \makebox[0pt][r]{\makebox[15pt]{\raisebox{25pt}{\rotatebox[origin=c]{0}{B1}}}}  \qquad
 			 \includegraphics[width=\sizp,height=\sizp]{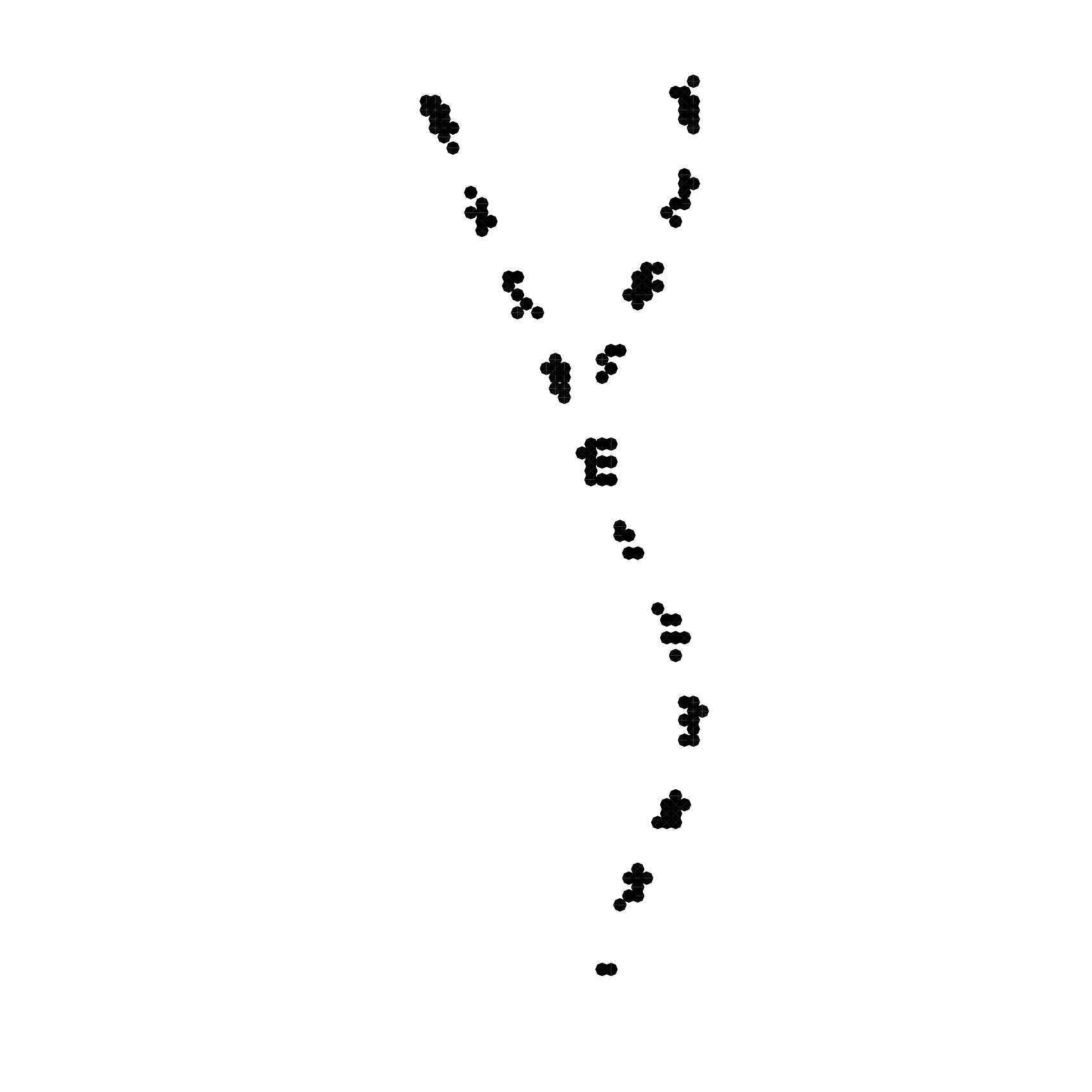} \qquad
			 	\includegraphics[width=\sizp,height=\sizp]{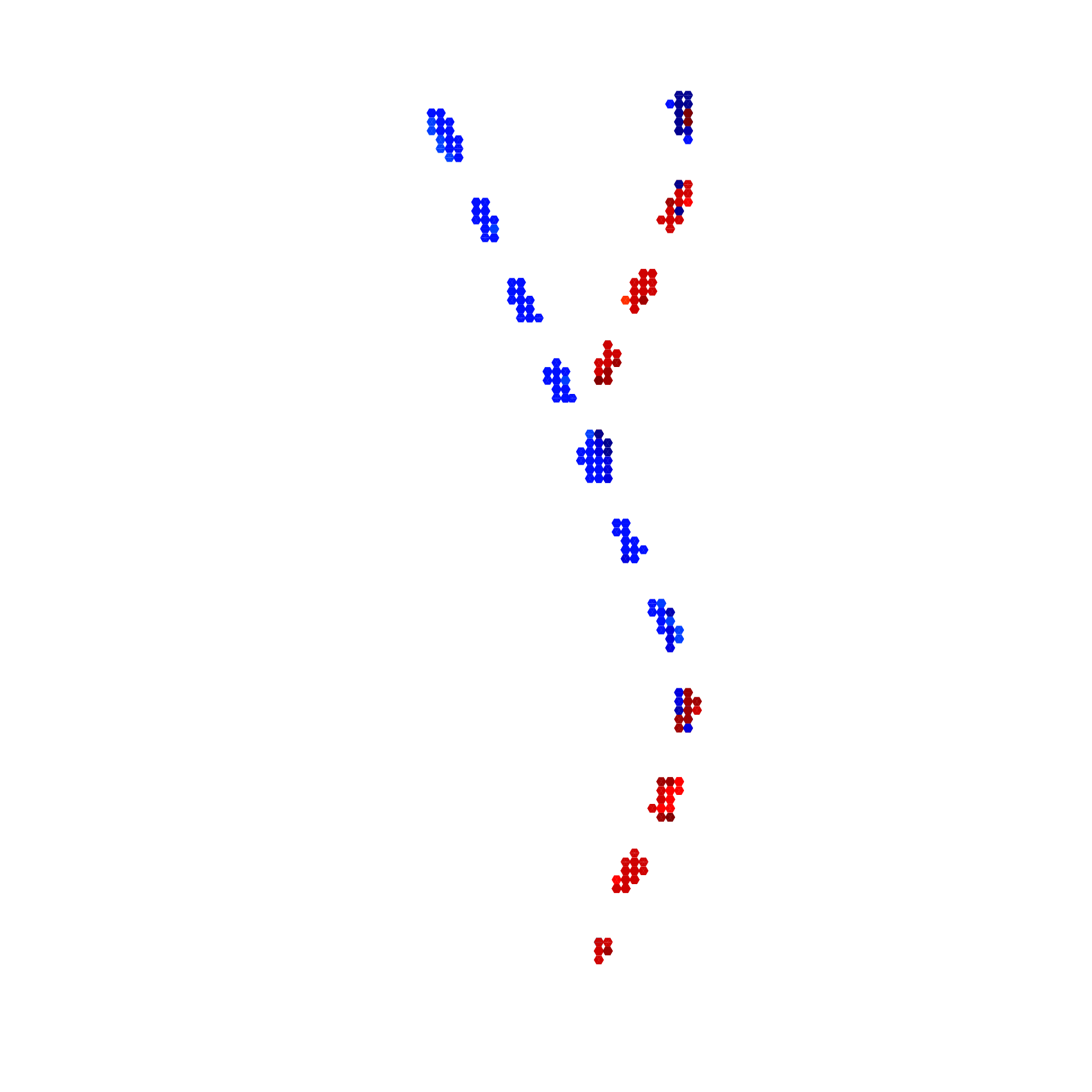} \qquad
	  			\includegraphics[width=\sizp,height=\sizp]{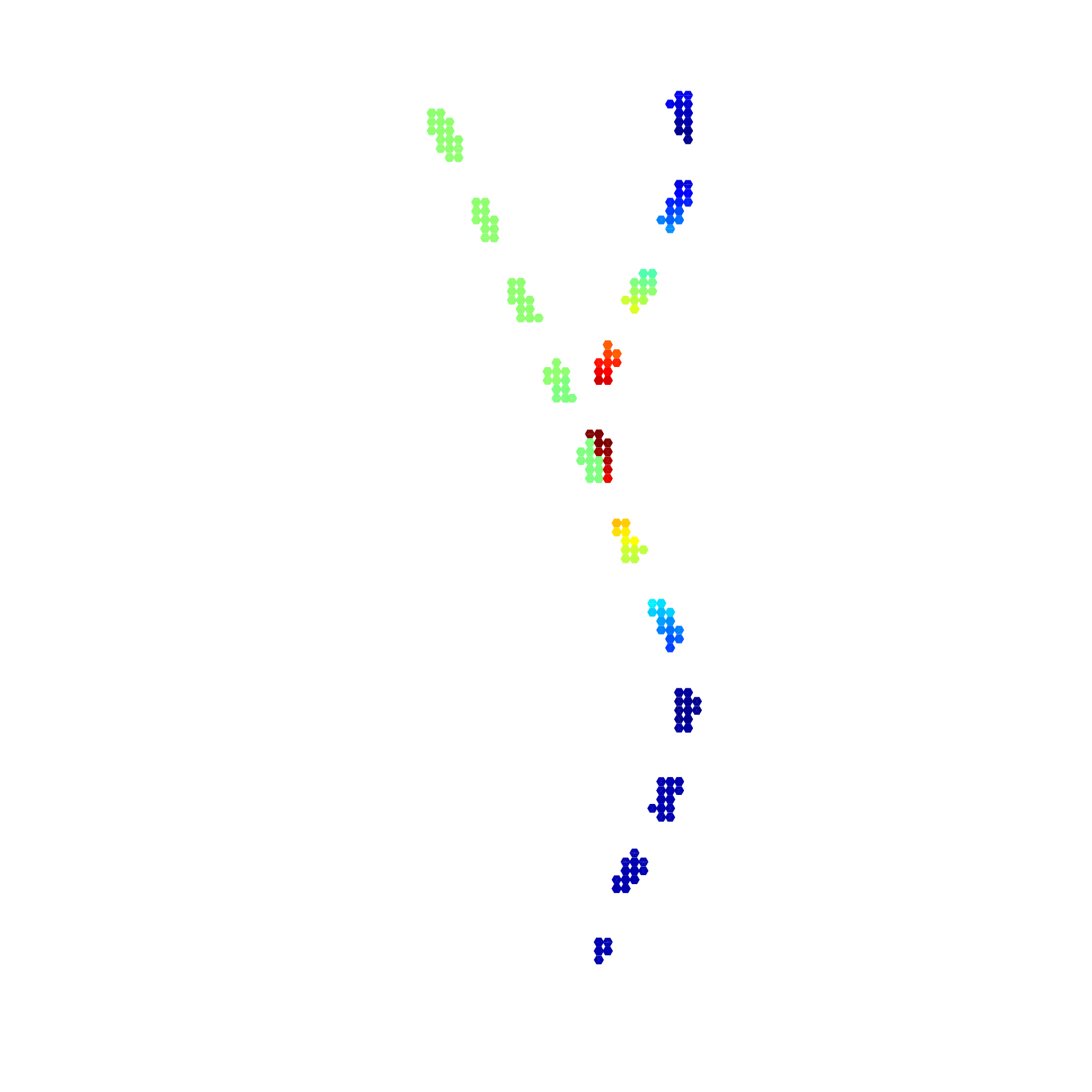}\qquad
	  			\includegraphics[width=\sizp,height=\sizp]{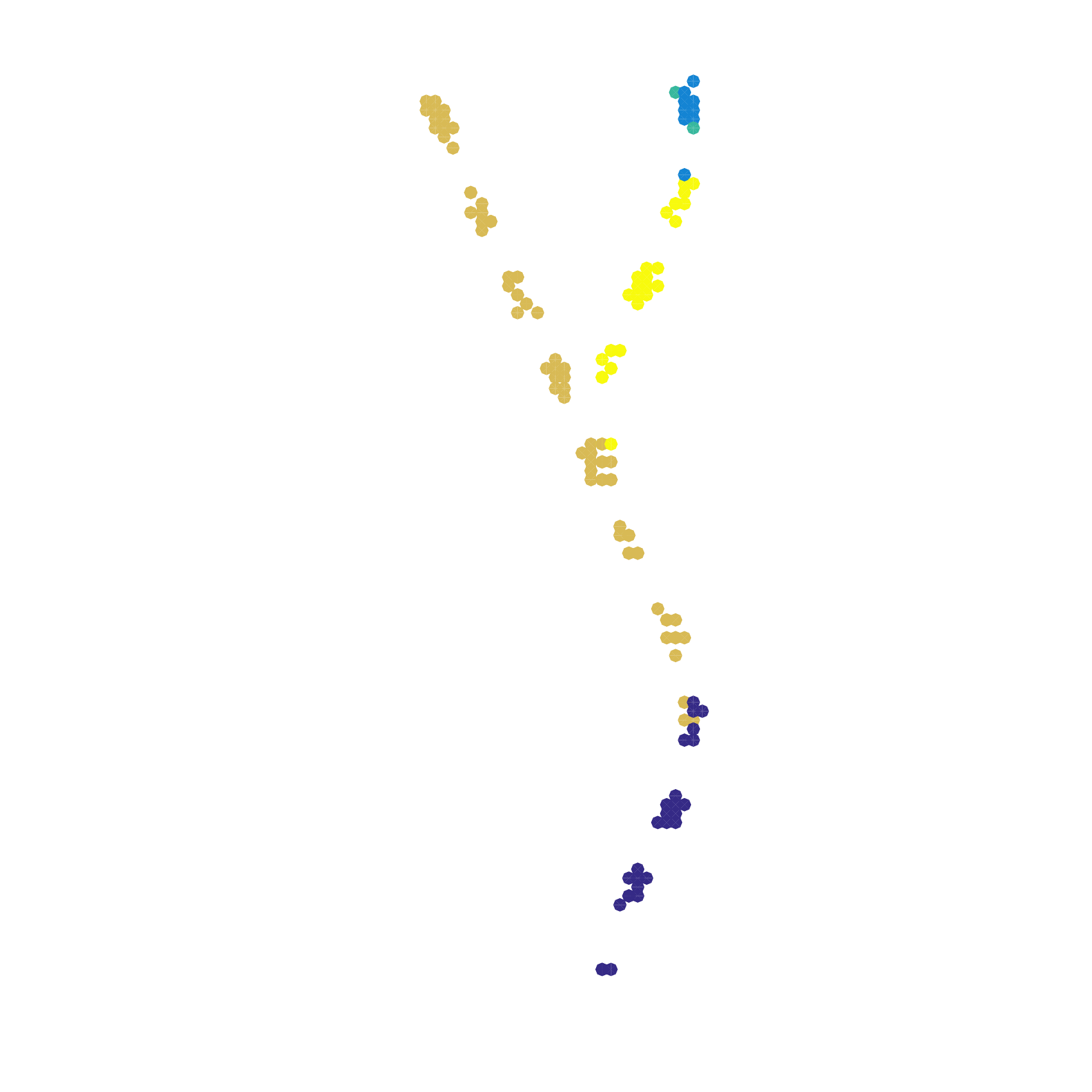} \qquad
	  			\includegraphics[width=\sizp,height=\sizp]{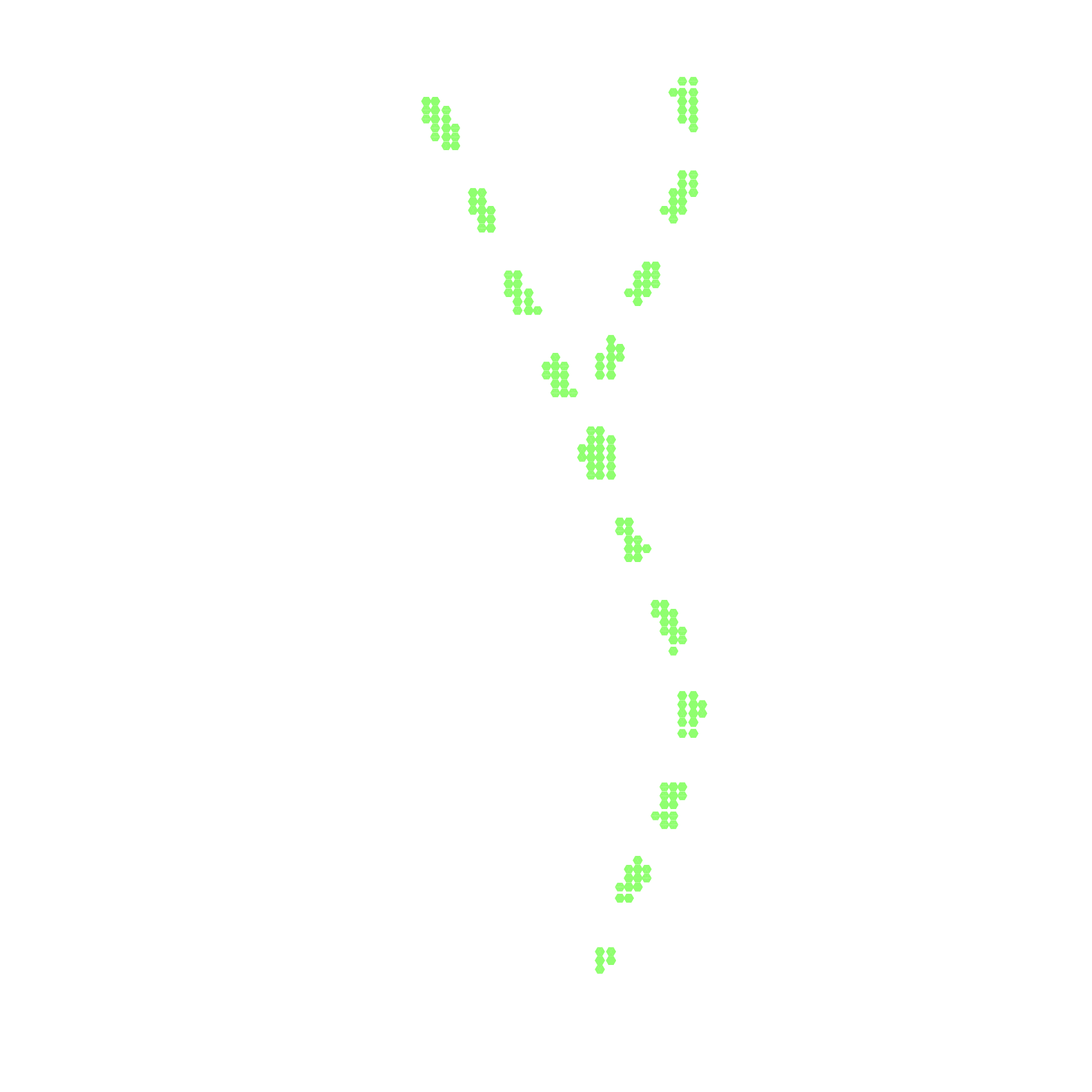} 
	\end{subfigure}
	\begin{subfigure}[b]{6.8in}
	\centering
 			  \makebox[0pt][r]{\makebox[15pt]{\raisebox{25pt}{\rotatebox[origin=c]{0}{C1}}}}  \qquad
				\includegraphics[width=\sizp,height=\sizp]{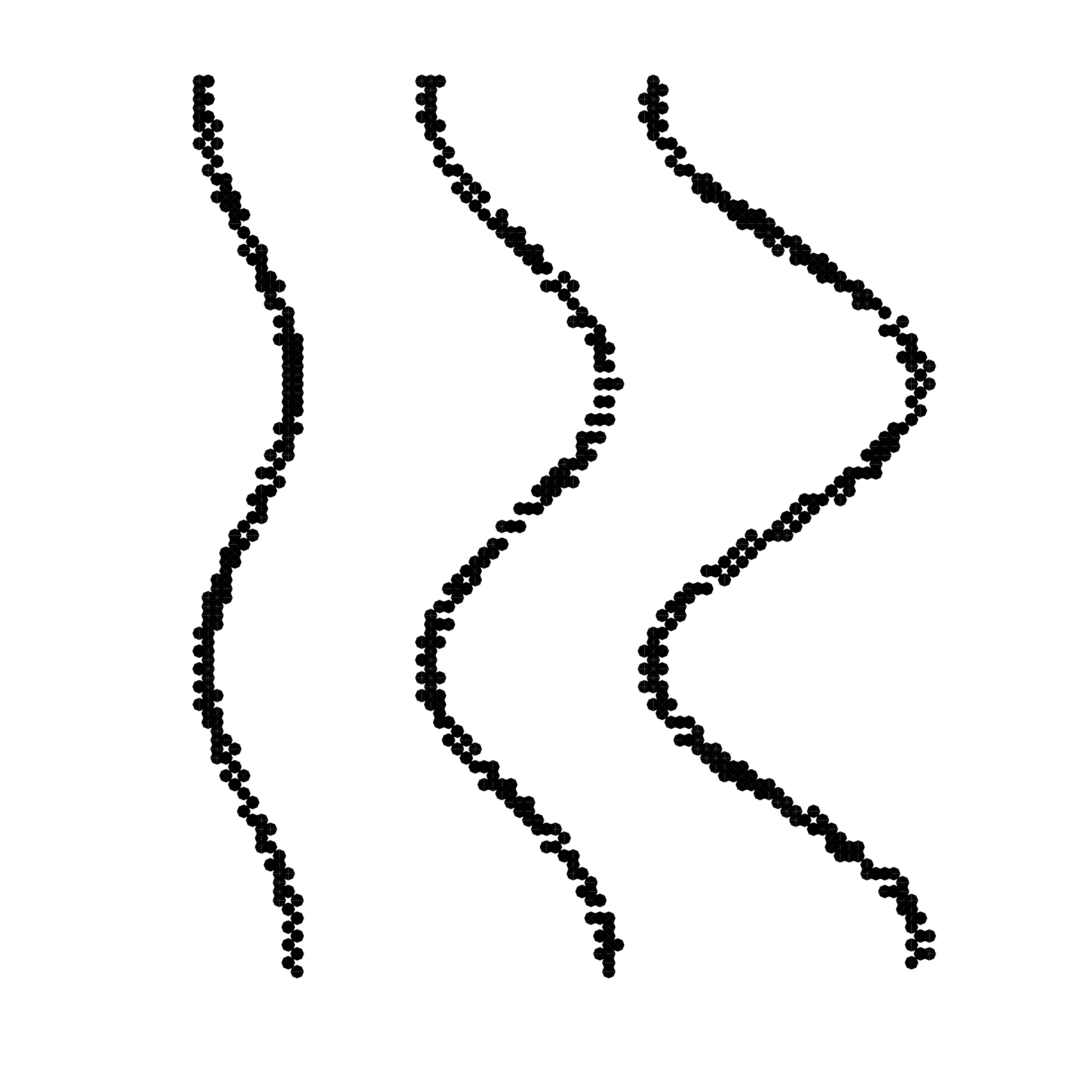} \qquad
			 	\includegraphics[width=\sizp,height=\sizp]{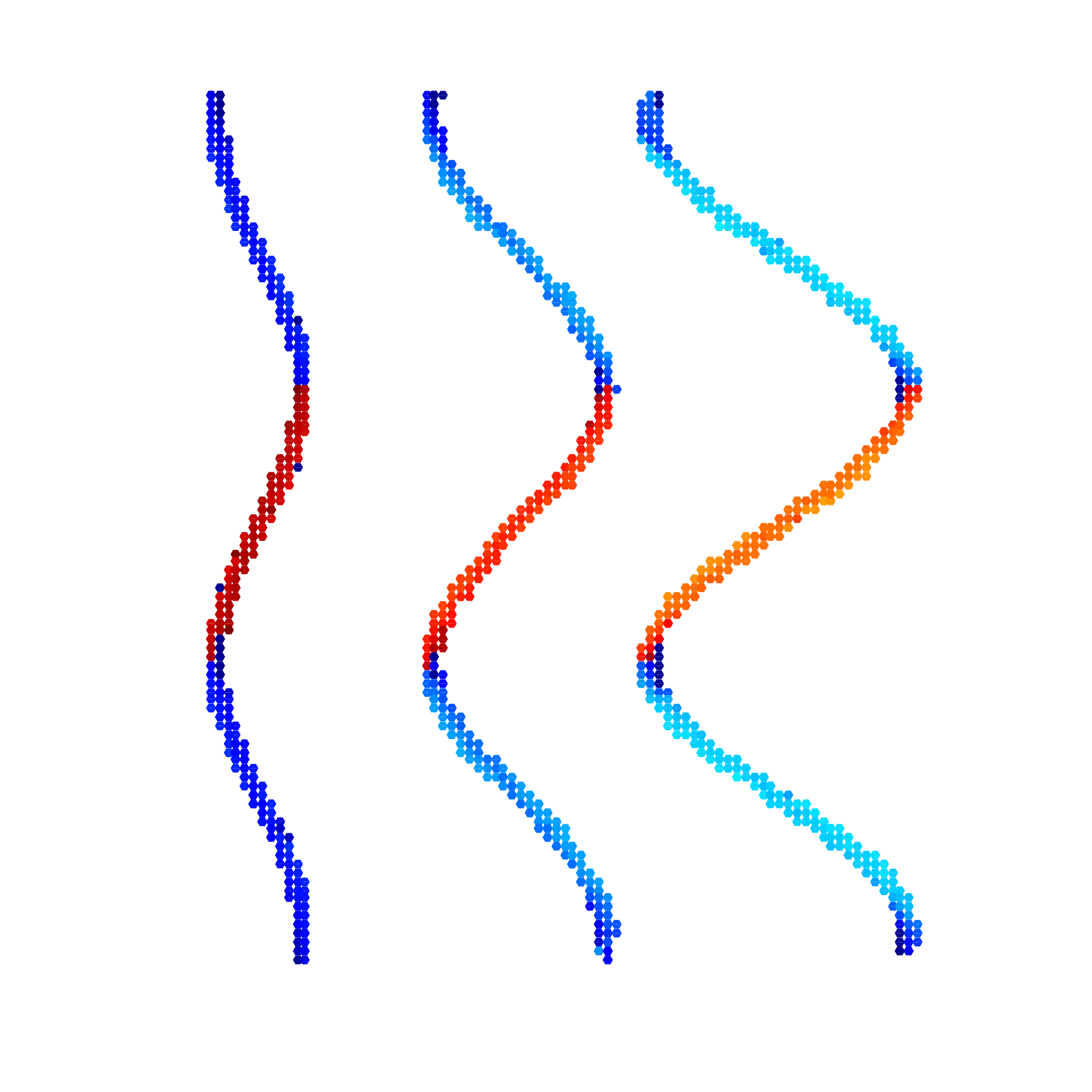} \qquad
	  			\includegraphics[width=\sizp,height=\sizp]{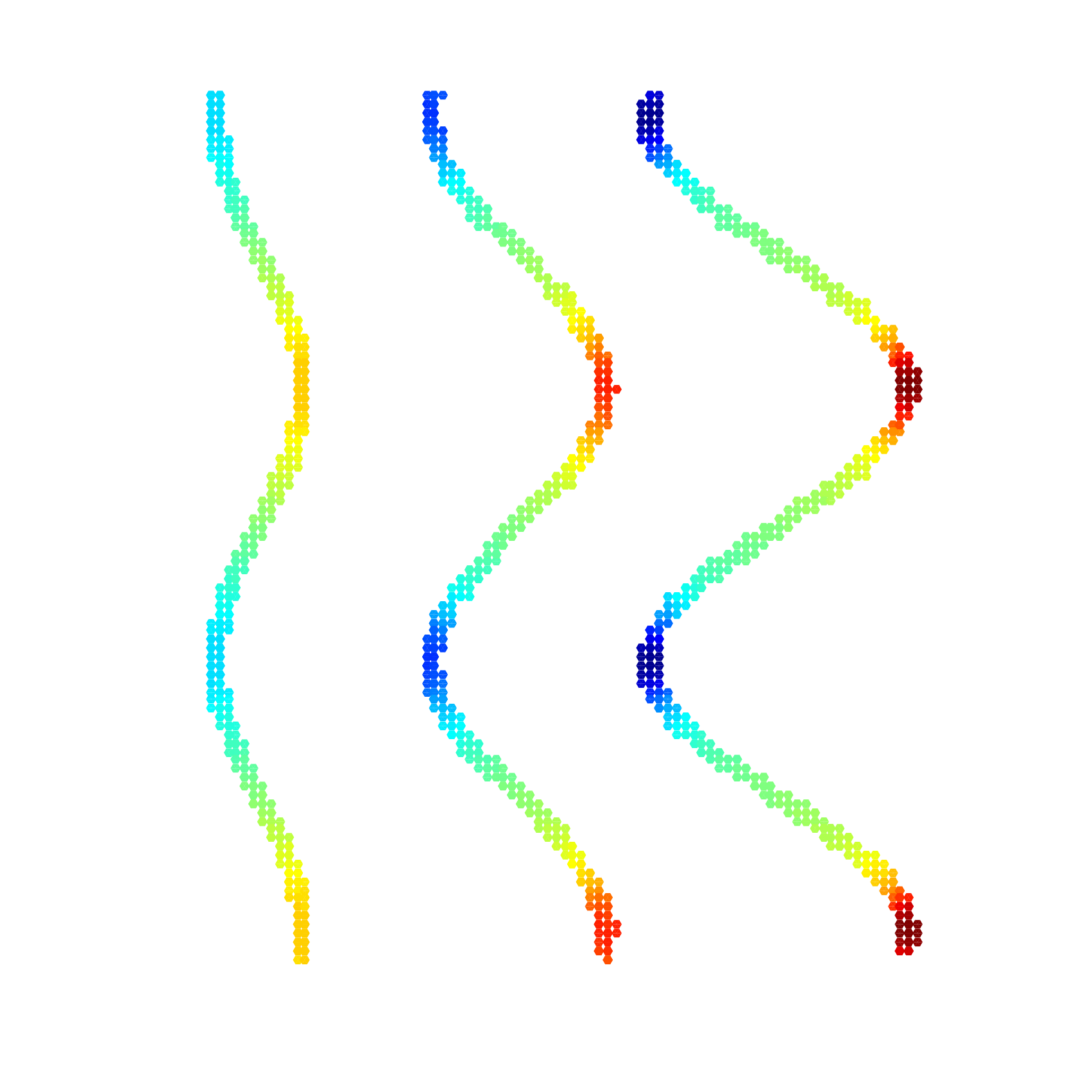}\qquad
	  			\includegraphics[width=\sizp,height=\sizp]{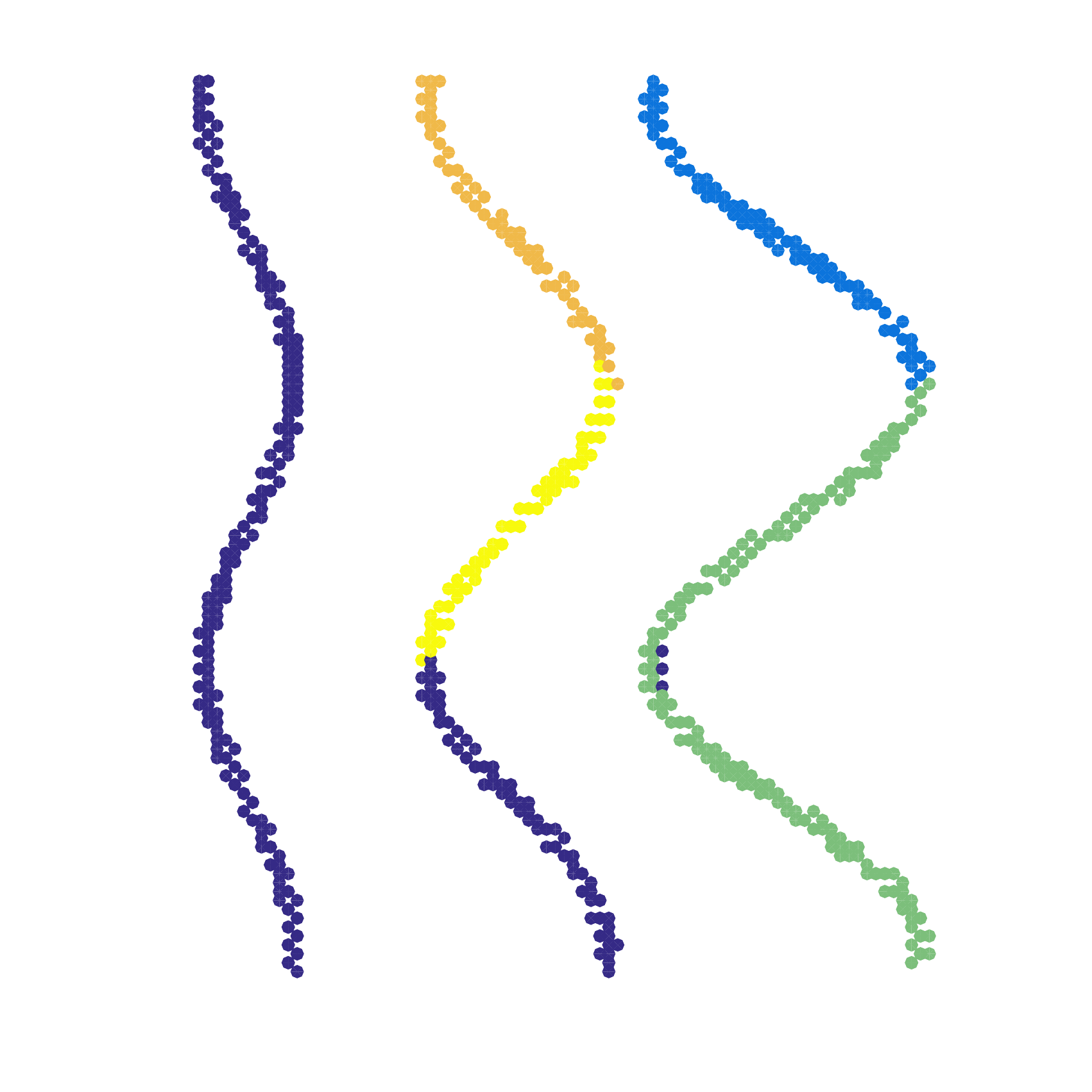} \qquad
	  			\includegraphics[width=\sizp,height=\sizp]{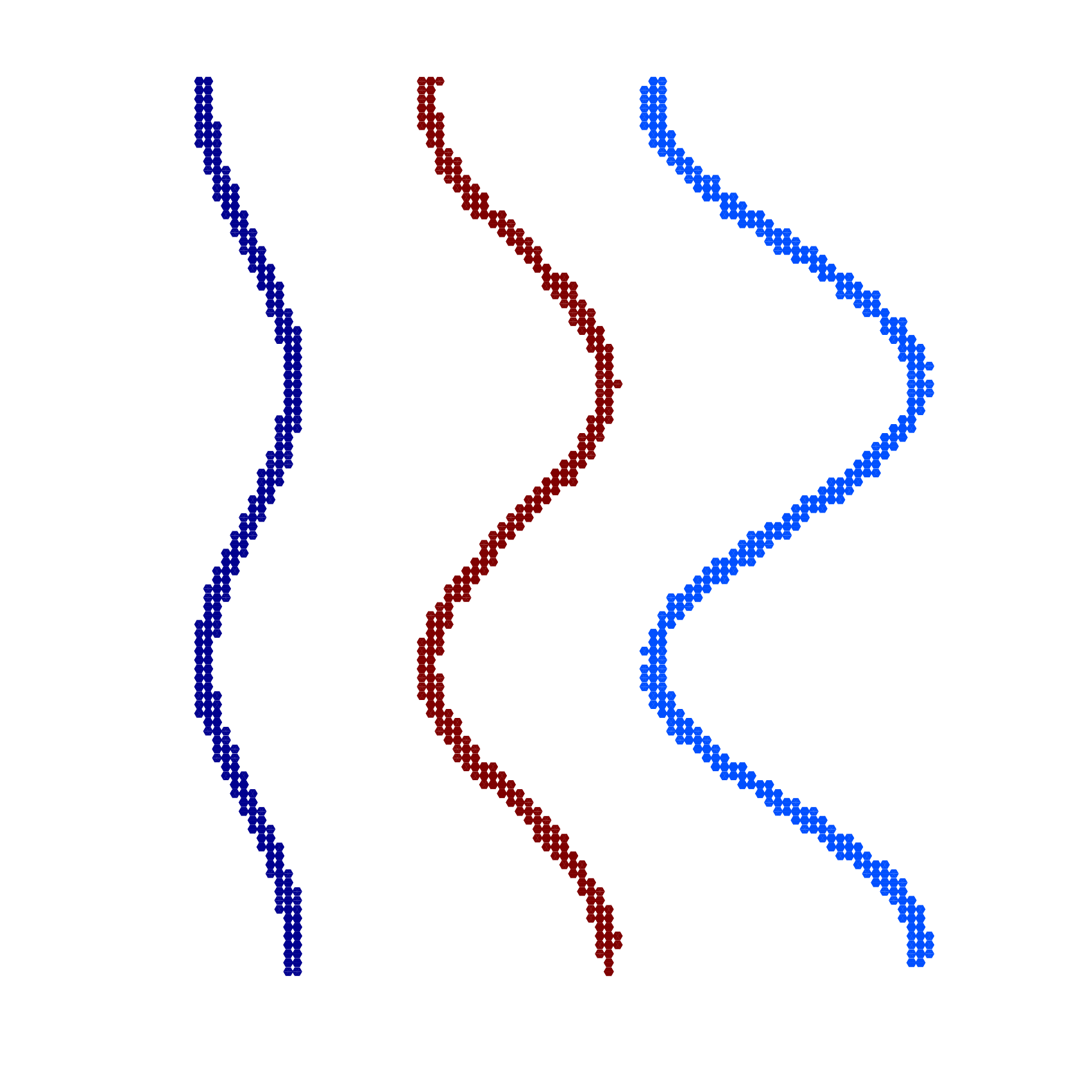} 
	\end{subfigure}
	\begin{subfigure}[b]{6.8in}
	\centering
 	 \makebox[0pt][r]{\makebox[15pt]{\raisebox{25pt}{\rotatebox[origin=c]{0}{D1}}}}  \qquad
 	 	\includegraphics[width=\sizp,height=\sizp]{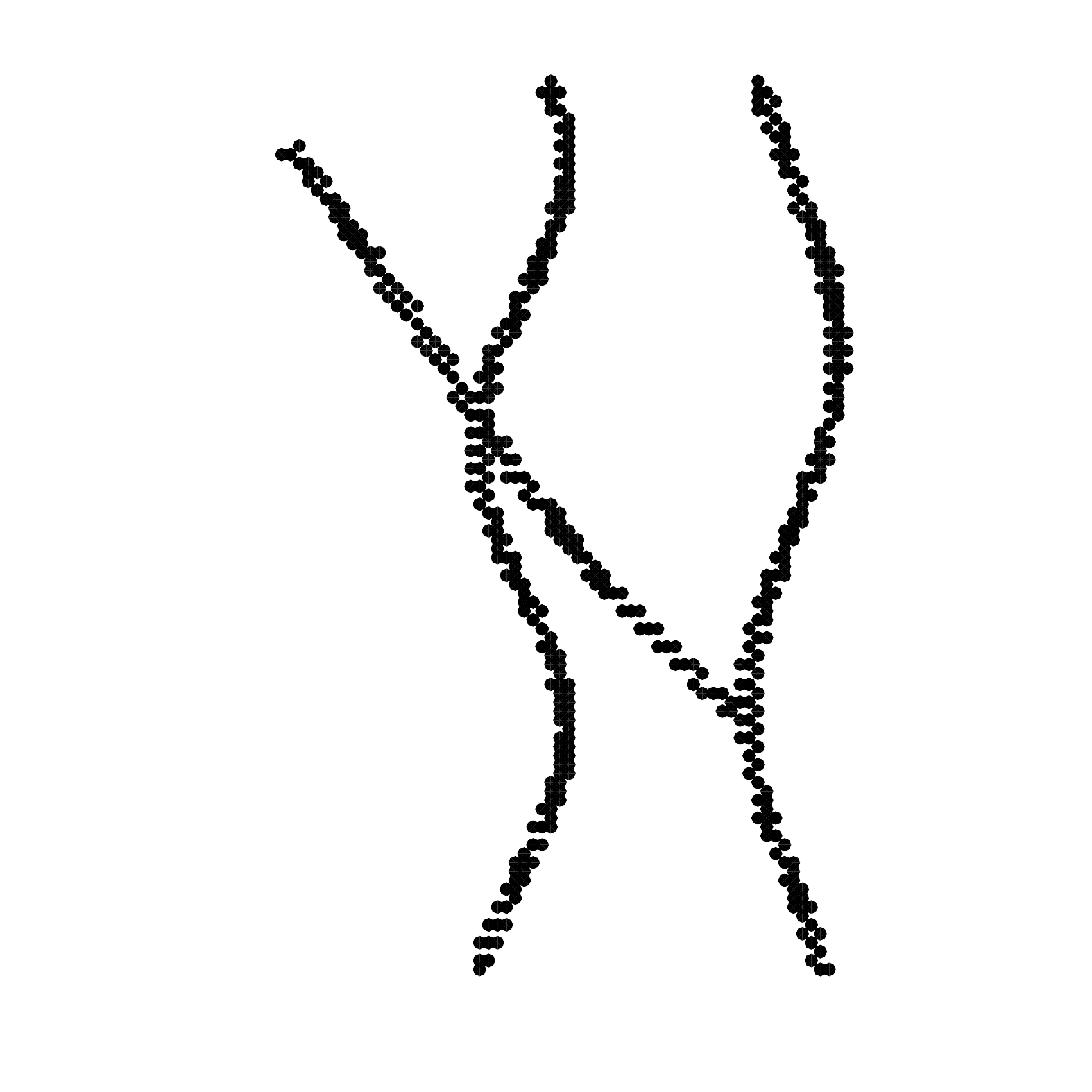} \qquad
			 	\includegraphics[width=\sizp,height=\sizp]{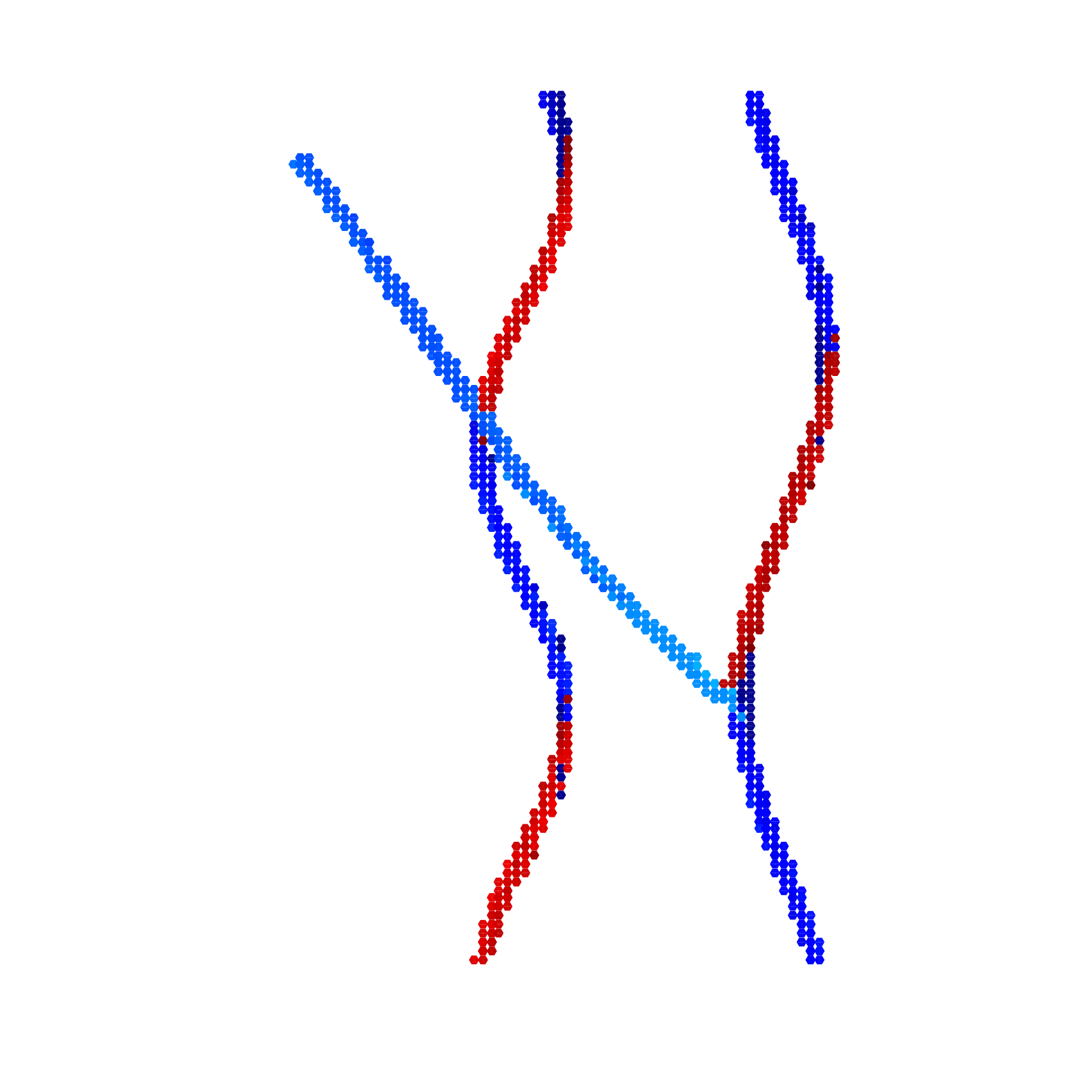} \qquad
	  			\includegraphics[width=\sizp,height=\sizp]{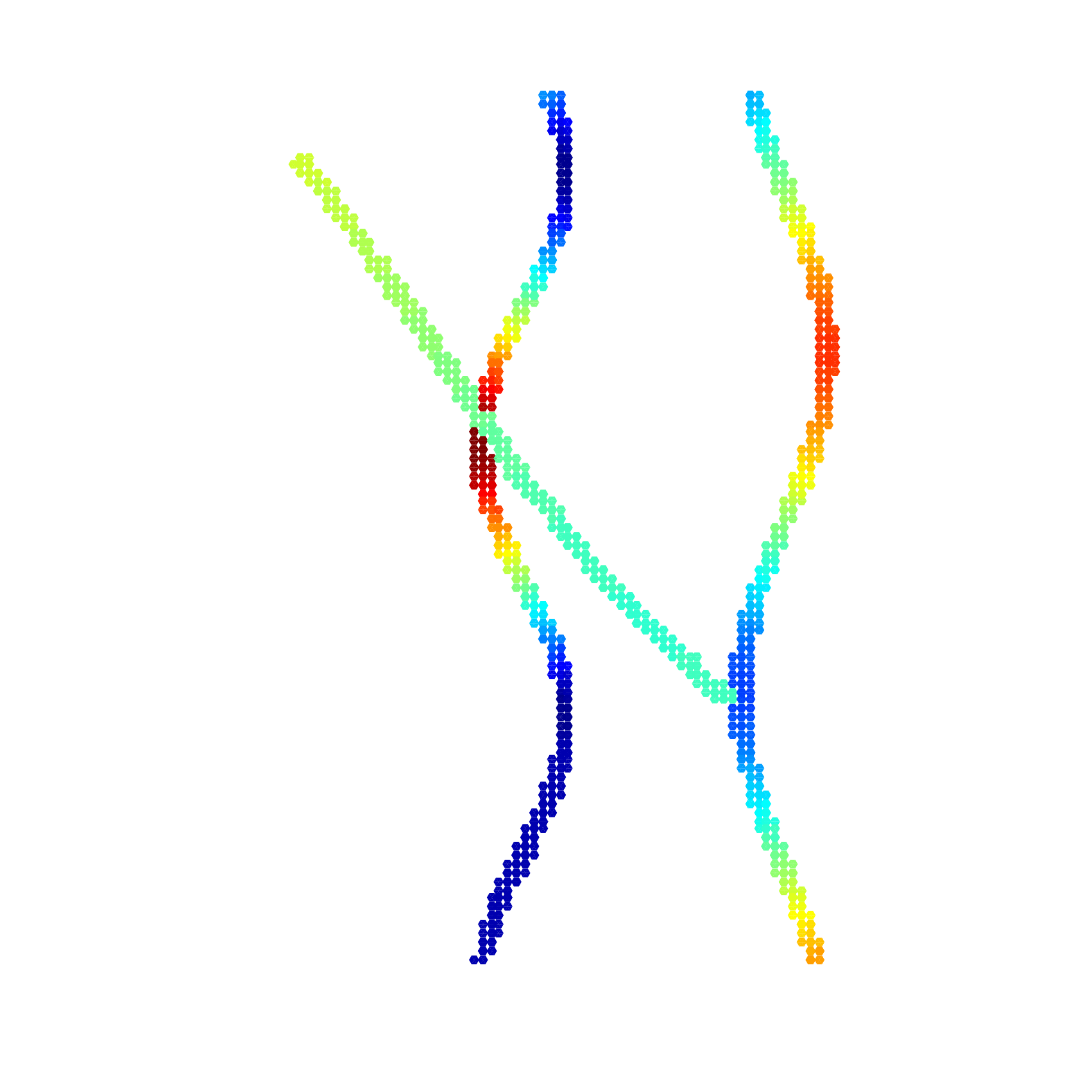}\qquad
	  			\includegraphics[width=\sizp,height=\sizp]{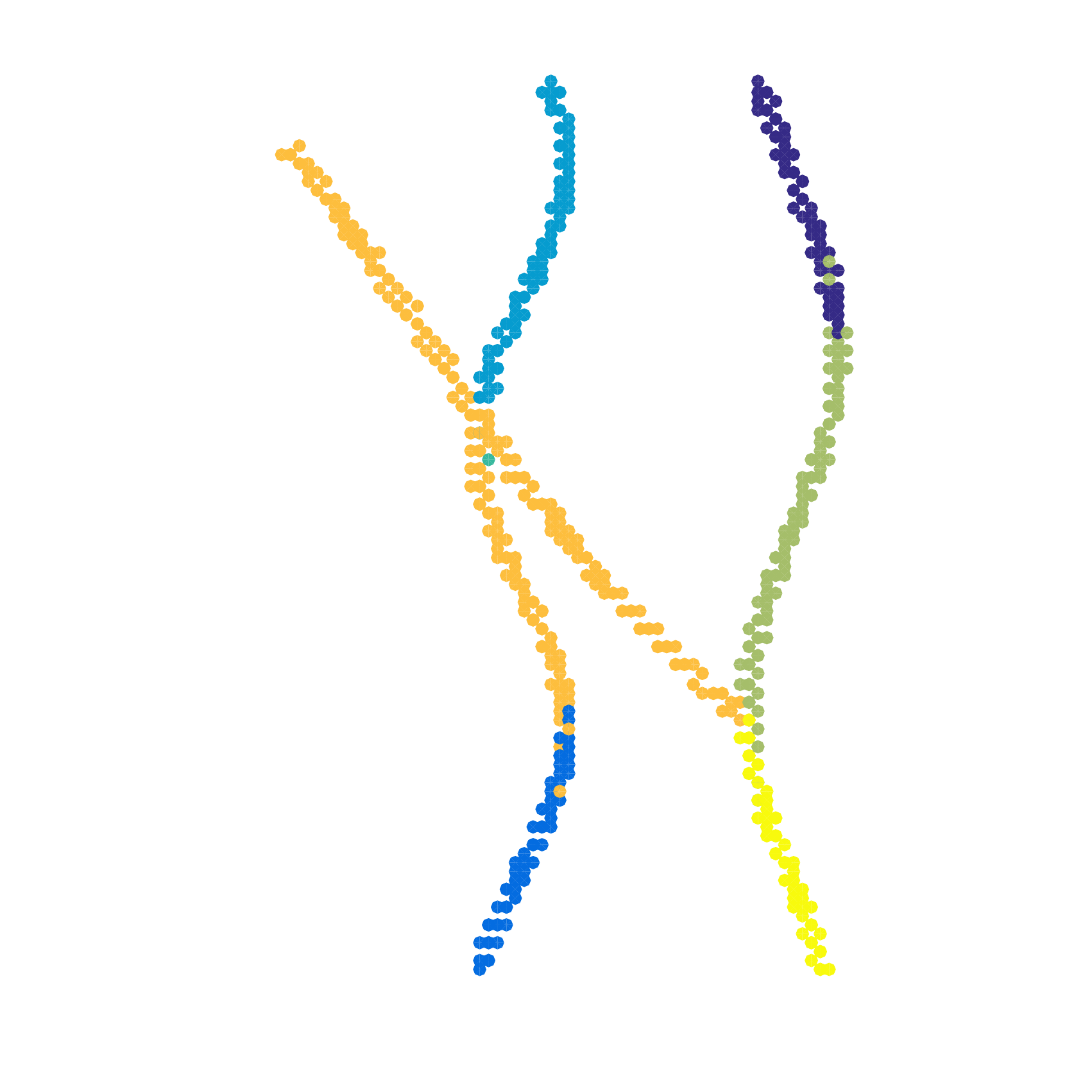} \qquad
	  			\includegraphics[width=\sizp,height=\sizp]{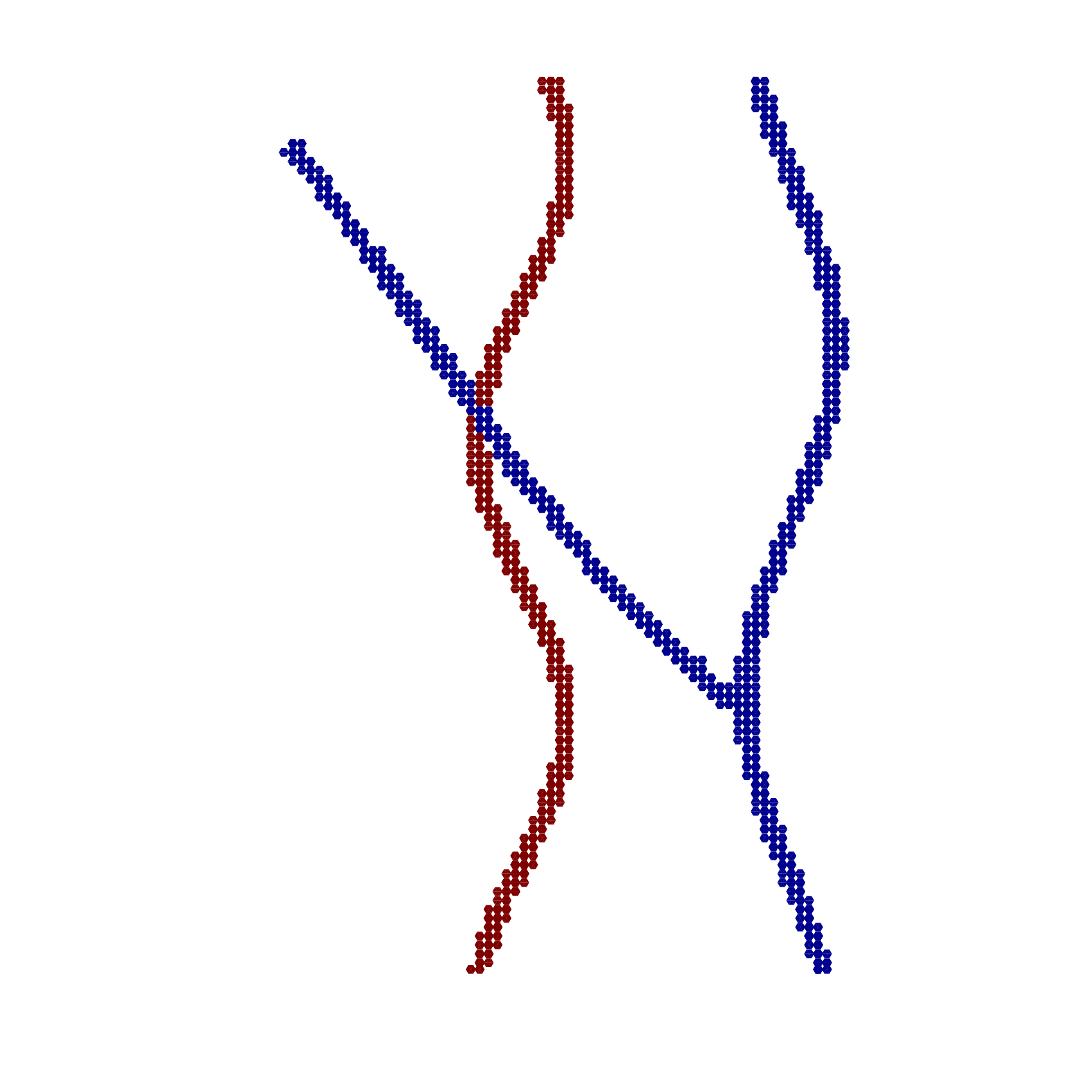} 

	\end{subfigure}
	\begin{subfigure}[b]{6.8in}
	\centering
 			 \makebox[0pt][r]{\makebox[15pt]{\raisebox{25pt}{\rotatebox[origin=c]{0}{E1}}}}   \qquad
 			 	\includegraphics[width=\sizp,height=\sizp]{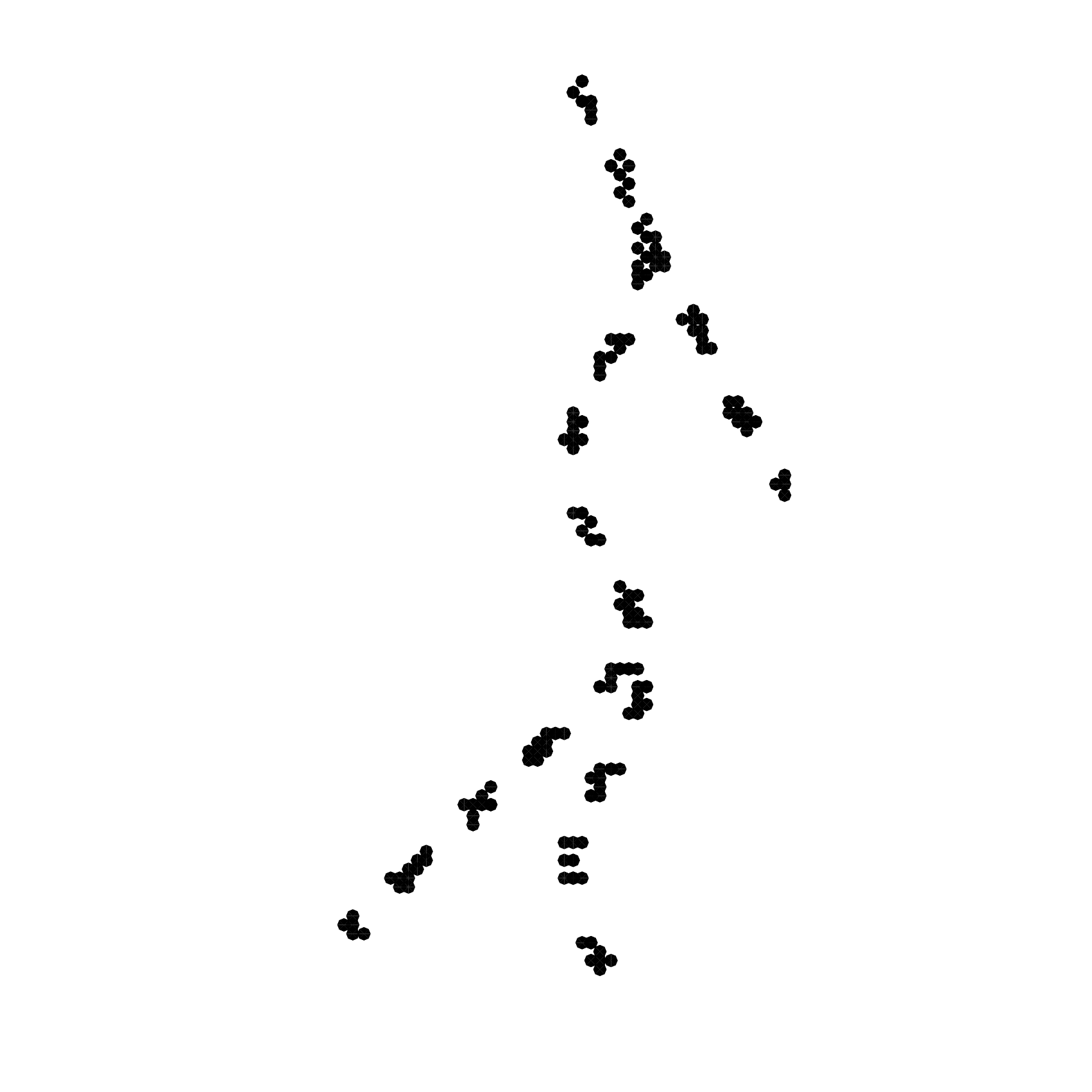} \qquad
			 	\includegraphics[width=\sizp,height=\sizp]{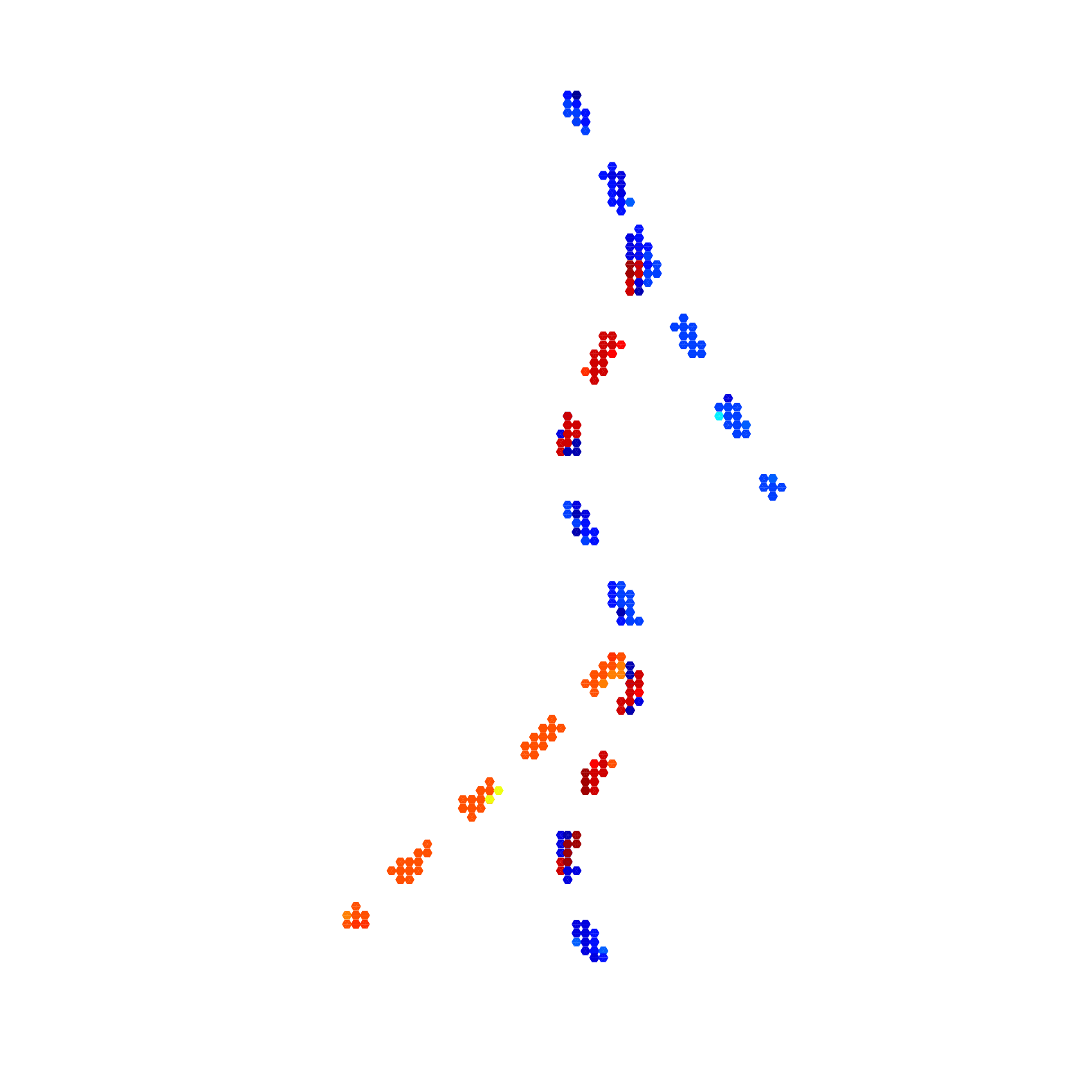} \qquad
	  			\includegraphics[width=\sizp,height=\sizp]{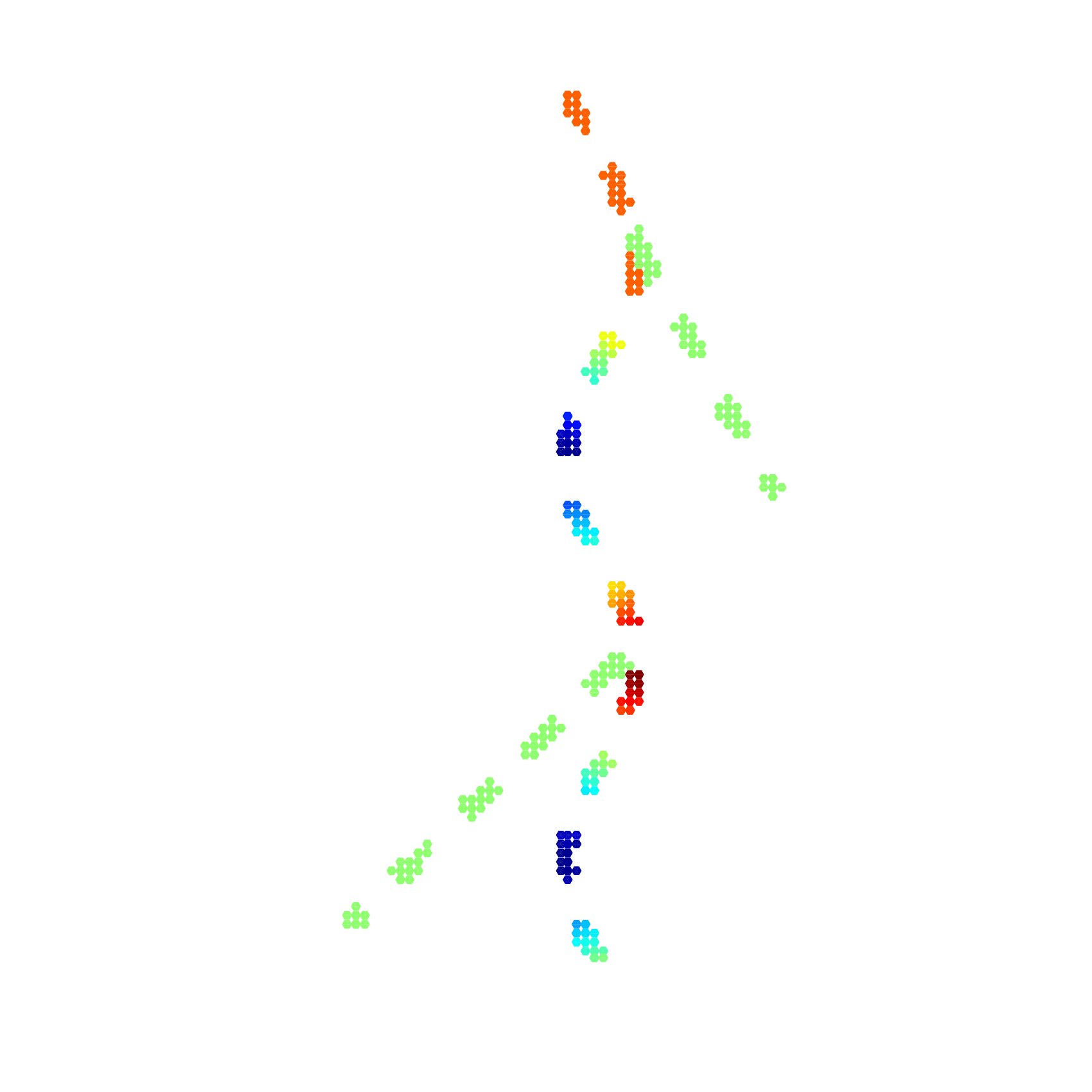}\qquad
	  			\includegraphics[width=\sizp,height=\sizp]{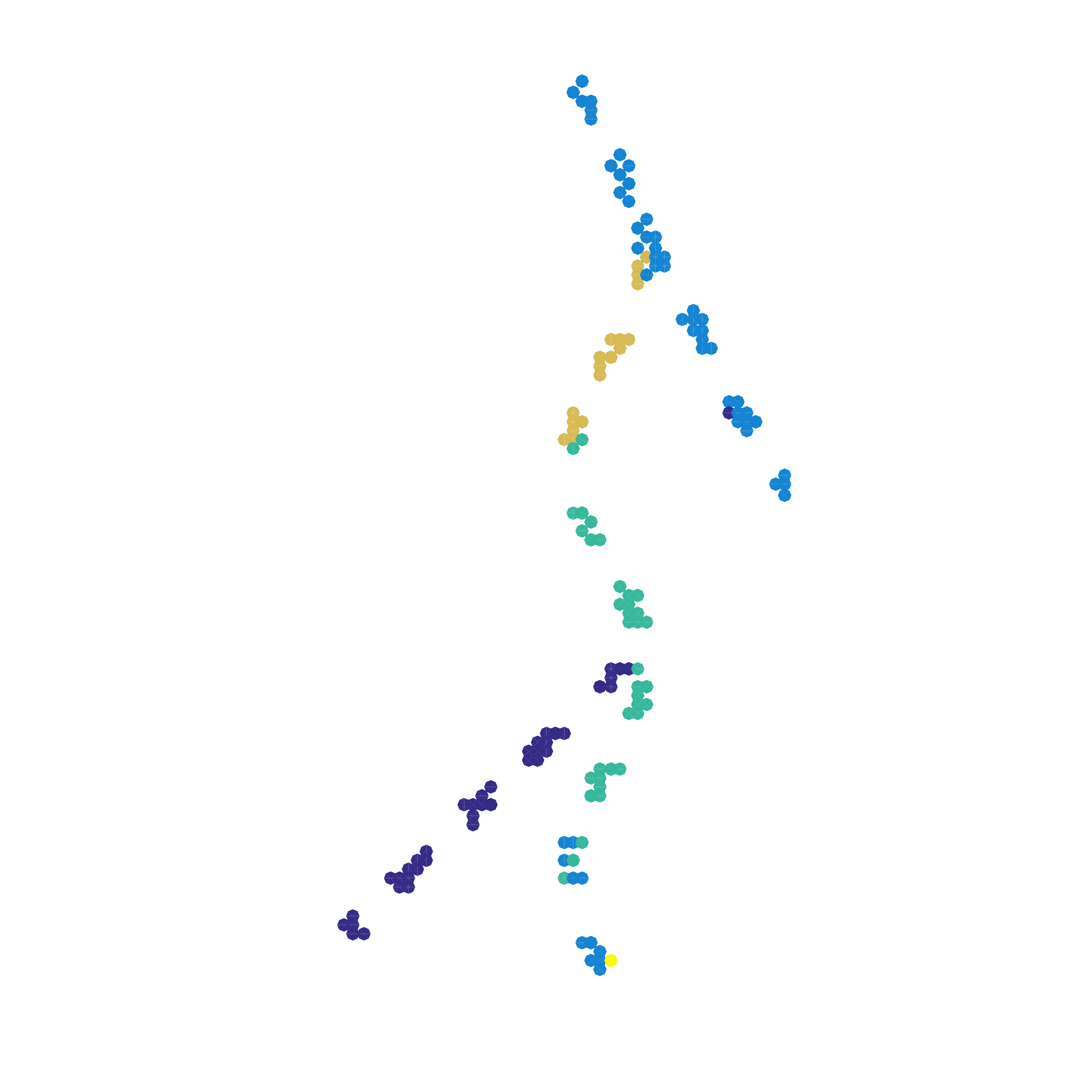} \qquad
	  			\includegraphics[width=\sizp,height=\sizp]{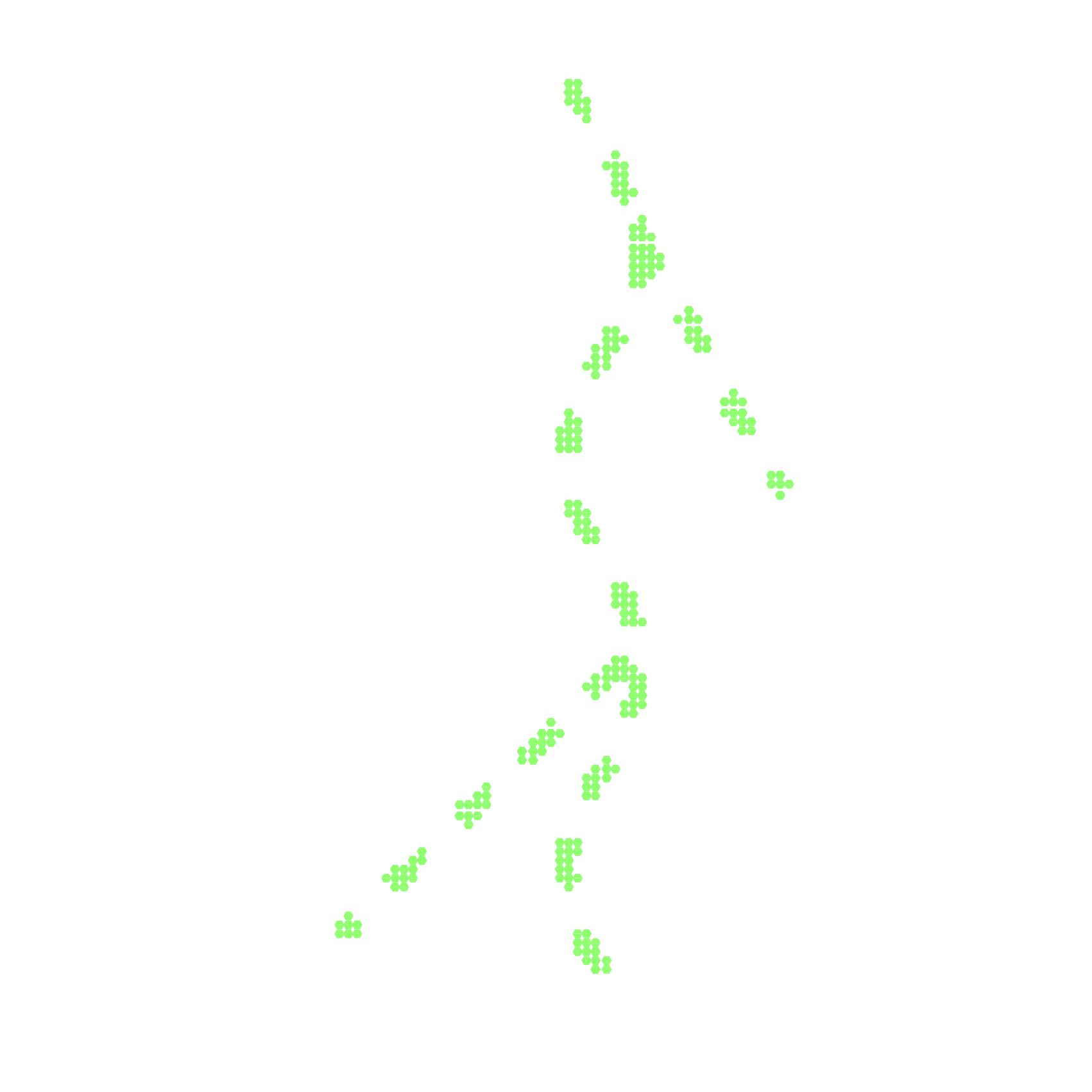} 

	\end{subfigure}
\caption{Samples of phantom images in different categories. From left to right, the images in each category represent:
stimulus, the orientation map, the curvature map and the clustering result with the previous~\cite{favali2016analysis} and the new kernel. The color of the curvature maps are scaled between the maximum and minimum values of the curvature in each image. }
  \label{fig:phantomResults}
\end{figure*}

To validate the method using the phantom images, the method presented in Algorithm~\ref{alg:ConnectAnal} is directly applied on the lifted images, starting from Step \ref{st:fundSol}. The orientation and curvature values of these images are available from the beginning. The intensity is constant for these images, so the intensity term in (\ref{eq:connectKernel}) is equal to 1. The last two columns of Fig.~\ref{fig:phantomResults} represent two different clustering results per image. For each image, two kernels have been used for obtaining the affinity matrix: the new kernel (adaptive, based on the curvature at each point); and the kernel used in our previous work~\cite{favali2016analysis}. This helps in highlighting the importance of including additional contextual information for connectivity analysis. Each color in the final results represents one detected unit. 
The parameter $\sigma_\kappa$ used in these simulations is 0.01 for the 4D kernel \cite{favali2016analysis} and 0.001 for the 5D one. As seen in this figure, the new method is capable of grouping the elongated, rotated and curved structures despite disconnections, high curvature points or small crossing angles. It not only differentiates well between curved crossing structures, but also groups the bifurcations within the main parent structure so that they construct one unique unit. For other examples we refer to Fig. 1 in the supplementary material.

\subsubsection{Retinal patches}
To validate the method on retinal image patches, after creating the patches in the lifted domain (\ref{eq:CrpLiftImg}), we perform the analysis starting from Step~\ref{st:fundSol} in Algorithm~\ref{alg:ConnectAnal}. All 272 patches have been examined semi-automatically. The number of patches processed per group is presented in the second column of Table~\ref{tbl:params}. 
In order to see if each detected unit represent one individual vessel, we create the corresponding 2D patch for each image from the artery-vein labeled images, which are available for this dataset. Existing vessels in each artery-vein labeled patch are compared with detected clusters. In case two vessels with similar labels in the ground truth image (artery or vein) belong to separate vessel trees (parents), they get a different label.  
 In addition, the operator performs a final check to control the final results. Two criteria are used to evaluate the performance. One is the Correct Detection Rate (CDR\%) defined as the percentages of correctly grouped patches among all examined cases, similar to~\cite{favali2016analysis}. The second criterion is $Q_{clust}$ obtained in the spectral clustering step (explained in \ref{sec:sepctral}). In this method, the best number of clusters is the one which minimizes the defined cost function or maximizes the quality ($0\leq Q_{clust}\leq 1$). This criterion represents how well aligned the elements of each group are. These two performance values have been measured for all the patches and presented in  the last two columns of Table~\ref{tbl:params} for each category separately. 

During the experiments the parameters used in numerical simulations and the calculation time of different parts of the experiments have been recorded. Several parameters are involved in creating the kernel. Some of them are determined automatically based on the available information in the data and others need to be set manually. 
The first set includes the size of the kernel in $x$ and $y$ dimension ($n_x$ and $n_y$ respectively). The other one is the number of discrete curvature values ($n_\kappa$), which for each a 3D section of fundamental solution with $\kappa$ fixed is created. Considering the step size as 1 pixel, $n_x$ and $n_y$ are determined by the difference between maximum and minimum coordinates of the vessel locations in $x$ and $y$ directions. Similarly, considering the step size of $0.05$ for discrete curvature values, $n_\kappa$ is obtained by division of the difference between maximum and minimum of the available curvature values in the patch over the step size. Moreover, the number of steps $H$ used in generating the random paths is set automatically as one third of the patch size. The second set of parameters is set manually. The number of discrete orientations ($n_\theta$), number of iterations in the Monte-Carlo simulation ($n$) and $\sigma_\kappa$ (used in (\ref{eq:connectKernel})) were set to 18, 100000 and 1 respectively and kept constant for all the cases. The other parameters are presented in Table~\ref{tbl:params} for each category separately and for all the cases (in \textit{mean} $\pm$ \textit{standard deviation} format). 
\begin{table}[htbp]
   \centering
   \caption{The number of analyzed patches, the parameters used during numerical simulation and the measured performance values per category and in total.}
\resizebox{3.4in}{!}{  
  \begin{tabular}{@{\extracolsep{1pt}}c  c*{4}{c} } 
  	 \hline 
	 \multirow{2}{*}{Group}&\multirow{2}{*}{Size}&\multicolumn{2}{c}{Parameters}&\multicolumn{2}{c}{Performance} \\
	 \cline{3-4}  \cline{5-6}
	 	&	&	$\sigma_\kappa$	&	$\sigma_{int}$	&	$CDR\%$	&	$Q_{clust}$	\\
		\hline \hline
      A		&	16	& 	$0.0079\pm0.0247$	&	$0.2532\pm0.0013$	&	0.8125	& 	0.9866	\\
      B		&	42	&	$0.0011\pm0.0027$	&	$0.2996\pm0.0026$	&	0.7143	&	0.9979	\\
     C		&	48	&	$0.0031\pm0.0044$	&	$0.2512\pm0.0093$	&	0.8542	&	0.9978	\\
     D		&	31	&	$0.0038\pm0.0047$	&	$0.2578\pm0.0016$	&	0.7742	&	0.9974	\\
     E		&	19	& 	$0.0025\pm0.0041$	&	$0.2699\pm0.0046$	&	0.7368	&	0.9967	\\
    [2ex] 
     A1	&	9	& 	$0.0018\pm0.0024$	&	$0.2574\pm0.0037$	&	0.6667	&	0.9756	\\
     B1	&	23	& 	$0.0004\pm0.0010$	&	$0.2951\pm0.0054$	&	0.6522	&	0.9987	\\
    C1	&	41	&	$0.0030\pm0.0044$	&	$0.2993\pm0.0078$	&	0.8537	&	0.9975	\\
     D1	&	27	& 	$0.0040\pm0.0048$	&	$0.2552\pm0.0018$	&	 0.8148	&	0.9974	\\
     E1	&	18	& 	$0.0021\pm0.0038$	&	$0.2926\pm0.0018$	&	0.7222	&	 0.9967	\\
    [2ex]  
     All 	&	274	& 	$0.0031\pm0.0087$	&	$0.2750\pm0.0072$	&	0.7602	&	0.9942	\\
     \hline 
    \end{tabular}
  }
   \label{tbl:params}
\end{table}

For recording the calculation times the whole process has been divided into four steps: the discretization step before creating the kernel; creating the kernel for several curvature values; creating the affinity matrix, and the spectral clustering step. The times are called $t_{disc}$, $t_{kernel}$, $t_{affinity}$ and $t_{clust}$, respectively, and they are affected by several parameters including the number of vessel pixels in each patch ($|v_i|~,i=1,\dots,M$), the number of discrete orientations ($n_\theta$), curvatures ($n_\kappa$) and the dimension of the kernel in $x$ and $y$ dimensions ($n_x$ and $n_y$). To consider these effects, the weighted average of calculation times for each image patch is obtained, so that the weight for each timing is defined as the product of the affecting parameters as. Thus we define the weights as $w_{disc}= n_{x,i} n_{y,i} n_{\theta,i} $, $w_{kernel}=n_{x,i} n_{y,i} n_{\theta,i} n_{\kappa,i}$, $w_{affinity}= w_{clust}=|v_i|^2$, where $i=1,\dots,N$ indicates the patch number among $N$ patches. It is worth mentioning that since the final number of clusters is determined by comparison among the clustering costs of several cluster sizes ($C$), $t_{clust}$ is additionally affected by the number of examined cluster sizes ($n_{c} = 20$ for all the cases), so the final clustering time is divided by $n_c$.
The size of the affinity matrix for patch $i$ is $|v_i^2|$. These weighted times and their normal average over all patches per step are presented in Table~\ref{tbl:timings}.
\begin{table}[htbp]
   \centering
   \caption{The weighted and normal mean of the processing time (in seconds) of each step in analyzing retinal patches.}
    \begin{tabular}{ c*{5}{c} } 
    \hline 
    &	$\overline{t}_{disc}$	&	$\overline{t}_{kernel}$	& $\overline{t}_{affinity}$	&	$\overline{t}_{clust}$\\
    \hline\hline
     mean(s) &	 0.06	&	43.26	&	17.86	&	0.86	\\
    weighted mean (s)  &		0.06	&	60.64 	&	17.73	&	1.06	\\
    \hline
    \end{tabular}
   \label{tbl:timings}
\end{table}

\begin{figure*}[htbp]
  \centering
 	 \begin{subfigure}[b]{6.8in}
	\centering
 				\qquad \hspace{0.05in}
			 	\includegraphics[width=1in]{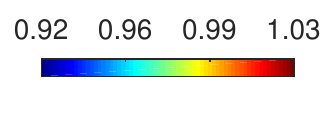} 
	  			\includegraphics[width=1in]{thetaColorRet}
	  			\includegraphics[width=1in]{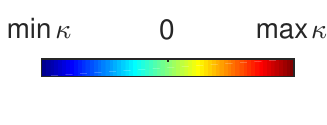} 
	\end{subfigure}
	\begin{subfigure}[b]{6.8in}
	\centering
 			 \makebox[0pt][r]{\makebox[15pt]{\raisebox{25pt}{\rotatebox[origin=c]{0}{A1}}}}  \qquad
				\includegraphics[width=\siz,height=\siz]{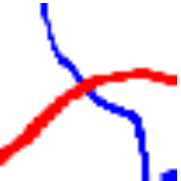} \qquad
	  			\includegraphics[width=\siz,height=\siz]{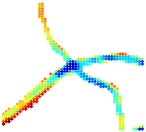}\qquad
	  			\includegraphics[width=\siz,height=\siz]{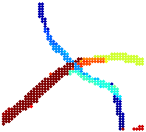} \qquad
	  			\includegraphics[width=\siz,height=\siz]{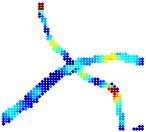} \qquad
	  			\includegraphics[width=\siz,height=\siz]{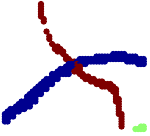} 
	\end{subfigure}
	\begin{subfigure}[b]{6.8in}
	\centering
 			 \makebox[0pt][r]{\makebox[15pt]{\raisebox{25pt}{\rotatebox[origin=c]{0}{B1}}}}  \qquad
				\includegraphics[width=\siz,height=\siz]{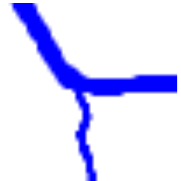} \qquad
	  			\includegraphics[width=\siz,height=\siz]{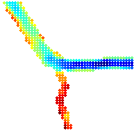}\qquad
	  			\includegraphics[width=\siz,height=\siz]{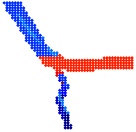} \qquad
	  			\includegraphics[width=\siz,height=\siz]{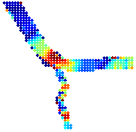} \qquad
	  			\includegraphics[width=\siz,height=\siz]{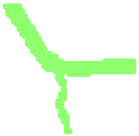} 
	\end{subfigure}
	\begin{subfigure}[b]{6.8in}
	\centering
 			 \makebox[0pt][r]{\makebox[15pt]{\raisebox{25pt}{\rotatebox[origin=c]{0}{C1}}}}   \qquad
				\includegraphics[width=\siz,height=\siz]{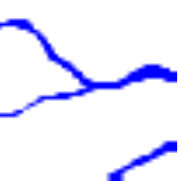} \qquad
	  			\includegraphics[width=\siz,height=\siz]{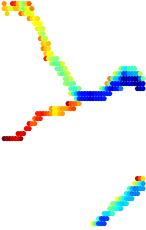} \qquad
	  			\includegraphics[width=\siz,height=\siz]{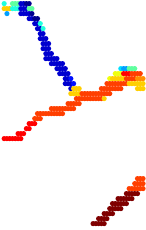}  \qquad
	  			\includegraphics[width=\siz,height=\siz]{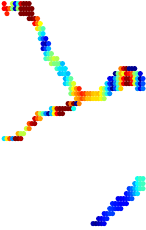}  \qquad
	  			\includegraphics[width=\siz,height=\siz]{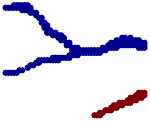} 
	\end{subfigure}
	\begin{subfigure}[b]{6.8in}
	\centering
 			 \makebox[0pt][r]{\makebox[15pt]{\raisebox{25pt}{\rotatebox[origin=c]{0}{D1}}}}  \qquad
				\includegraphics[width=\siz,height=\siz]{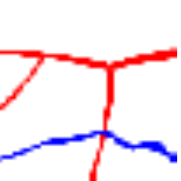} \qquad
	  			\includegraphics[width=\siz,height=\siz]{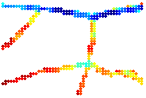} \qquad
	  			\includegraphics[width=\siz,height=\siz]{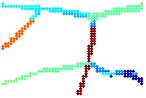}  \qquad
	  			\includegraphics[width=\siz,height=\siz]{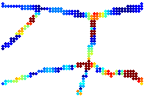} \qquad
	  			\includegraphics[width=\siz,height=\siz]{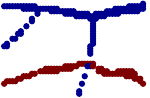} 
	\end{subfigure}
	\begin{subfigure}[b]{6.8in}
	\centering
 			 \makebox[0pt][r]{\makebox[15pt]{\raisebox{25pt}{\rotatebox[origin=c]{0}{E1}}}}   \qquad
				\includegraphics[width=\siz,height=\siz]{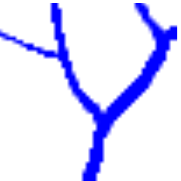} \qquad
	  			\includegraphics[width=\siz,height=\siz]{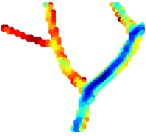} \qquad
	  			\includegraphics[width=\siz,height=\siz]{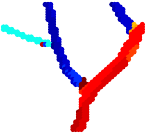}  \qquad
	  			\includegraphics[width=\siz,height=\siz]{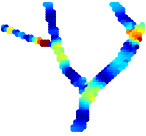}  \qquad
	  			\includegraphics[width=\siz,height=\siz]{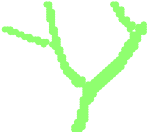} 
	\end{subfigure}
  \caption{Samples of retinal patches in different categories. From left to right, the images in each category represent: artery/vein vessel ground truth, intensity, orientation, curvature and clustering results. The color of the curvature maps are scaled between the maximum and minimum values of the curvature in each image patch.}
  \label{fig:resRetPatch}
\end{figure*}

A set of sample results for various kinds of patches in challenging groups is depicted in Fig.~\ref{fig:resRetPatch}. In this figure, the first column shows the cropped patch from the artery/vein ground truth image. 
The second column depicts the color-coded normalized intensity values taken from the preprocessed image. As seen in these figures, the variation of intensity is high even for small children vessels belonging to one parent vessel. The third and fourth columns represent the color-coded orientations and curvature values. Finally, the last column represents the clusters found in each patch, each shown in an individual color. Other examples are presented in Fig. 2 in the supplementary material.
The presented results, parameters and timings are discussed in the next section.
\subsection{Discussion}
\begin{figure*}[htbp]
  \centering
 	 \begin{subfigure}[b]{6.8in}
 				\hspace{1.4in}
			 	\includegraphics[width=1in]{intColorRet} \hspace{-0.05in}
	  			\includegraphics[width=1in]{thetaColorRet} \hspace{-0.05in}
	  			\includegraphics[width=1in]{curvColorRet} 
	\end{subfigure}
  	\begin{subfigure}[b]{6.8in}
	\centering
 			 \makebox[0pt][r]{\makebox[15pt]{\raisebox{25pt}{\rotatebox[origin=c]{0}{A1}}}}  \qquad
			 	\includegraphics[width=\sizz,height=\sizz]{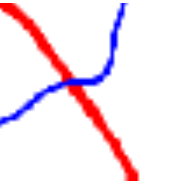} \qquad
	  			\includegraphics[width=\sizz,height=\sizz]{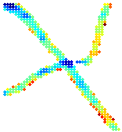}\qquad
	  			\includegraphics[width=\sizz,height=\sizz]{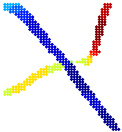} \qquad
	  			\includegraphics[width=\sizz,height=\sizz]{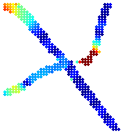} \qquad
	  			\includegraphics[width=\sizz,height=\sizz]{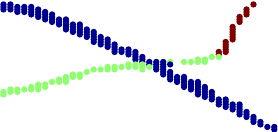} \qquad
				\includegraphics[width=\sizz,height=\sizz]{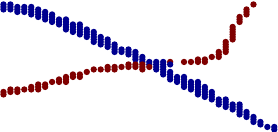} 
	\end{subfigure}
	\begin{subfigure}[b]{6.8in}
	\centering
 			 \makebox[0pt][r]{\makebox[15pt]{\raisebox{25pt}{\rotatebox[origin=c]{0}{C1}}}}  \qquad
			 	\includegraphics[width=\sizz,height=\sizz]{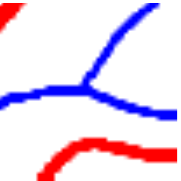} \qquad
	  			\includegraphics[width=\sizz,height=\sizz]{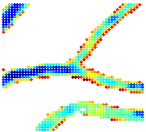}\qquad
	  			\includegraphics[width=\sizz,height=\sizz]{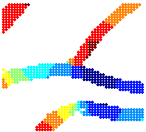} \qquad
	  			\includegraphics[width=\sizz,height=\sizz]{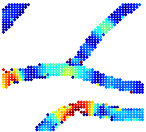} \qquad
	  			\includegraphics[width=\sizz,height=\sizz]{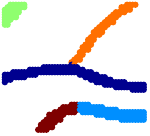} \qquad
				\includegraphics[width=\sizz,height=\sizz]{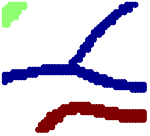} 
	\end{subfigure}
	  \caption{Comparison between clustering results obtained using the connectivity kernel introduced in this work and the one proposed by~\cite{favali2016analysis}. From left to right: artery/vein vessel ground truth, intensity, orientation, curvature and clustering results with the previous and the new kernels.}
  \label{fig:resRetPatchOldKernel}
\end{figure*}

Based on the results presented in the previous section, the main advantage of the new method is that by including the curvature information as an additional contextual information, the kernel adapts itself naturally according to the available data. If the curvature is high, the kernel rotates as well, otherwise it finds a closer path to the points which are collinear with respect to the reference point. In both datasets, the bifurcations are grouped with the parent vessel, but at crossovers with small crossing angles, despite their similar appearance to junctions, the vessels are totally separated. The main reason is that the curvature at junction points is high (because of sudden change of orientation), while for crossings the orientation for individual vessels changes only slightly (in most of the cases). This is advantageous not only in differentiating between junctions and crossings, but also in separation of arteries from veins or crossing tree structures from each other. Fig.~\ref{fig:resRetPatchOldKernel} shows the clustering results for two retinal patches obtained using the new 5D kernel and the previously introduced 4D kernel by~\cite{favali2016analysis}. This helps in depicting the differences between the two methods visually.  

As presented in Fig.~\ref{fig:resRetPatch}, the intensity is a less informative feature compared to the geometrical features because of its large variation within a small neighborhood; however, in some cases, it is a good local criterion differentiating arteries from veins. Thus it is also included in the final affinity matrix with a smaller effect (using a relatively large $\sigma_{int}$).  

\begin{figure*}[htbp]
	\centering 
	\begin{subfigure}[b]{6.8in}
 				\hspace{2in}
	  			\includegraphics[width=1in]{thetaColorRet} \hspace{0.15in}
	  			\includegraphics[width=1in]{curvPhantoms} 
	\end{subfigure}
	\begin{subfigure}[b]{1.2in}
		\includegraphics[width=0.9in,height=0.9in]{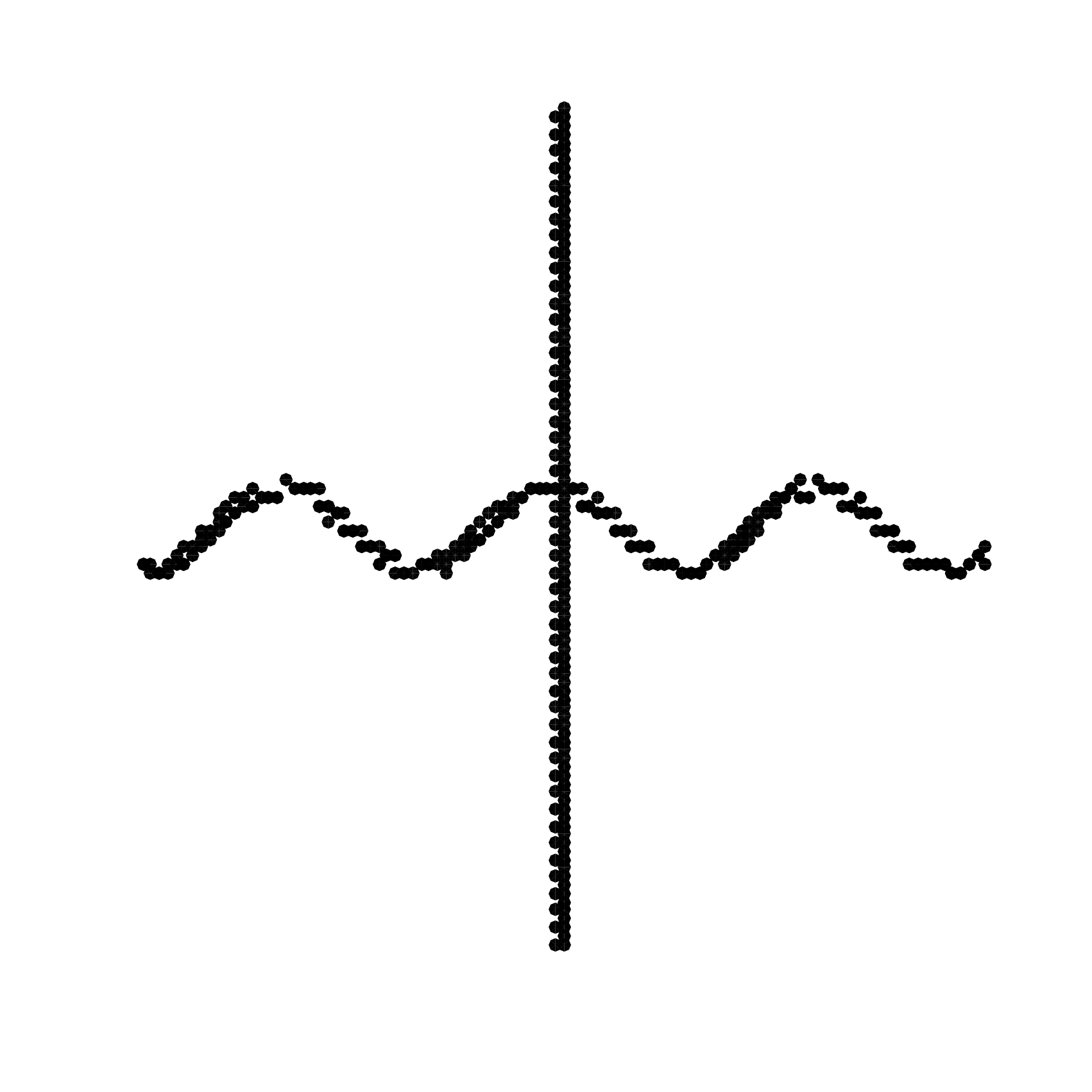} \vspace{-0.2in}
		\includegraphics[width=\sizp,height=\sizp]{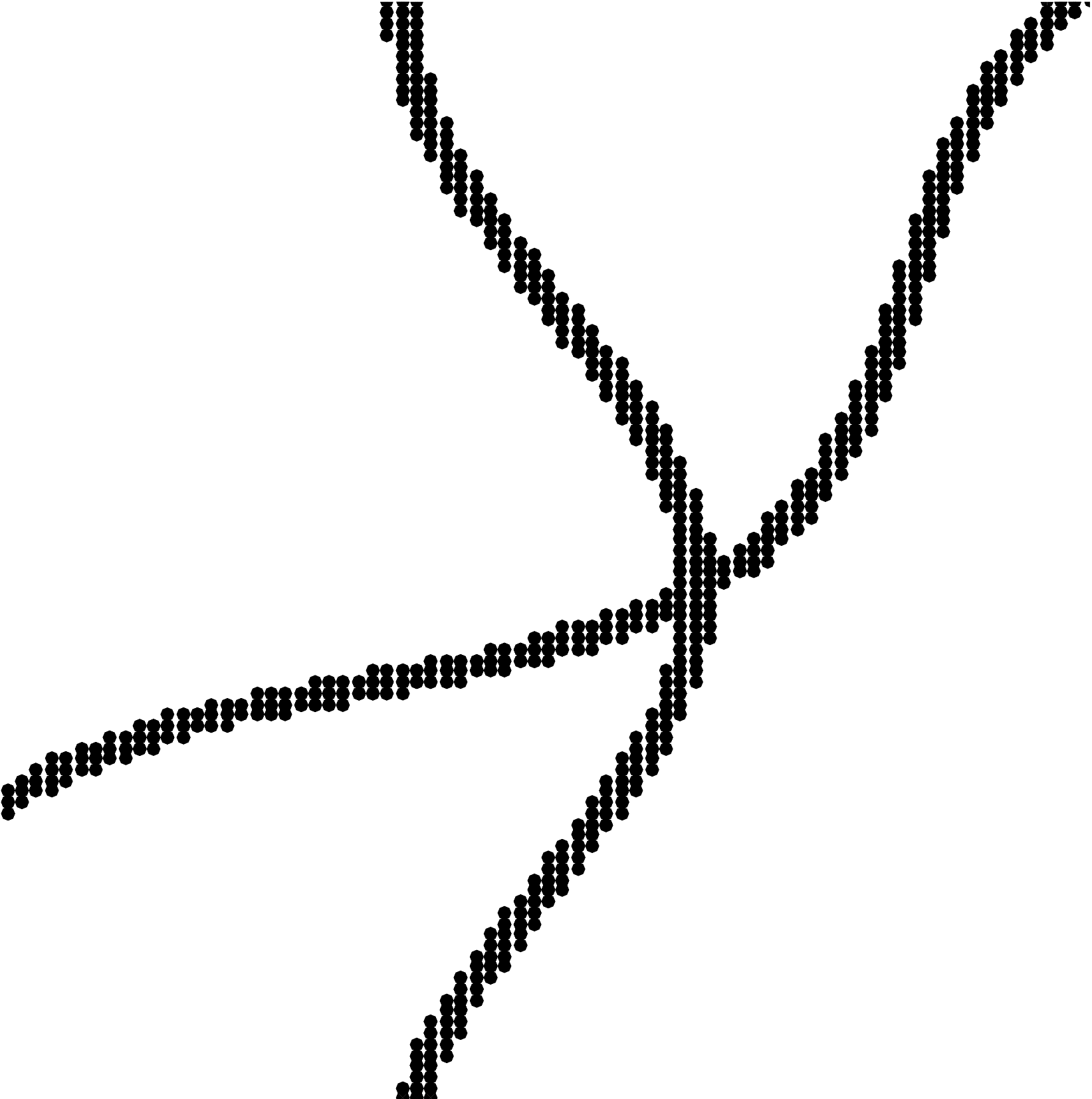} \caption{} \label{fig:limOrig}
	\end{subfigure}
	\begin{subfigure}[b]{1.2in}
		\includegraphics[width=0.9in,height=0.9in]{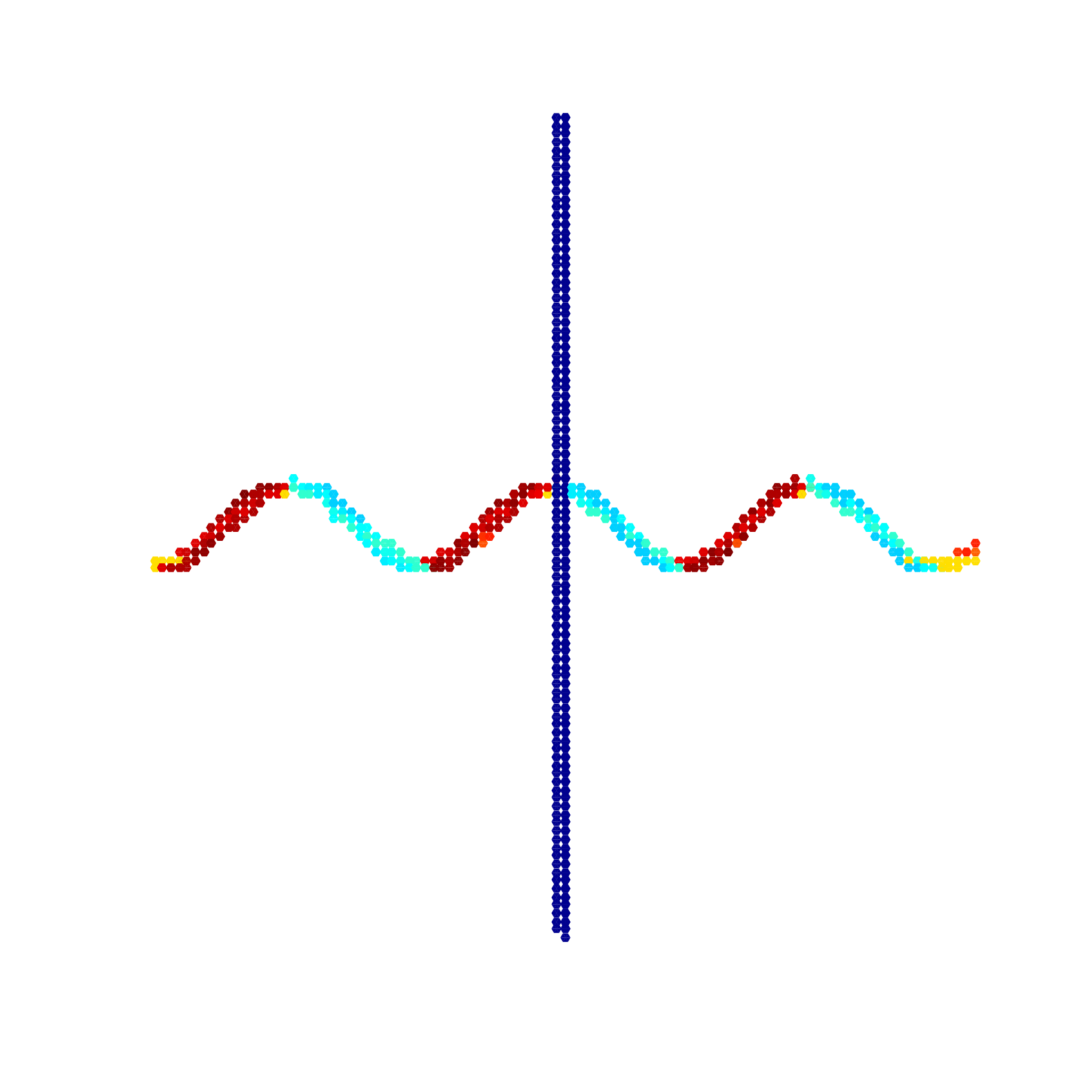}  \vspace{-0.2in}
		\includegraphics[width=\sizp,height=\sizp]{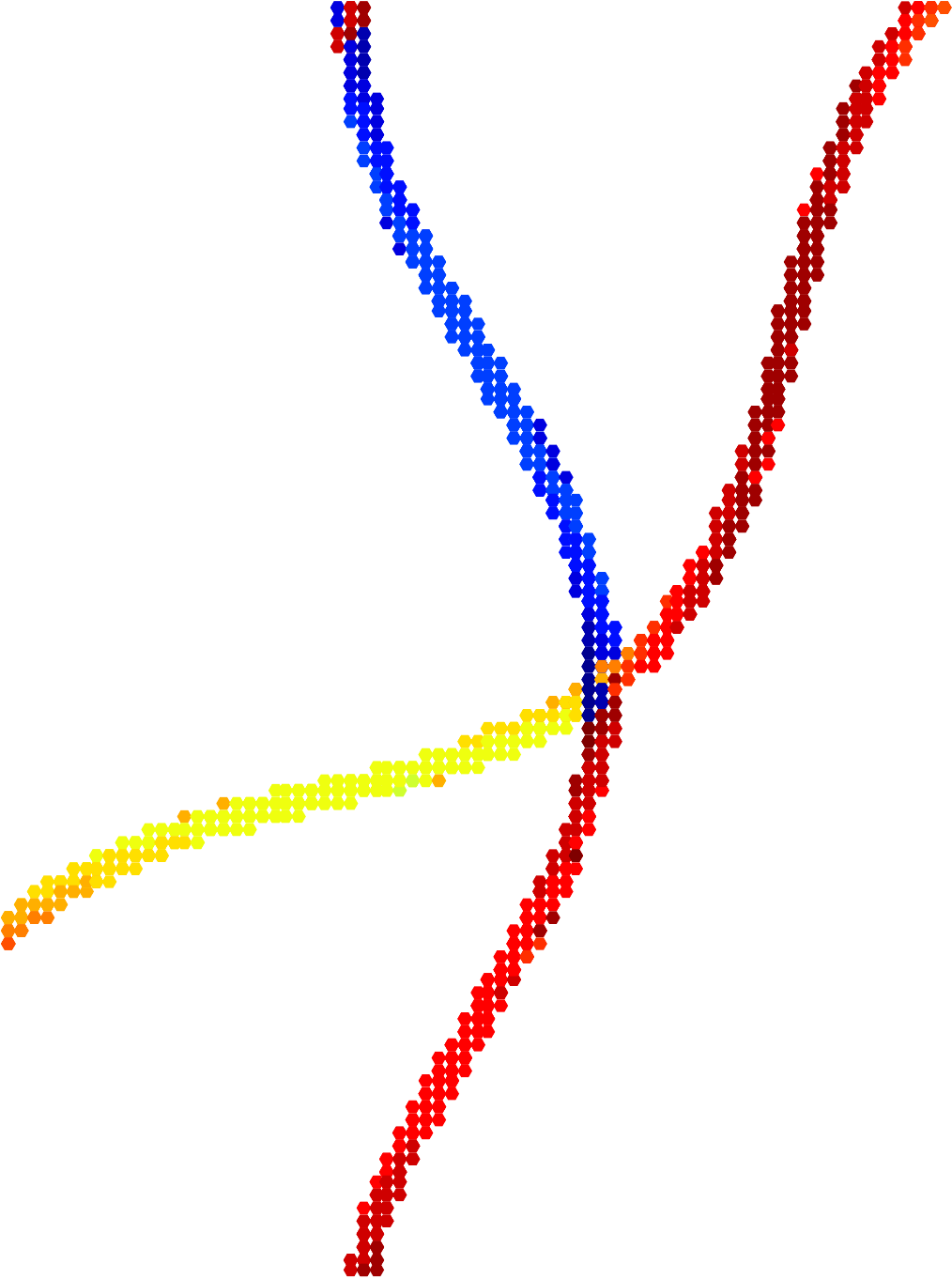} \caption{} \label{fig:limOs}
	\end{subfigure}
	\begin{subfigure}[b]{1.2in}
		\includegraphics[width=0.9in,height=0.9in]{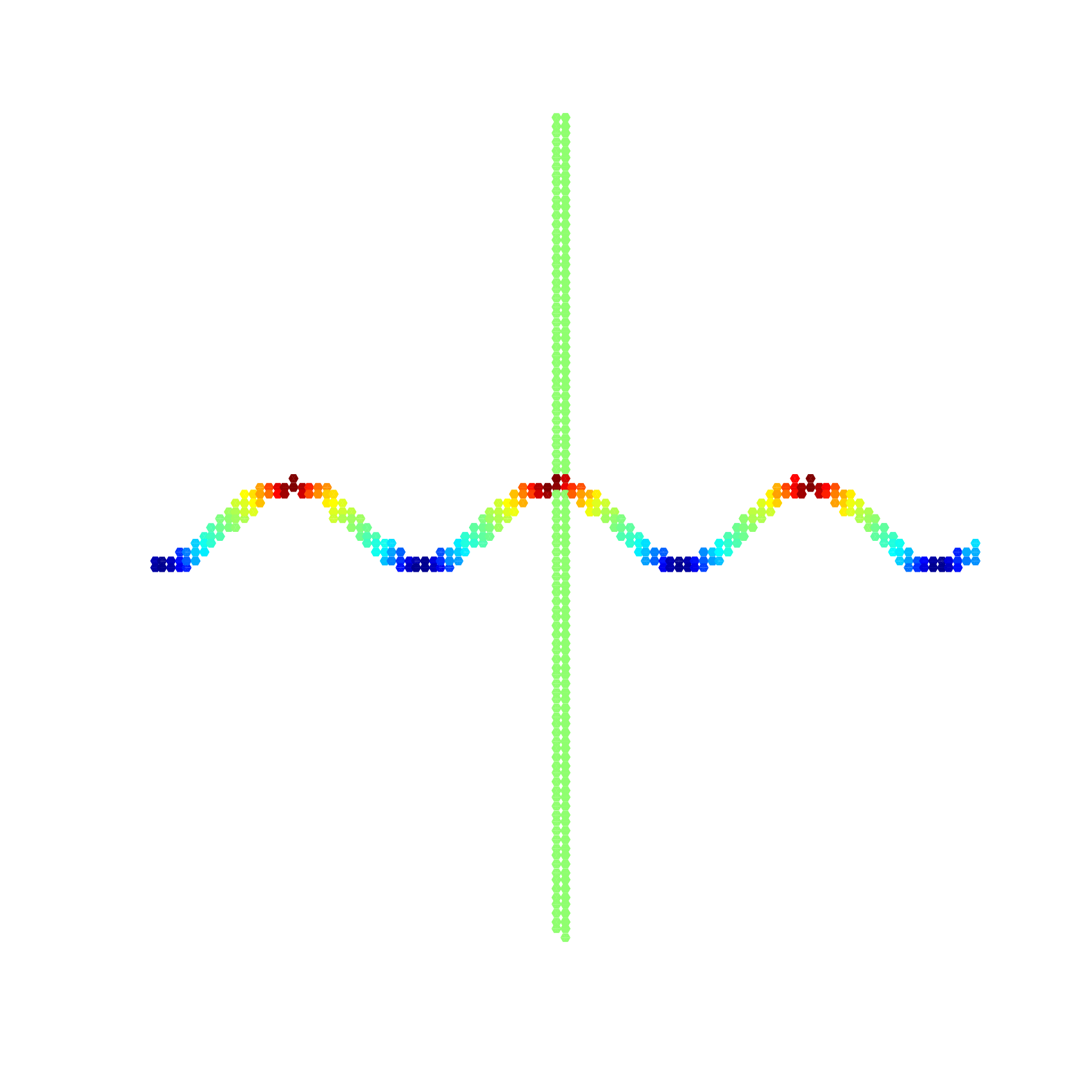} 	\vspace{-0.2in}	
		\includegraphics[width=\sizp,height=\sizp]{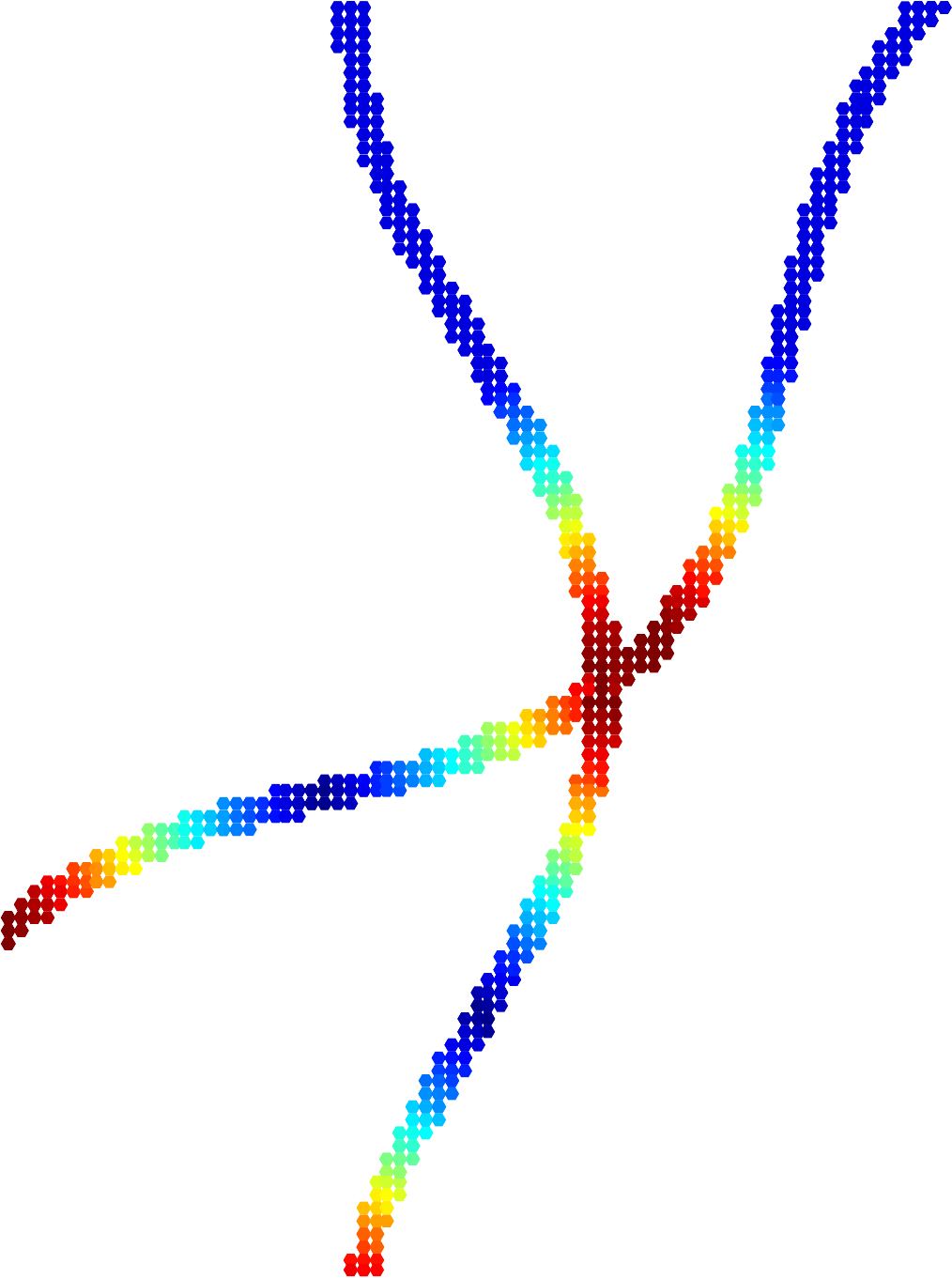} \caption{} \label{fig:limCurv}
	\end{subfigure}
	\begin{subfigure}[b]{1.2in}
		\includegraphics[width=0.9in,height=0.9in]{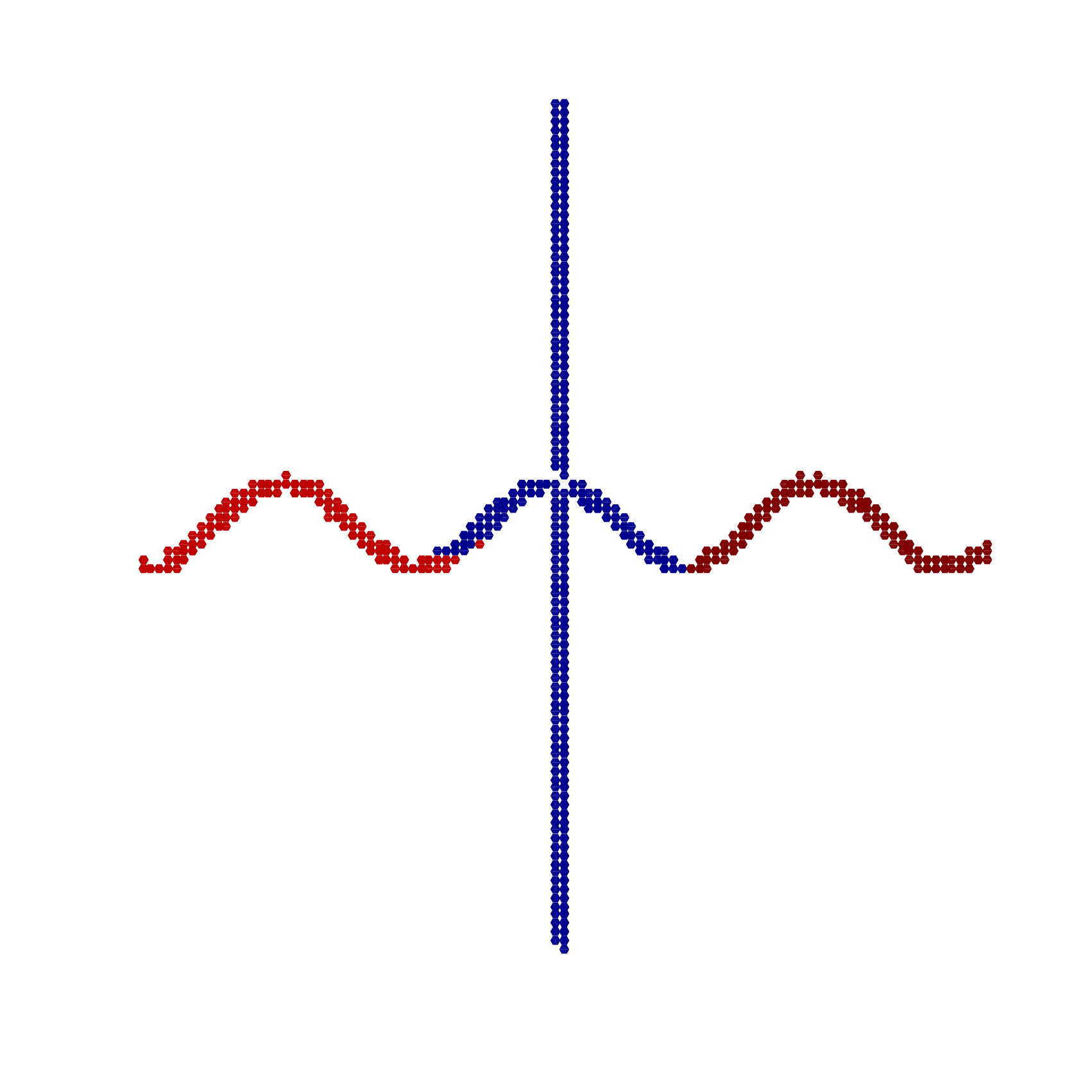} 	\vspace{-0.2in}	
		\includegraphics[width=\sizp,height=\sizp]{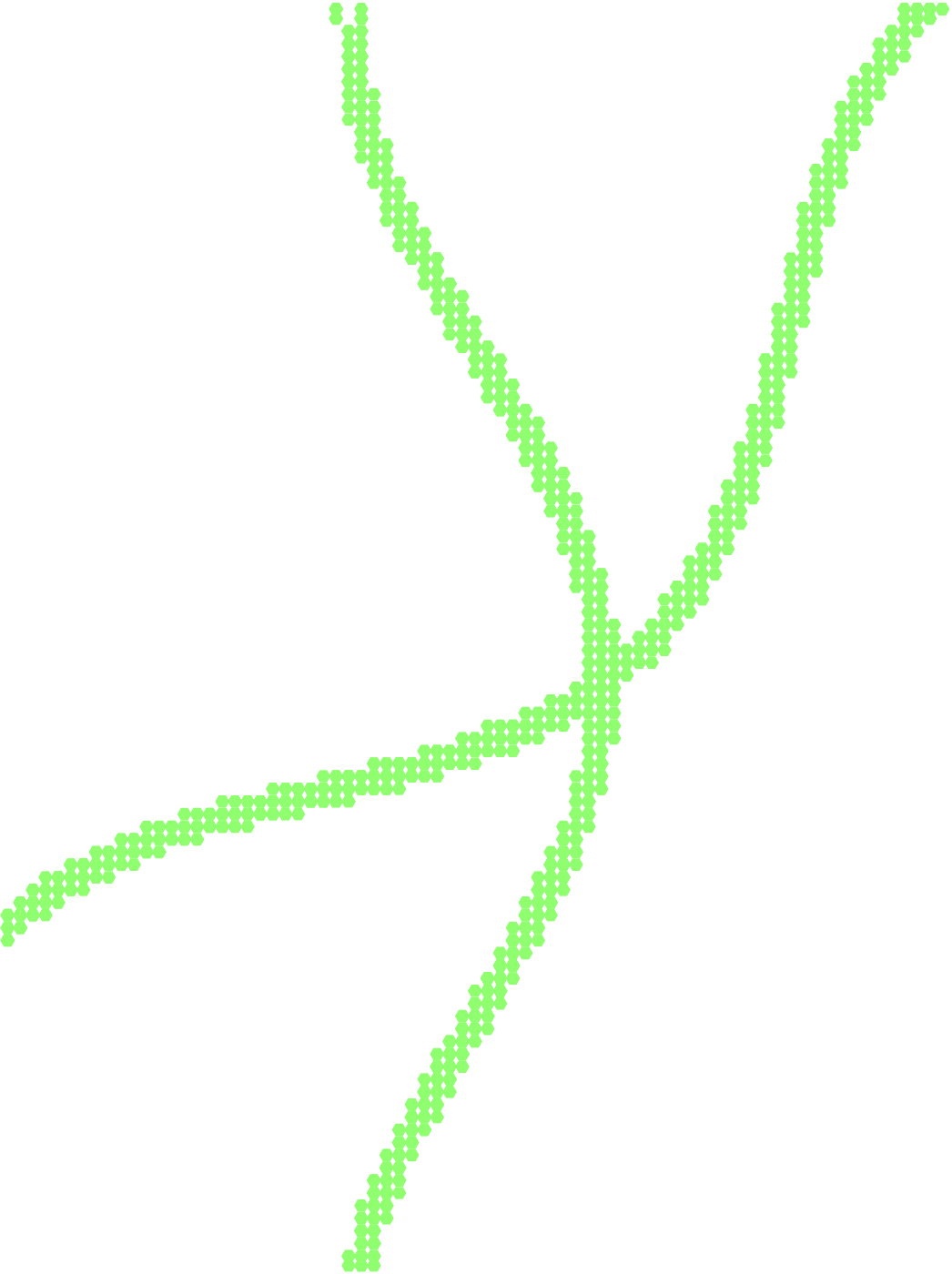} \caption{} \label{fig:limClust}
	\end{subfigure}
	\caption{From left to right: the stimuli (\subref{fig:limOrig}), the orientation maps (\subref{fig:limOs}), the curvature maps (\subref{fig:limCurv}) and the clustering results with the new kernel (\subref{fig:limClust}). }
	\label{fig:lim}
\end{figure*}

\begin{figure*}[htbp]
  \centering
 	  \begin{subfigure}[b]{6.8in}
	\centering
			 	\includegraphics[width=1in]{intColorRet} 
	  			\includegraphics[width=1in]{thetaColorRet}
	  			\includegraphics[width=1in]{curvColorRet} 
	\end{subfigure}
  	\begin{subfigure}[b]{6.8in}
	\centering
			 	\includegraphics[width=\sizz,height=\sizz]{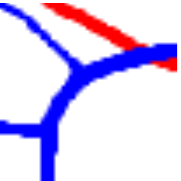} \qquad
	  			\includegraphics[width=\sizz,height=\sizz]{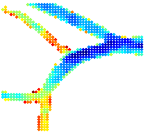}\qquad
	  			\includegraphics[width=\sizz,height=\sizz]{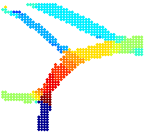} \qquad
	  			\includegraphics[width=\sizz,height=\sizz]{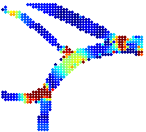} \qquad
				\includegraphics[width=\sizz,height=\sizz]{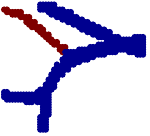} 
	\end{subfigure}
	\begin{subfigure}[b]{6.8in}
	\centering
			 	\includegraphics[width=\sizz,height=\sizz]{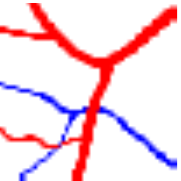} \qquad
	  			\includegraphics[width=\sizz,height=\sizz]{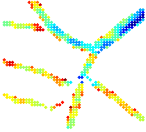}\qquad
	  			\includegraphics[width=\sizz,height=\sizz]{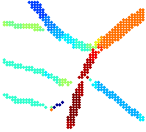} \qquad
	  			\includegraphics[width=\sizz,height=\sizz]{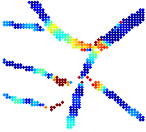} \qquad
				\includegraphics[width=\sizz,height=\sizz]{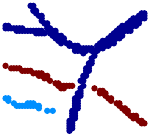} 
	\end{subfigure}
	  \caption{Wrong clustering results on two retinal patches. From left to right: the artery/vein ground truth, vessel intensity, orientation, curvature and clustering results.}
  \label{fig:resRetPatchFailure}
\end{figure*}

Some examples of the limitations of the method in clustering the phantom and retinal patches are represented in Fig.~\ref{fig:lim} and \ref{fig:resRetPatchFailure} respectively. For phantom cases, the presence of very high curvature combined with the co-circularity and co-linearity of vessels does not allow to obtain a good clustering result. Considering other features, as the intensity, could be helpful in solving this problem. However, as shown in the top row of Fig.~\ref{fig:resRetPatchFailure}, the feature of intensity is not useful for correct clustering. In this image, one of the bifurcations has been assigned as a vessel crossing the other one because it is almost orthogonal to its parent vessel; while the other crossing vessel has been wrongly clustered as a bifurcation. In the bottom row, one of the small bifurcations is totally missing in the segmentation and the other small one is not clustered with its parent vessel because of lack of information. 

The statistical analysis on the parameters used during numerical simulations is presented in Table~\ref{tbl:params}. Based on these results the curvature diffusion constant parameter ($\sigma_\kappa$) changes in a small range for simple and challenging cases per group, except for category A and A1, considering the fact that the number of available challenging patches in category A is small compared to the others. The $\sigma_{int}$ parameter has a small variation as well. Therefore, it is reasonable to use the mean value in general for examining new patches in each group. It is worth mentioning that all the patches examined in this work have been selected from retinal images with the same resolution and respective pixel size. If the pixel size increases or decreases, the $\sigma_\kappa$ controlling the size of the 3D kernel needs to be increased or decreased accordingly. It is also important to mention that compared to previous work, the eigensystem analysis is fully automatic in this work, due to the use of self-tuning spectral clustering. Therefore, no additional parameters need to be tuned for this step.

Qualitative and quantitative results indicate the better performance of the method on all kinds of retinal patches and challenging structures. Compared to the previous work~\cite{favali2016analysis}, the $CDR\%$ performance values have changed for some groups. There are two main reasons. On one hand, in the previous work, the $CDR\%$ was calculated for 20 patches per group; while in this work each group is categorized into two groups depending on the available structure and also the number of patches per group is different. On the other hand, in this work we evaluated the performance with the assumption that the bifurcations need to be grouped with the main parent vessel, while in previous work, due to use of one elongated kernel, the assumption was to have at least two separate units depending on the bifurcation angle. Another minor difference is that the dataset has changed and the patches have been selected from a different set of retinal images. Therefore, it is not fair to make a one-by-one comparison to the previous results. 

Last but not least point is about the computation times. The codes are implemented in Matlab and the times are measured on an Apple Macbook Air, Intel Core i7, 1.7 GHz processor and 8GB of memory. The most time consuming step as presented in Table~\ref{tbl:timings} is the calculation of several 3D kernels. The number of kernels depends on the number of discrete curvature values ($n_\kappa$) existing in the data, and the computation time as mentioned before depend on its size. However, the kernels can be calculated only once for all possible curvature values and then used for analyzing all other patches. The next most time consuming part is the calculation of the affinity matrix which is performed per pair of points. The weighted average times are good indicators of the complexity of each step. Although they are already still relatively small, they can be improved both from hardware and implementation points of view.

\section{Conclusion}
\label{sec:conclusion}

In this work a new method for analyzing the connectivities in images containing elongated, rotated and curved structures is proposed. The new connectivity model is inspired by the geometry of the primary visual cortex, where the connectivity between all the lifted points into the five-dimensional space of positions, orientations, intensity and curvatures is represented by a 5D kernel. 
We use two sets of patches: a retinal and an artificial one, to identify perceptual units in these figures, showing that this can be considered as  a good quantitative model for their constitution and in general for the law of good continuation. Including the geometrical contextual properties in addition to local features, such as intensity, makes the method very robust against different kinds of variations and missing information which could exist in clinical images. We analyze different challenging cases that are very informative for clinical studies and we show how the proposed method is successful in solving the problem of grouping, also for blood vessels with high curvature. That was a limitation in the previously proposed method~\cite{favali2016analysis} and many of the state-of-the-art techniques. 

In this work, we only analyzed small patches. However, by replacing the self-tuning spectral clustering step with label propagation methods or using the affinity matrix in an optimization framework (e.g. integer programming), the constitution of the entire vasculature network in full retinal images becomes possible. When errors appear in the lifting step due to noise or low contrast regions, and if the measured curvature or orientations are not accurate enough, then the method fails. Therefore, it is essential to validate the curvature and orientation measurement methods in advance. 

The proposed method has a great potential in discrimination and separation of arteries from veins in retinal images and, in a general view, separation of all the tree structures crossing each other in the vasculature network. Most of the segmentation or artery/vein separation methods are local, pixel based techniques which do not take into account the global connectivity of the blood vessels in the network. By including this global connectivity criterion, most of the errors and wrong detections will be eliminated and the problem of missing information will be handled appropriately. This technique is also highly suitable for other application areas, such as finding roads in aerial photographs or detecting cracks and faults in materials.


%
\IEEEpeerreviewmaketitle

\section*{Acknowledgment}
We thank Dr. Erik Bekkers who provided us some parts of his codes for measuring the curvatures in retinal images and creating the phantom images.

\ifCLASSOPTIONcaptionsoff
  \newpage
\fi



%


\bibliographystyle{IEEEtran}

\bibliography{bare_jrnl}
\end{document}


%
\title{Cortically-Inspired Spectral Clustering for Connectivity Analysis in Retinal Images }
%
%
%

%
\markboth{Supplementary Material}%
{Supplementary Material}

 \label{sec:supp}
 \begin{figure*}[htbp]
 	\centering
 	
 	\begin{subfigure}[b]{6.8in}
 		\centering
 		\makebox[0pt][r]{\makebox[15pt]{\raisebox{25pt}{\rotatebox[origin=c]{0}{A}}}}  \qquad
 		\includegraphics[width=\sizp,height=\sizp]{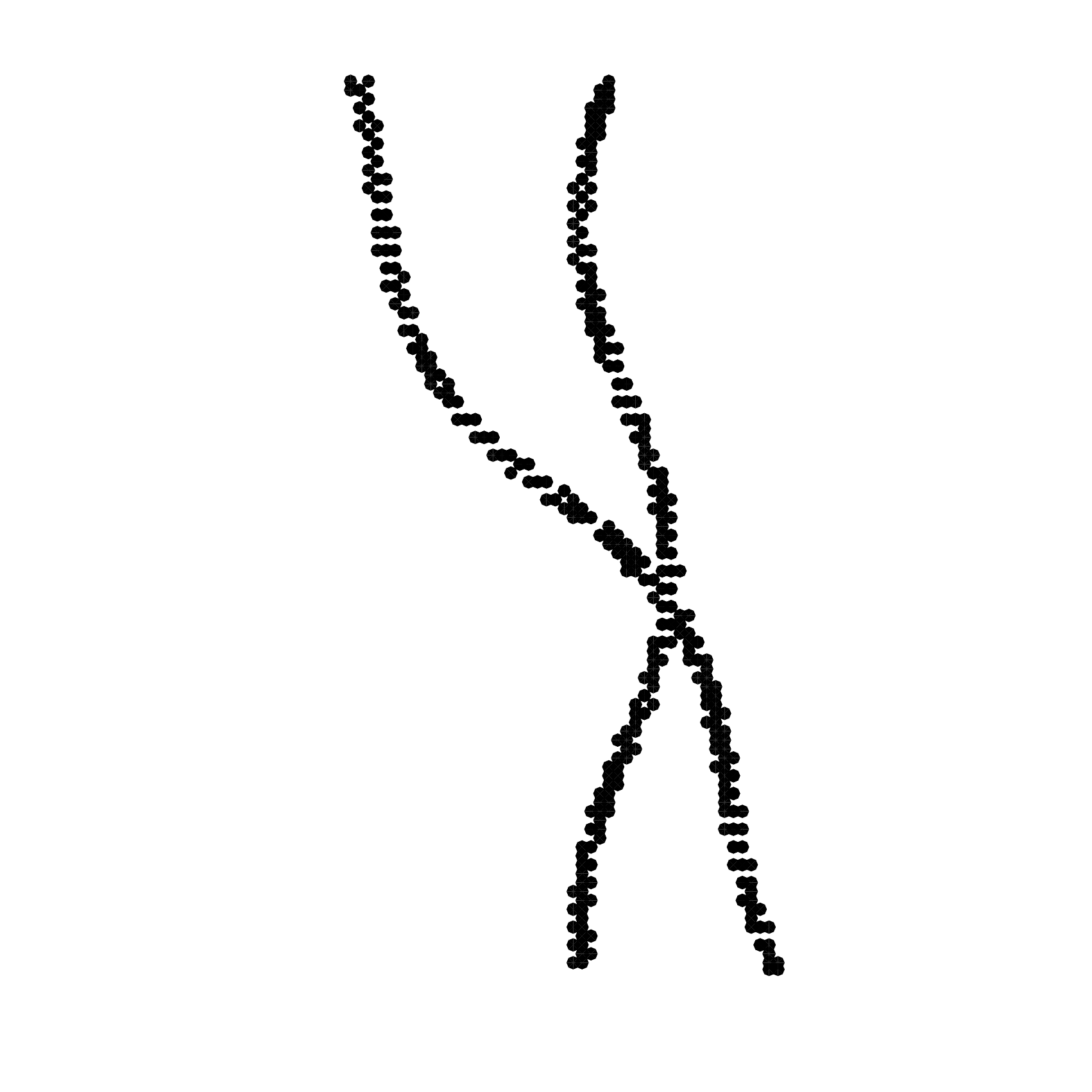} \qquad
 		\includegraphics[width=\sizp,height=\sizp]{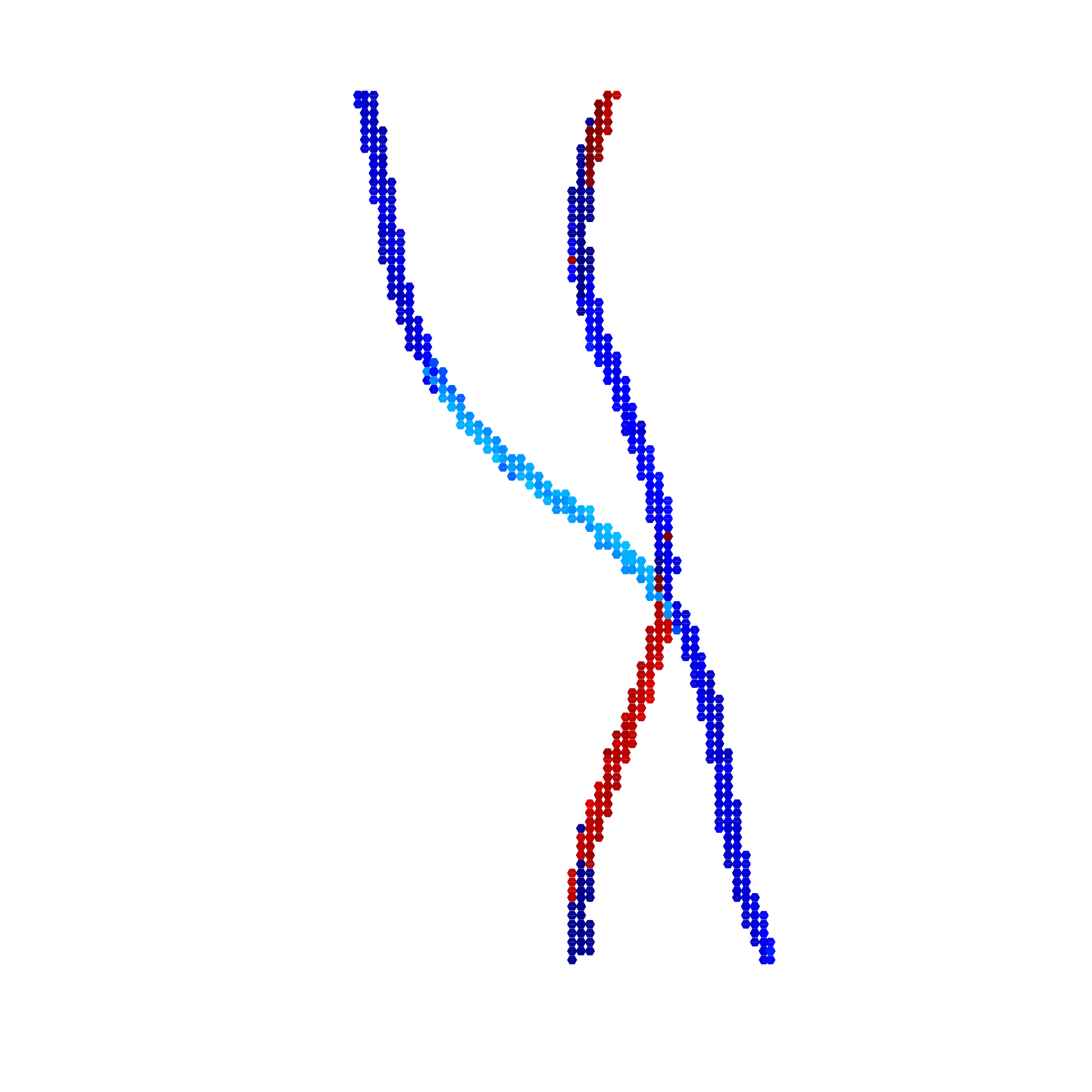} \qquad
 		\includegraphics[width=\sizp,height=\sizp]{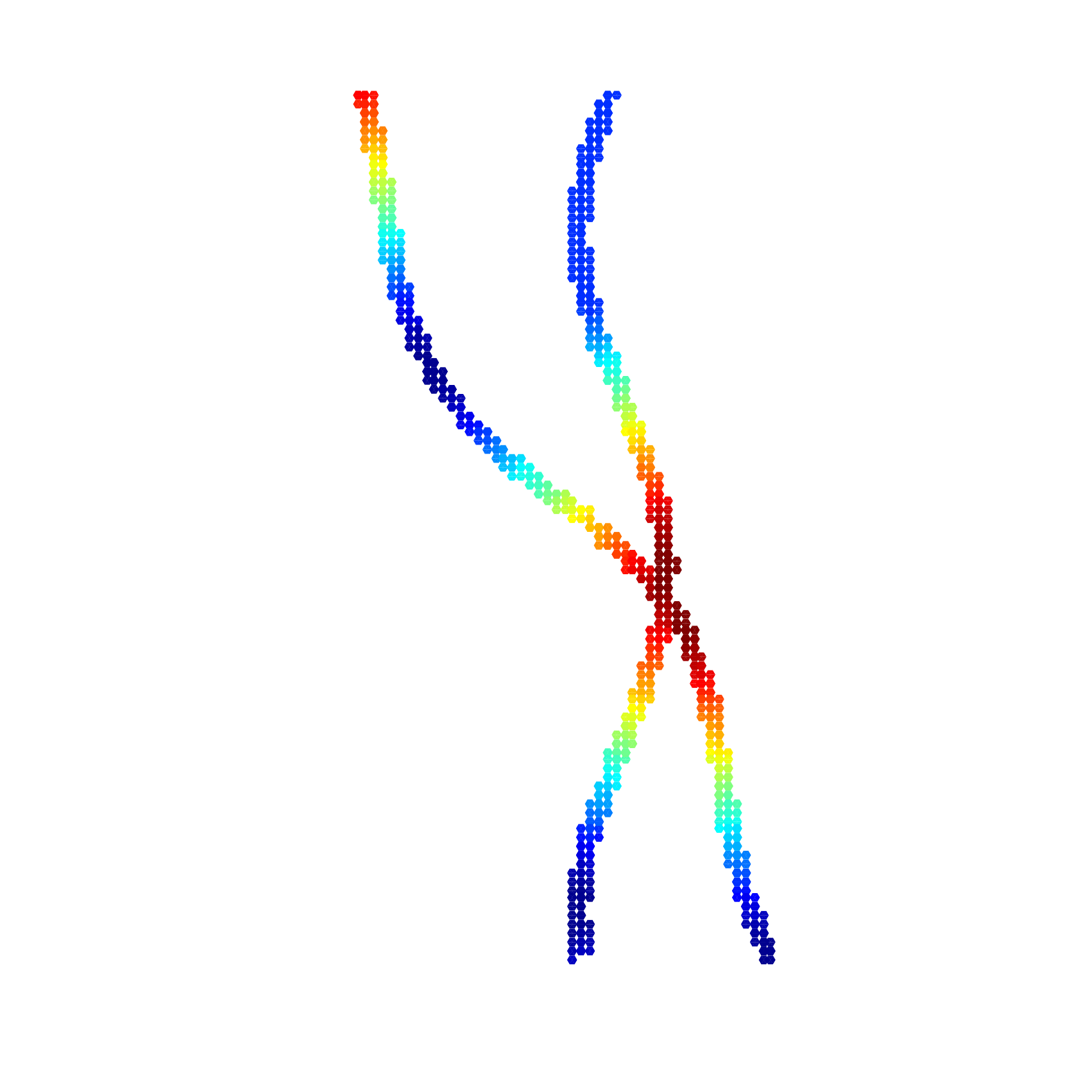}\qquad
 		\includegraphics[width=\sizp,height=\sizp]{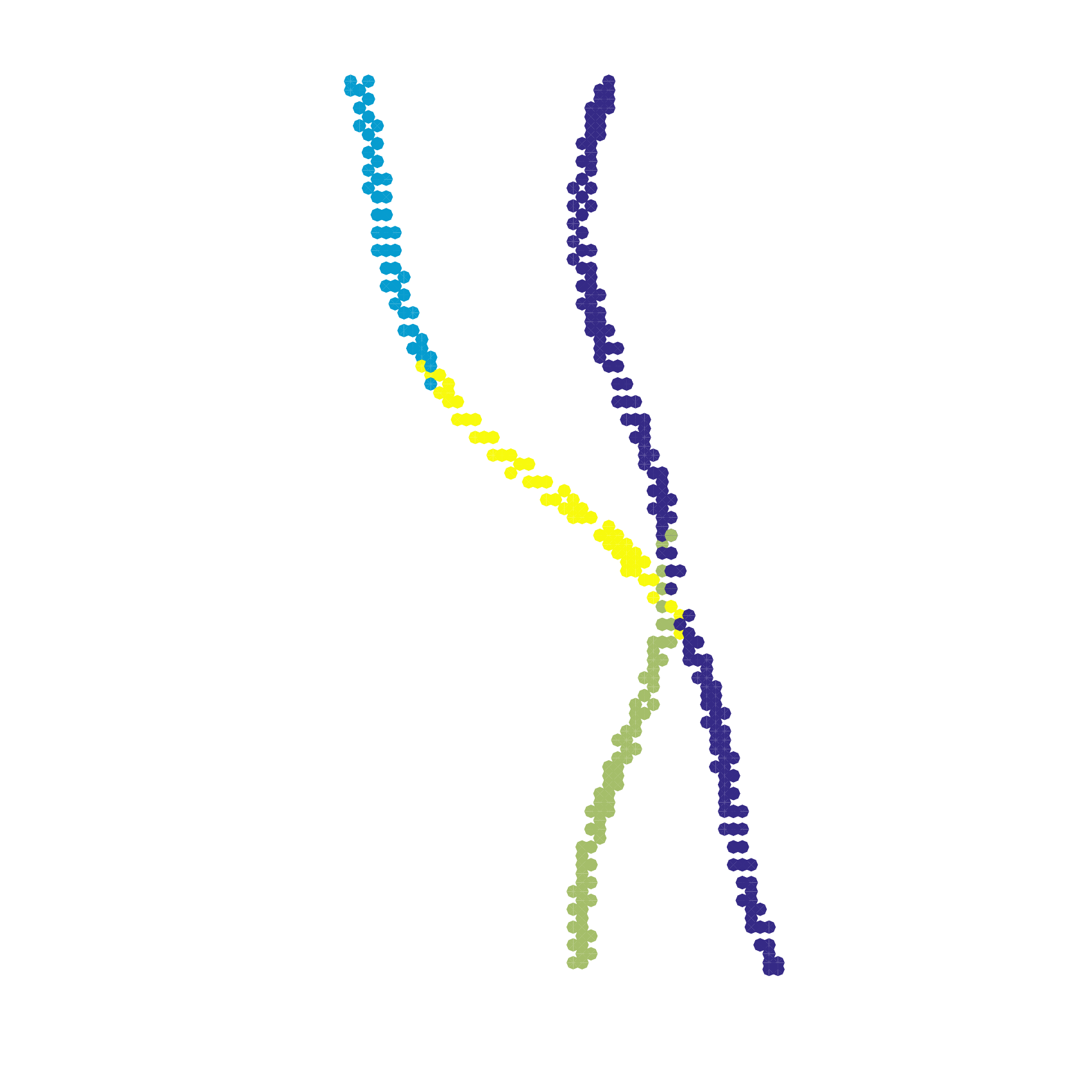} \qquad
 		\includegraphics[width=\sizp,height=\sizp]{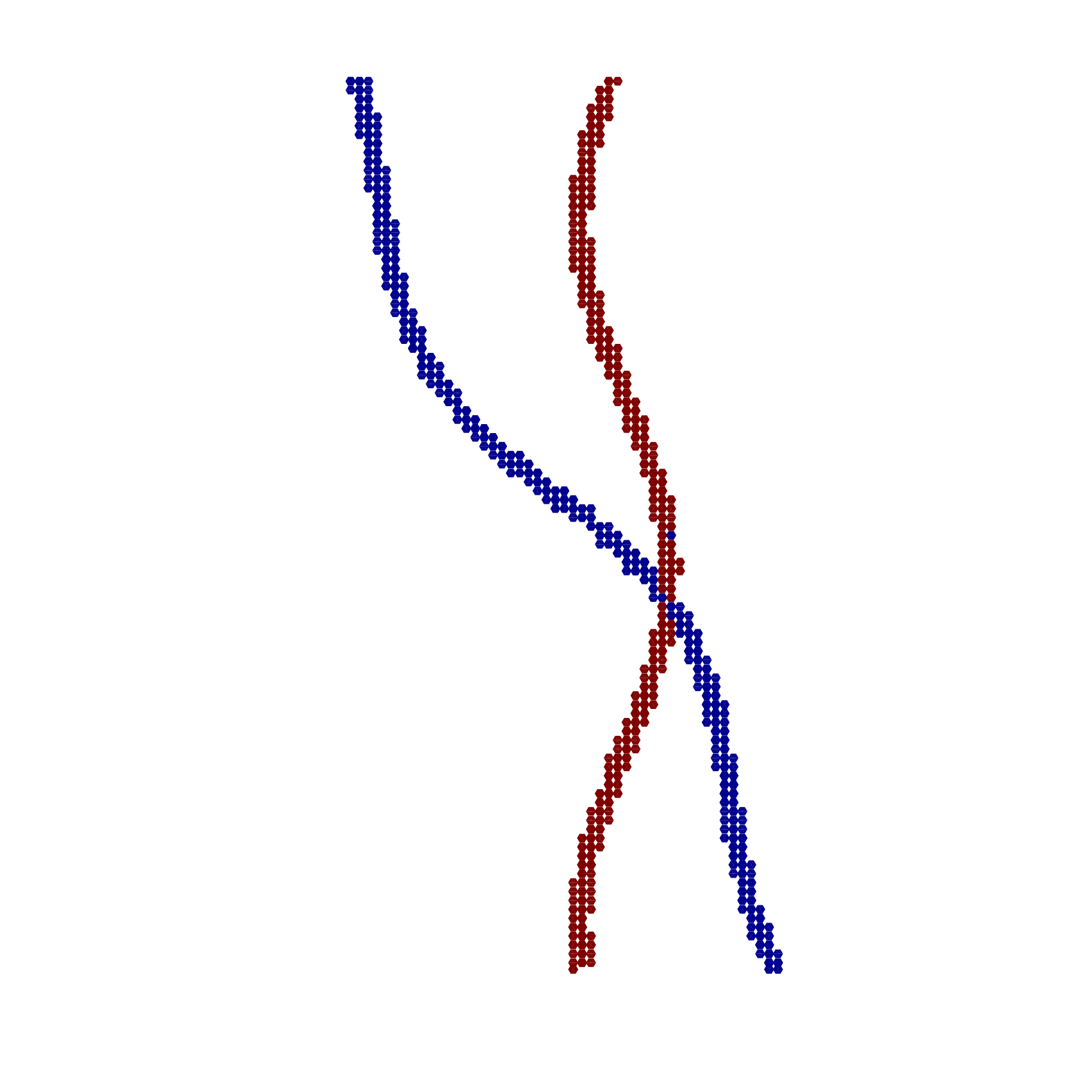} 
 	\end{subfigure}
 	
 	\begin{subfigure}[b]{6.8in}
 		\centering
 		\makebox[0pt][r]{\makebox[15pt]{\raisebox{25pt}{\rotatebox[origin=c]{0}{B}}}}  \qquad
 		\includegraphics[width=\sizp,height=\sizp]{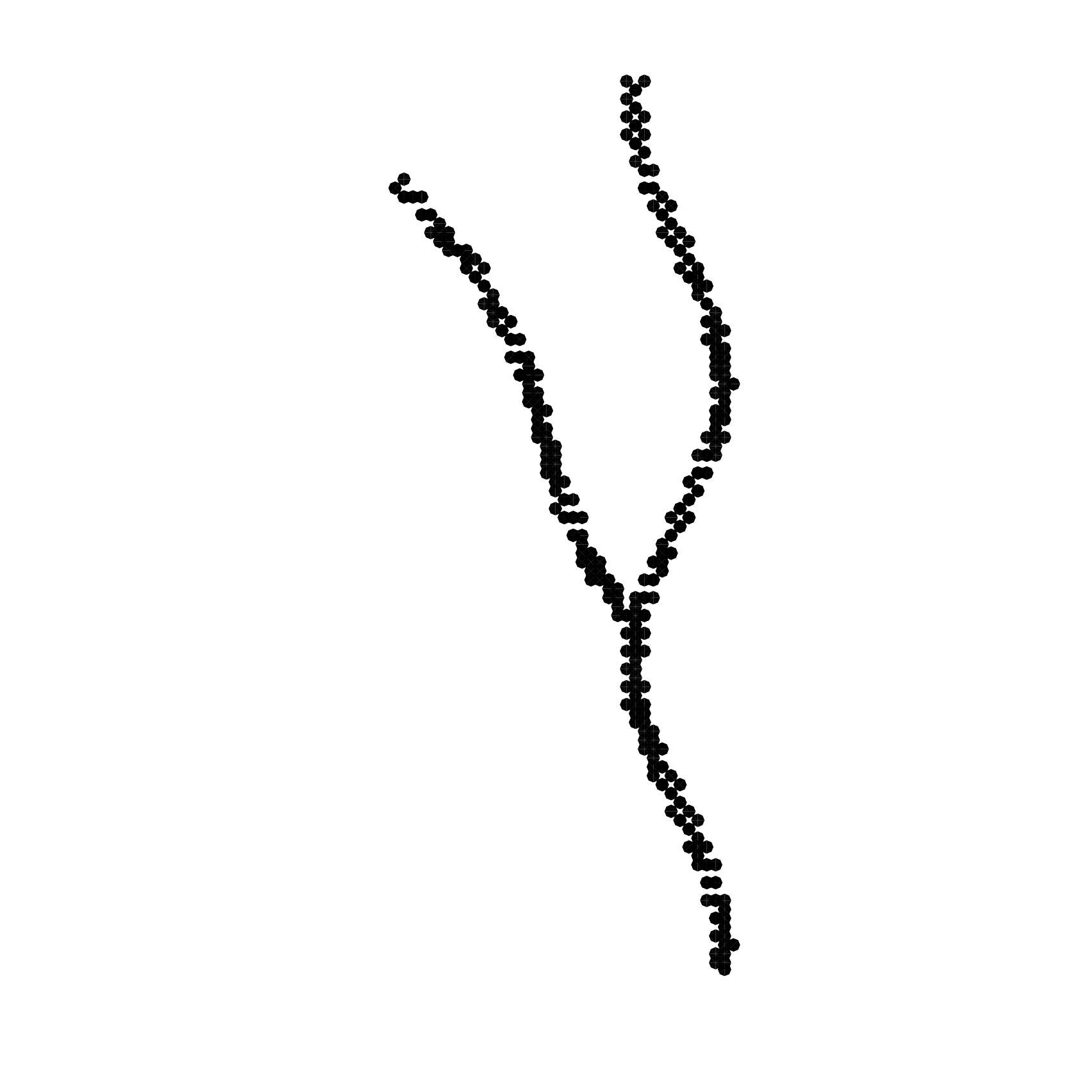} \qquad
 		\includegraphics[width=\sizp,height=\sizp]{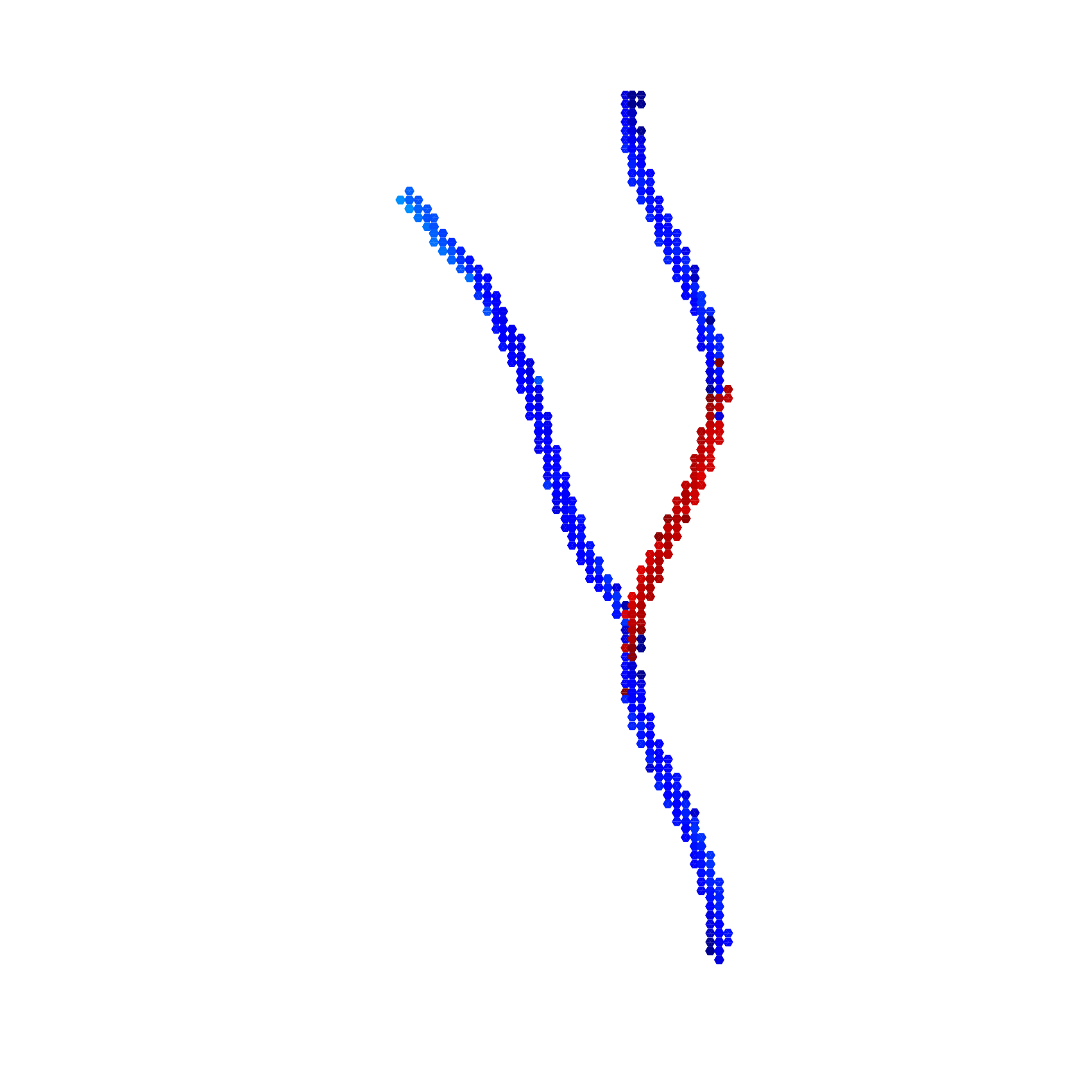} \qquad
 		\includegraphics[width=\sizp,height=\sizp]{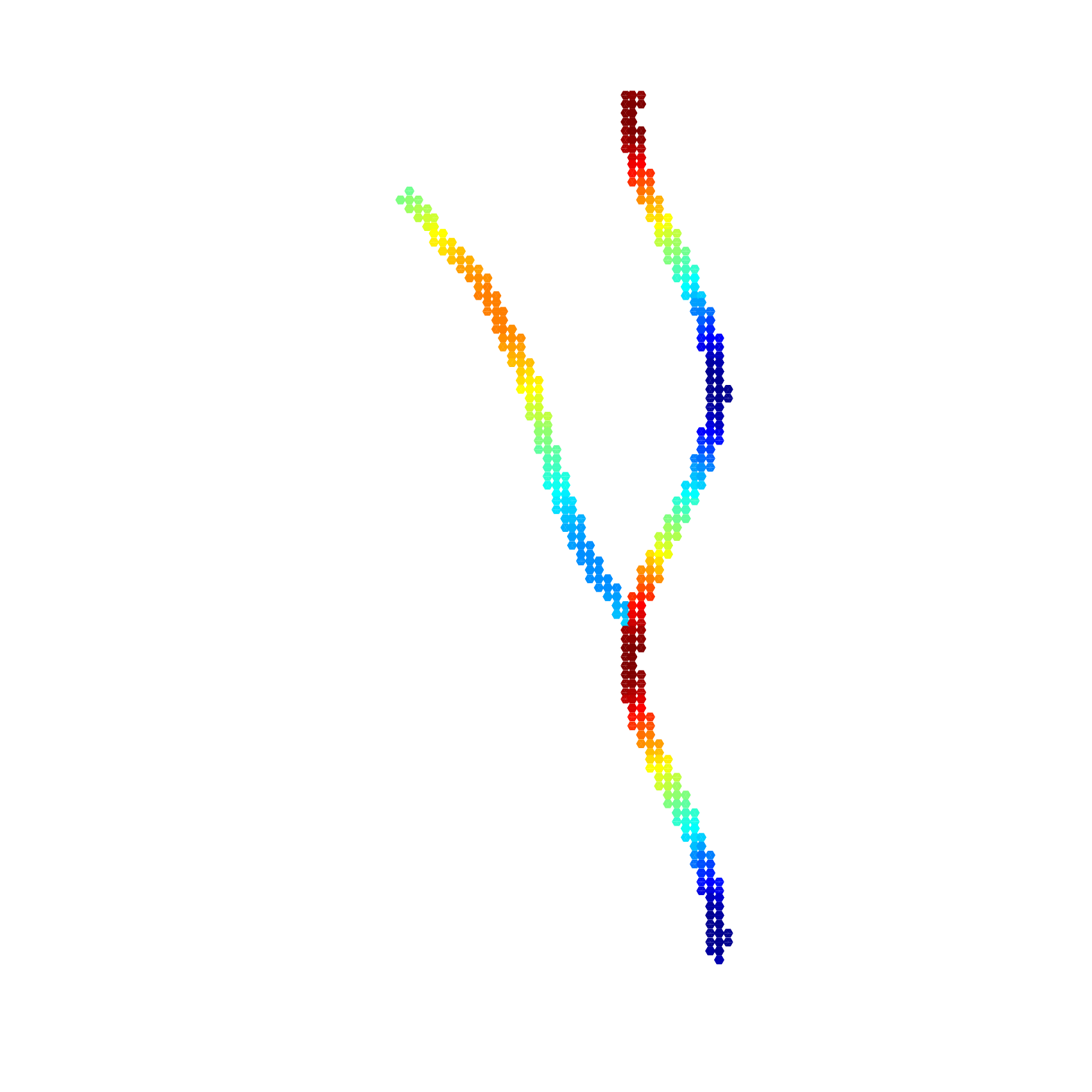}\qquad
 		\includegraphics[width=\sizp,height=\sizp]{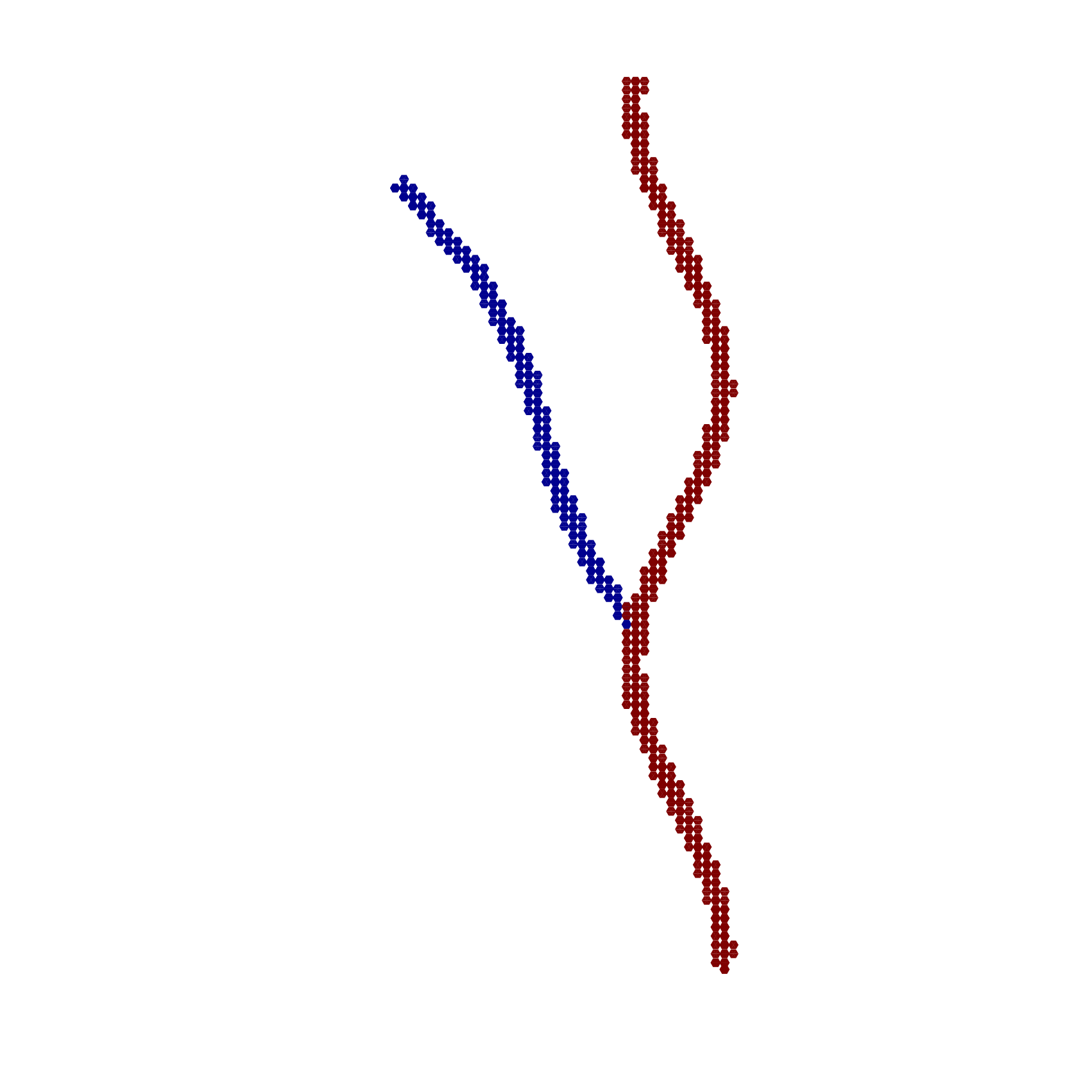} \qquad
 		\includegraphics[width=\sizp,height=\sizp]{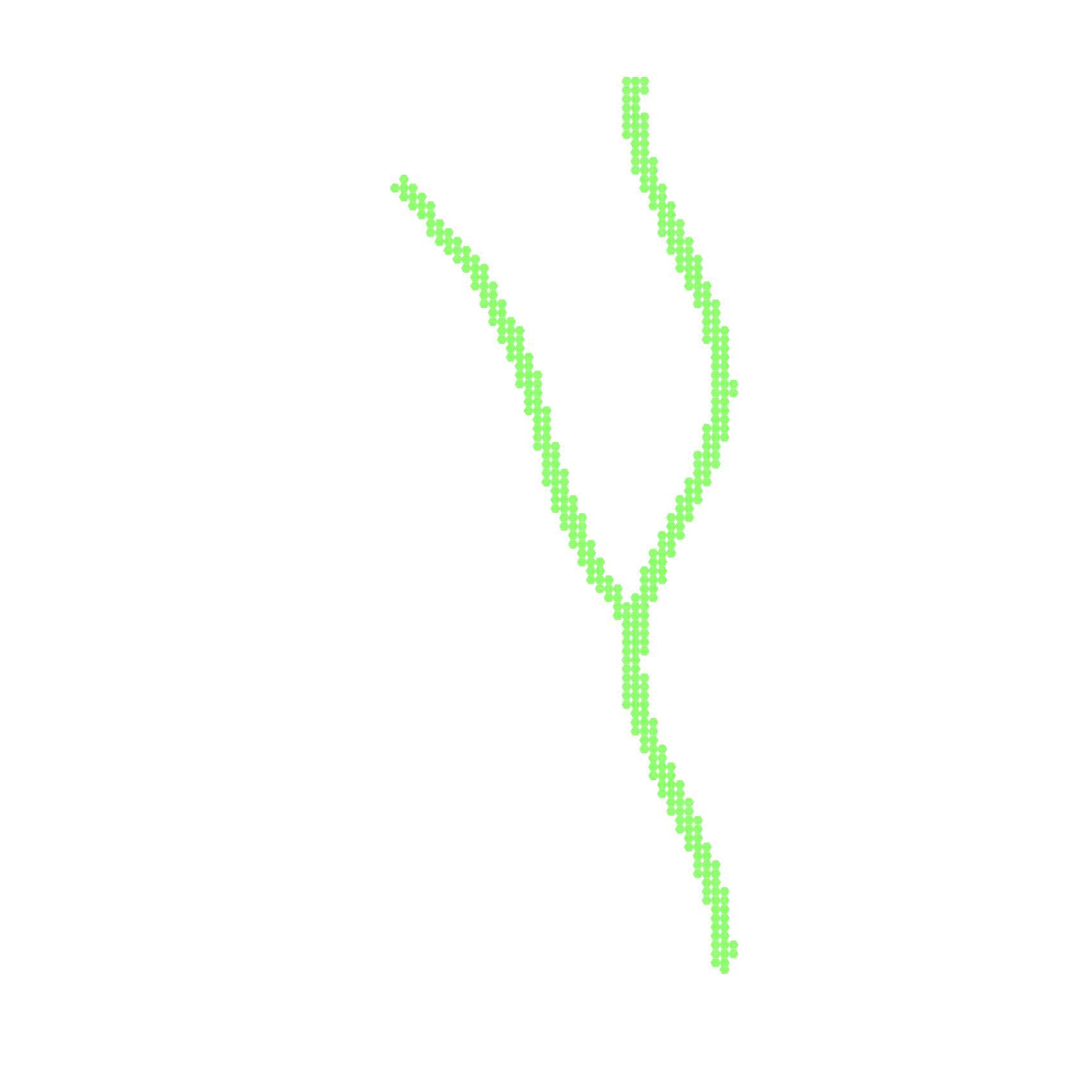} 
 	\end{subfigure}
 	
 	\begin{subfigure}[b]{6.8in}
 		\centering
 		\makebox[0pt][r]{\makebox[15pt]{\raisebox{25pt}{\rotatebox[origin=c]{0}{C}}}}  \qquad
 		\includegraphics[width=\sizp,height=\sizp]{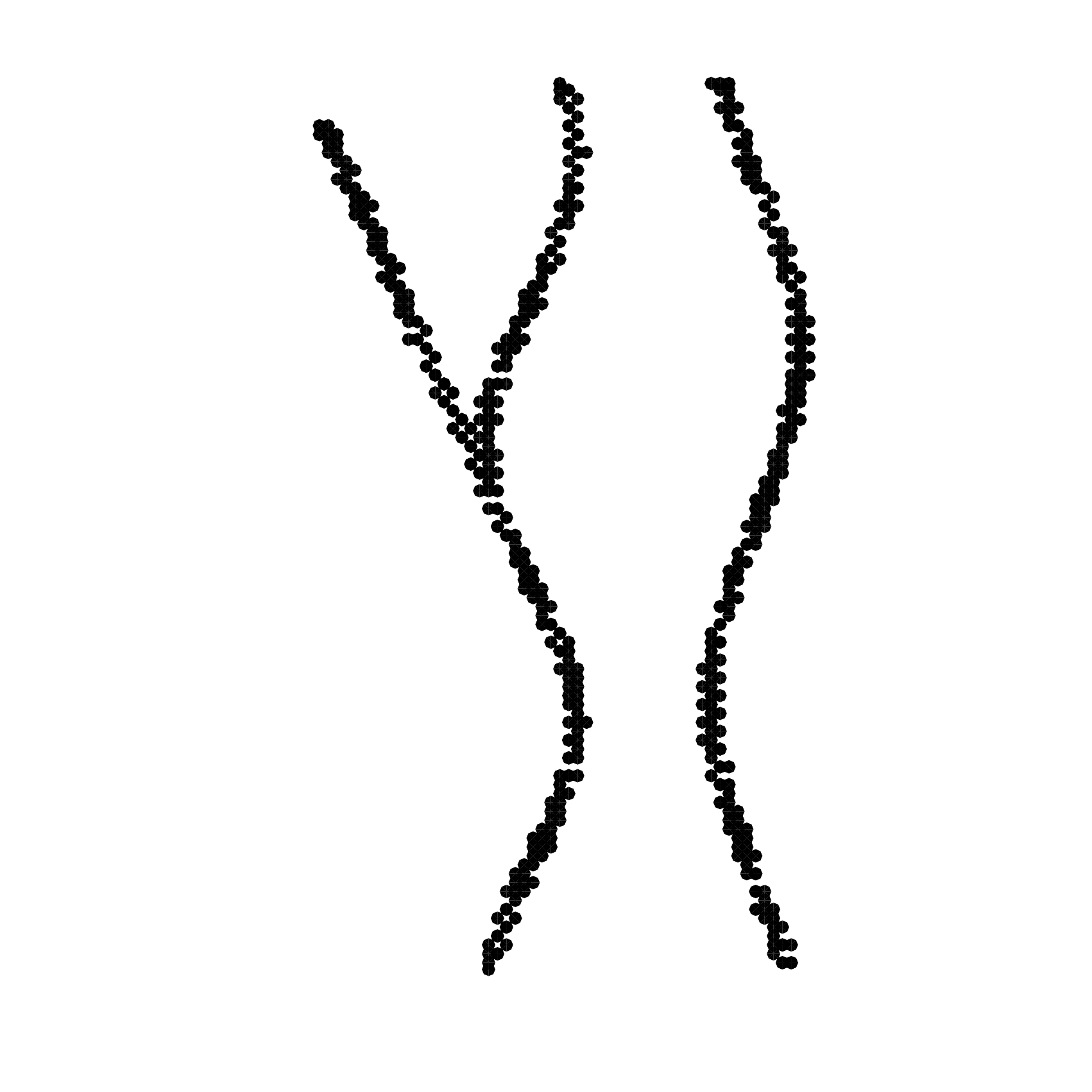} \qquad
 		\includegraphics[width=\sizp,height=\sizp]{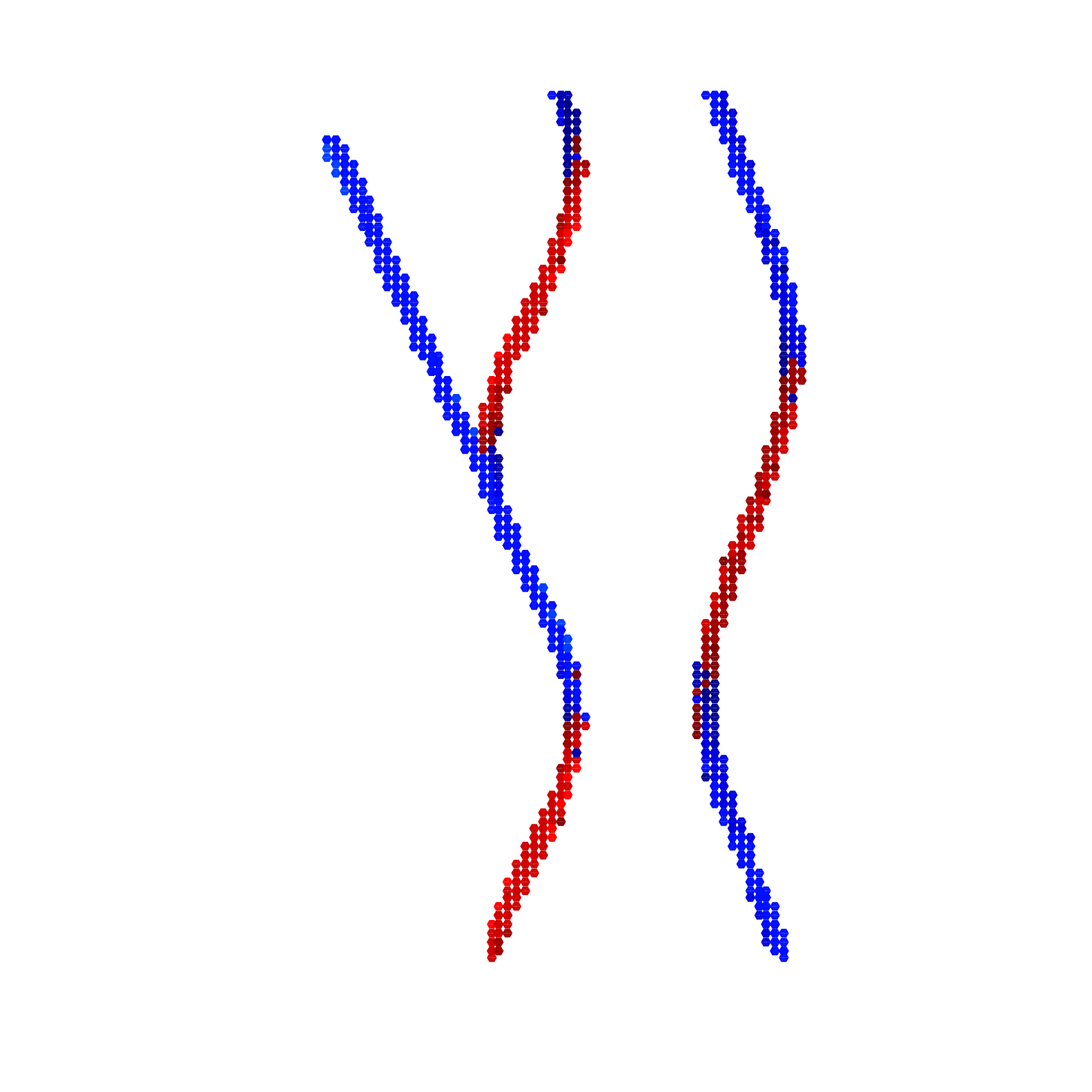} \qquad
 		\includegraphics[width=\sizp,height=\sizp]{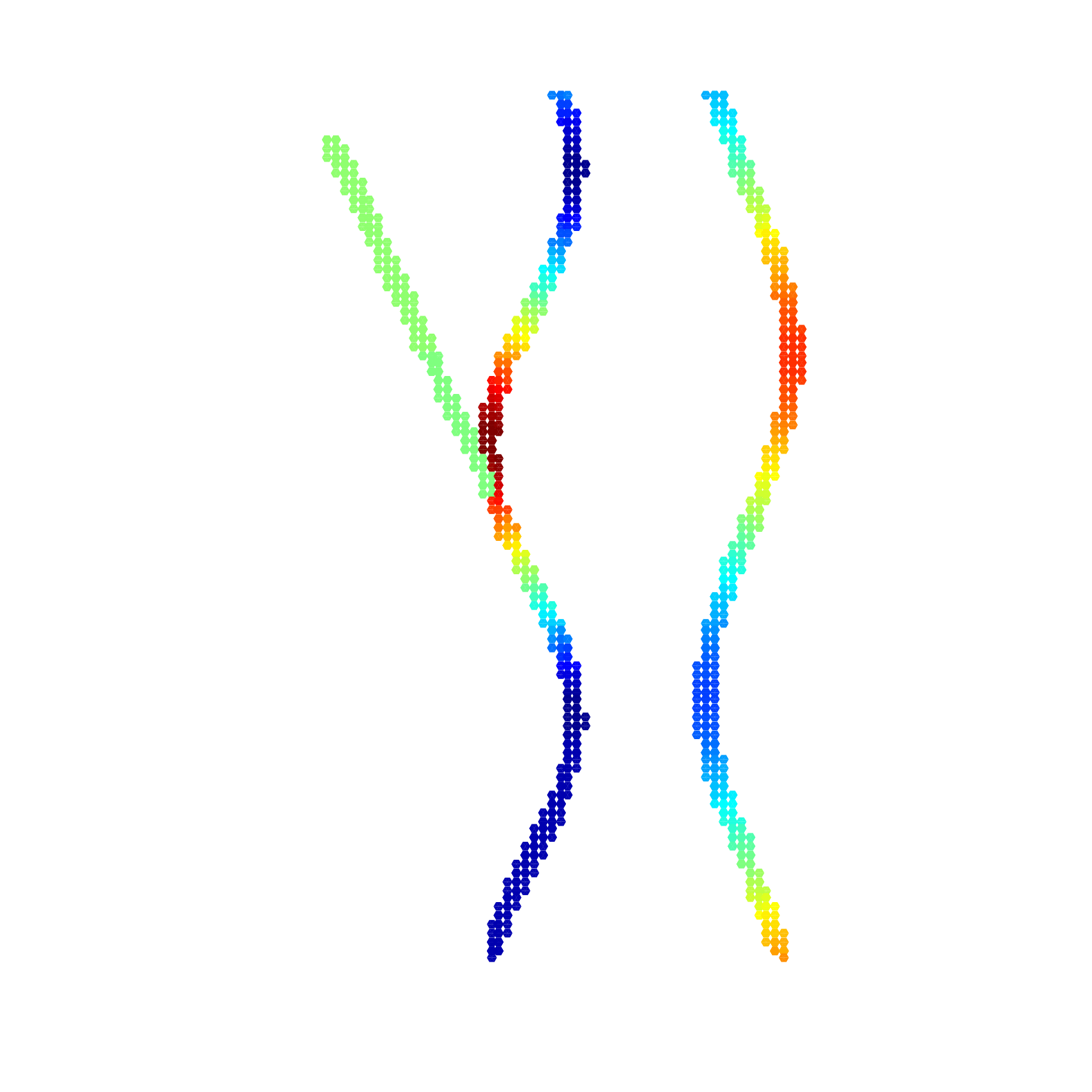}\qquad
 		\includegraphics[width=\sizp,height=\sizp]{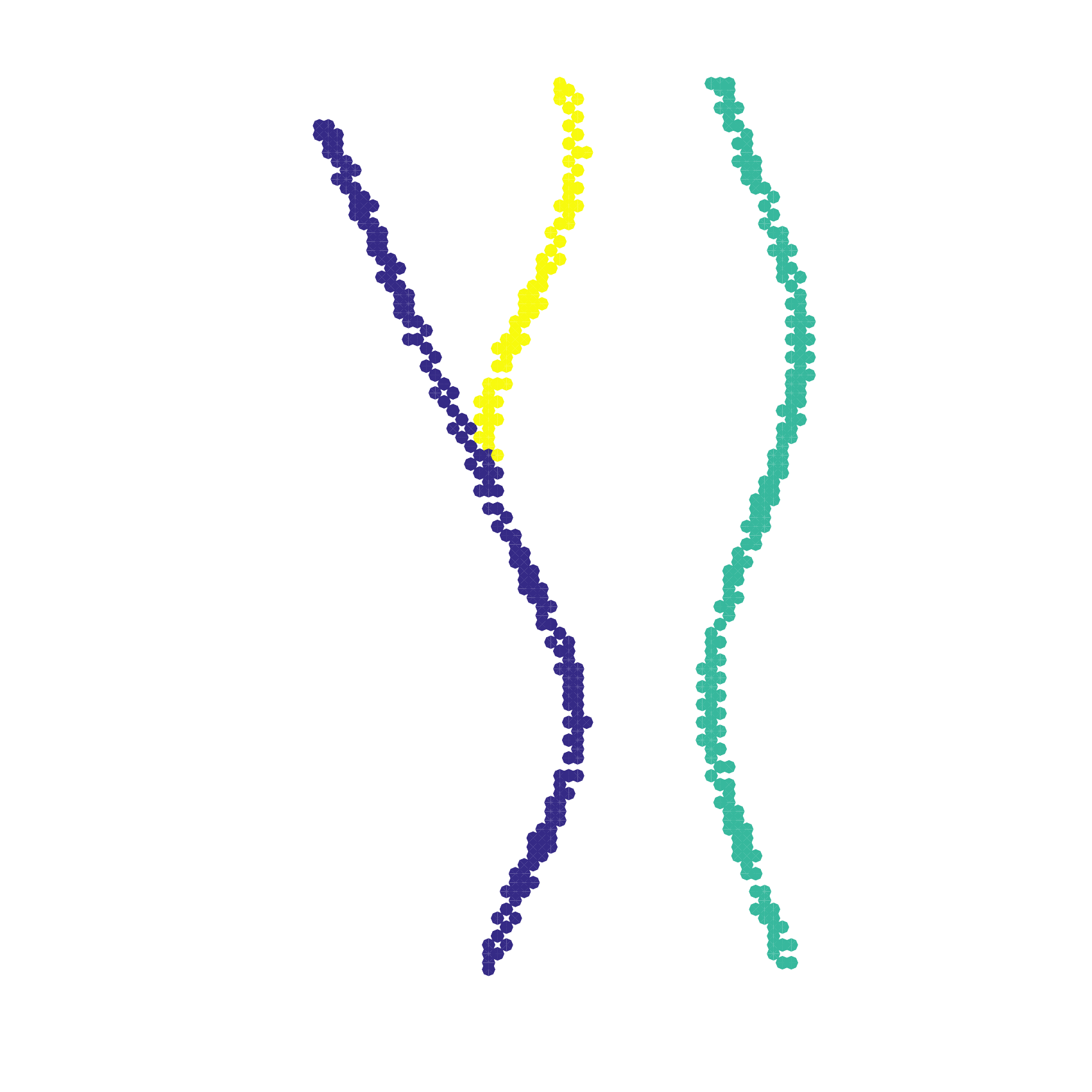} \qquad
 		\includegraphics[width=\sizp,height=\sizp]{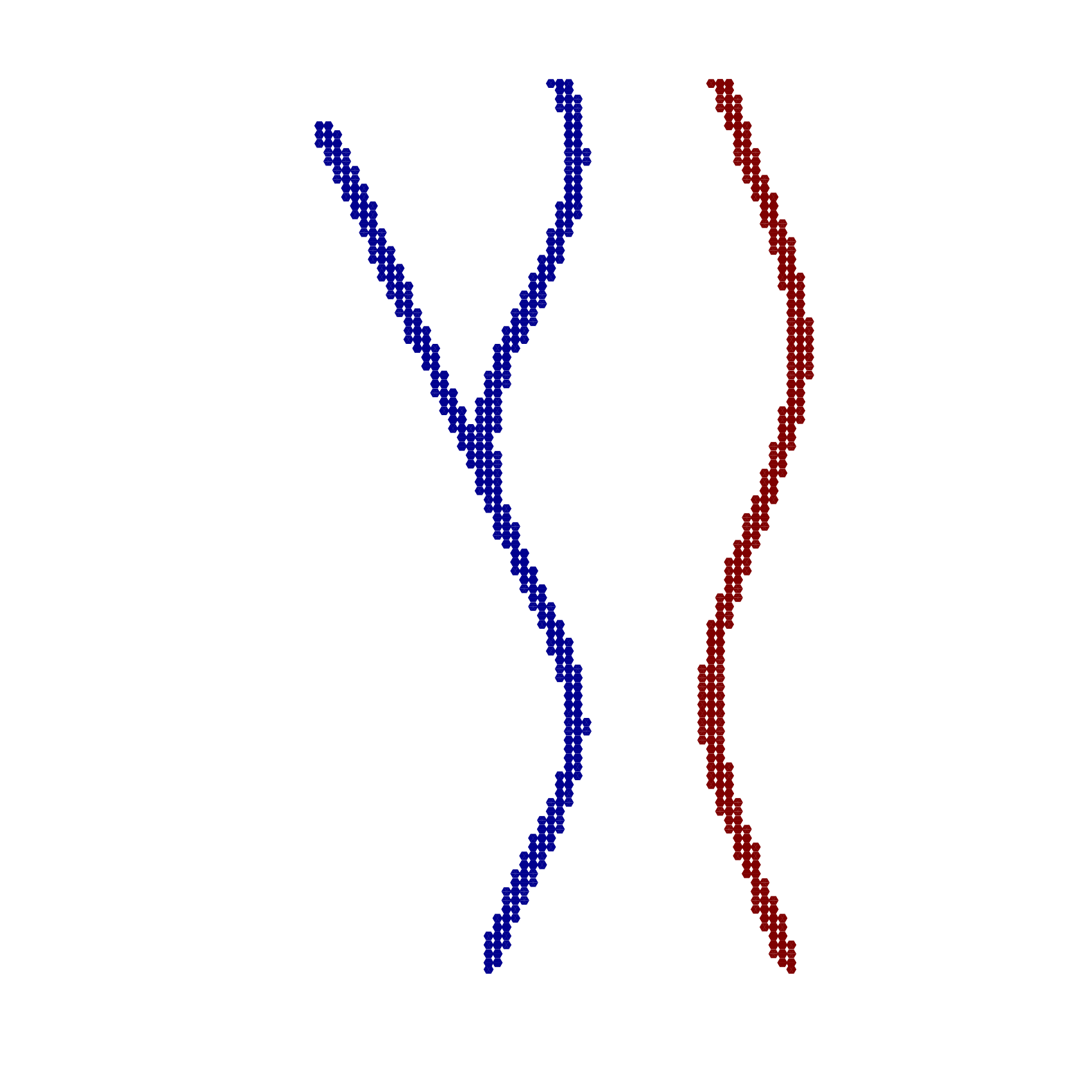} 
 	\end{subfigure}
 	
 	\begin{subfigure}[b]{6.8in}
 		\centering
 		\makebox[0pt][r]{\makebox[15pt]{\raisebox{25pt}{\rotatebox[origin=c]{0}{D}}}}  \qquad
 		\includegraphics[width=\sizp,height=\sizp]{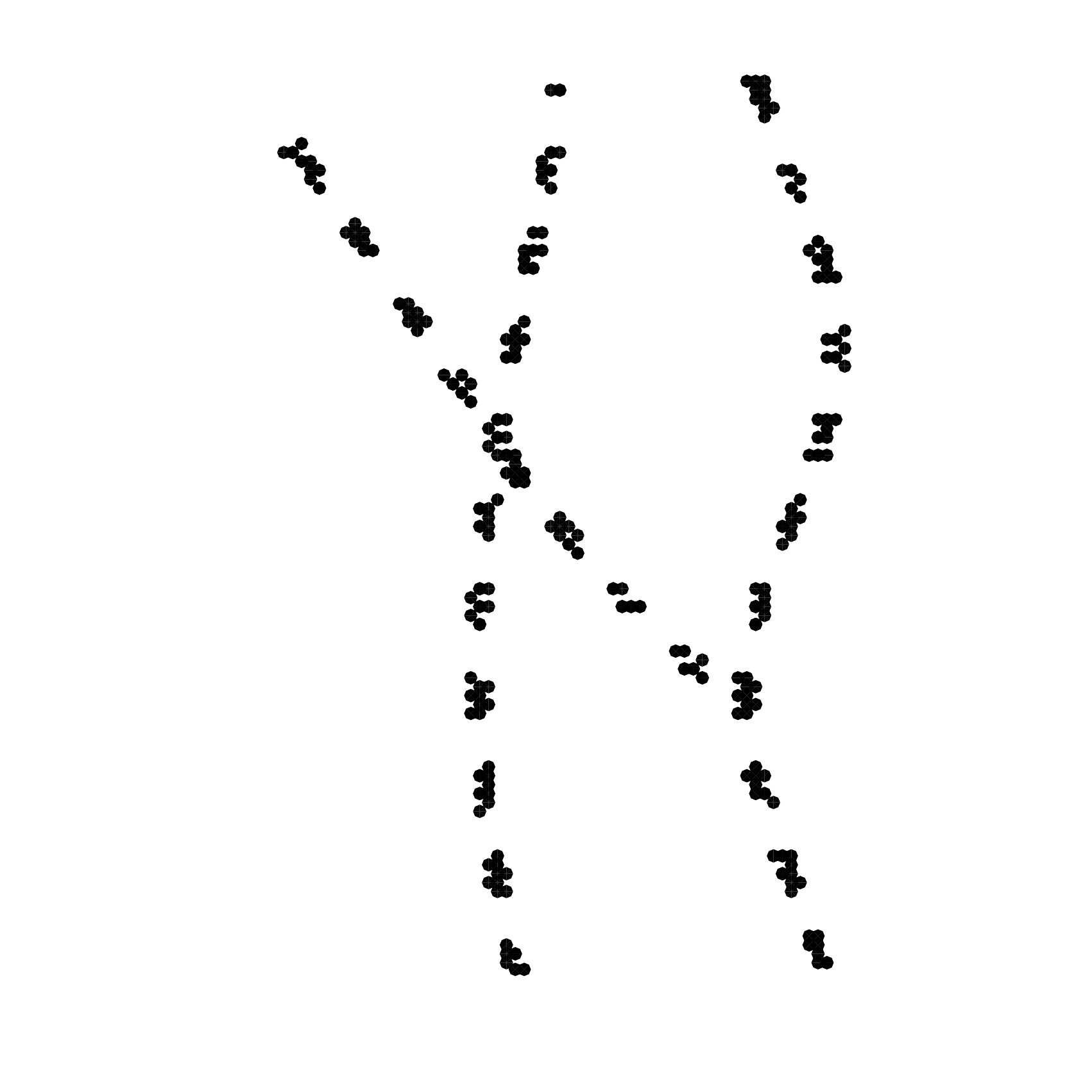} \qquad
 		\includegraphics[width=\sizp,height=\sizp]{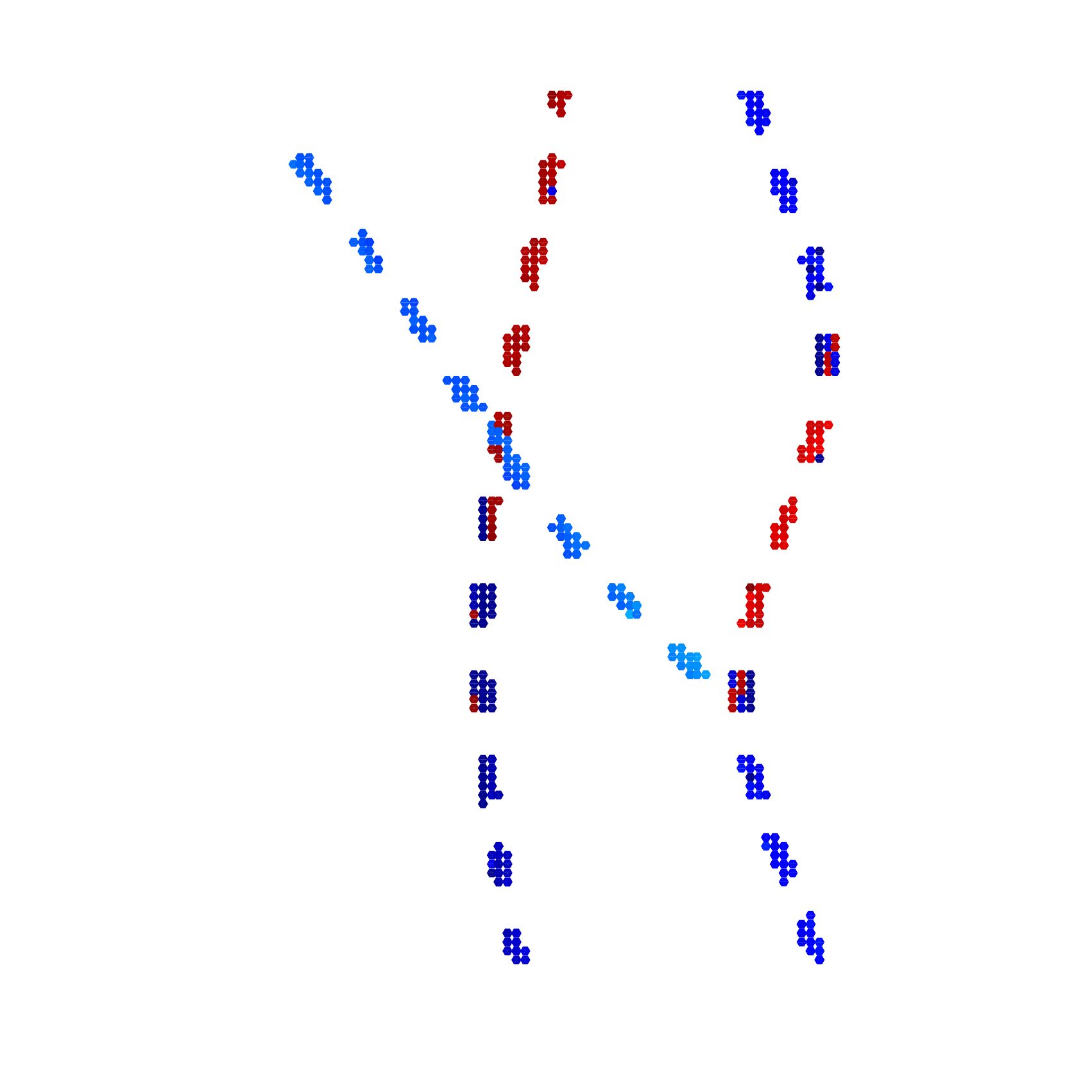} \qquad
 		\includegraphics[width=\sizp,height=\sizp]{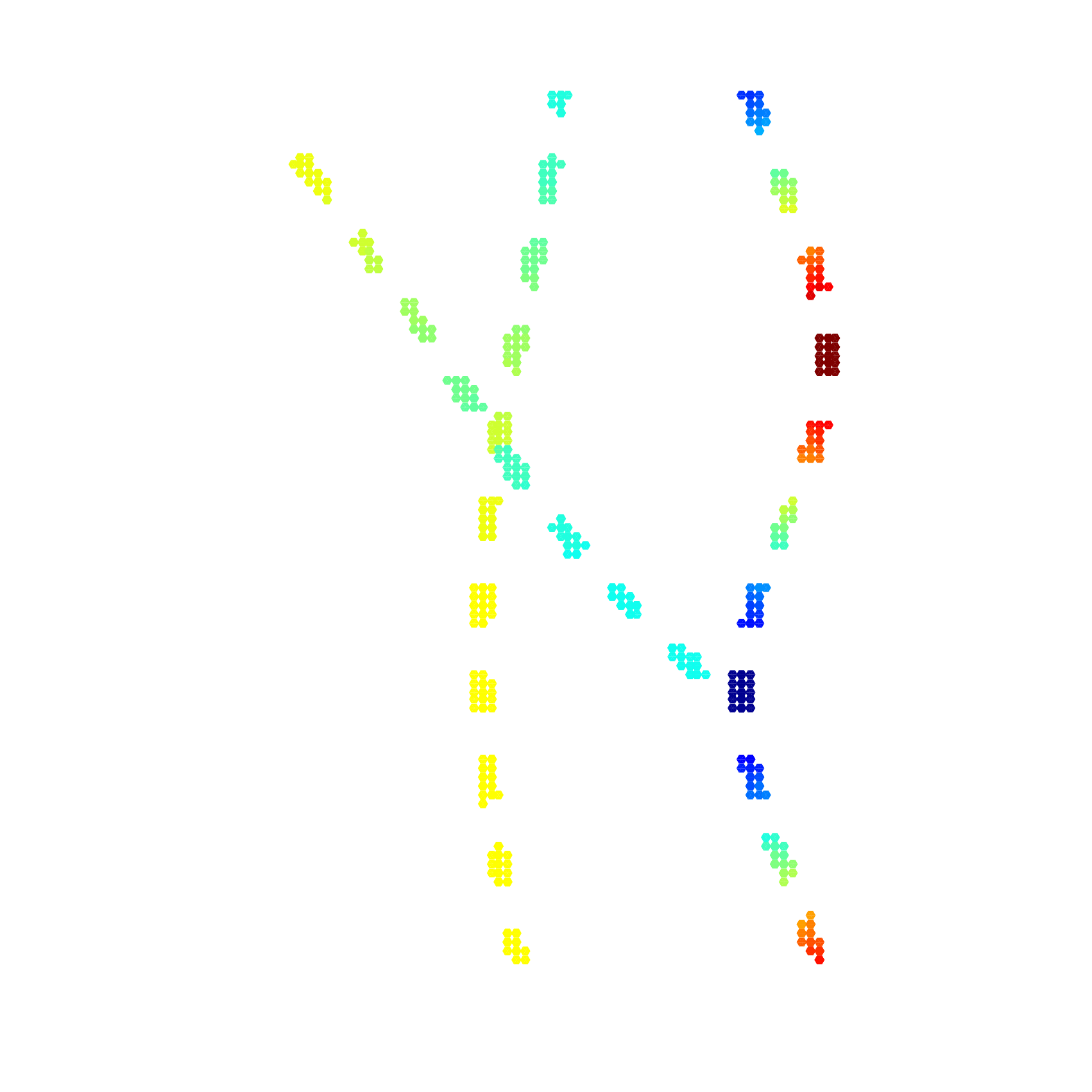}\qquad
 		\includegraphics[width=\sizp,height=\sizp]{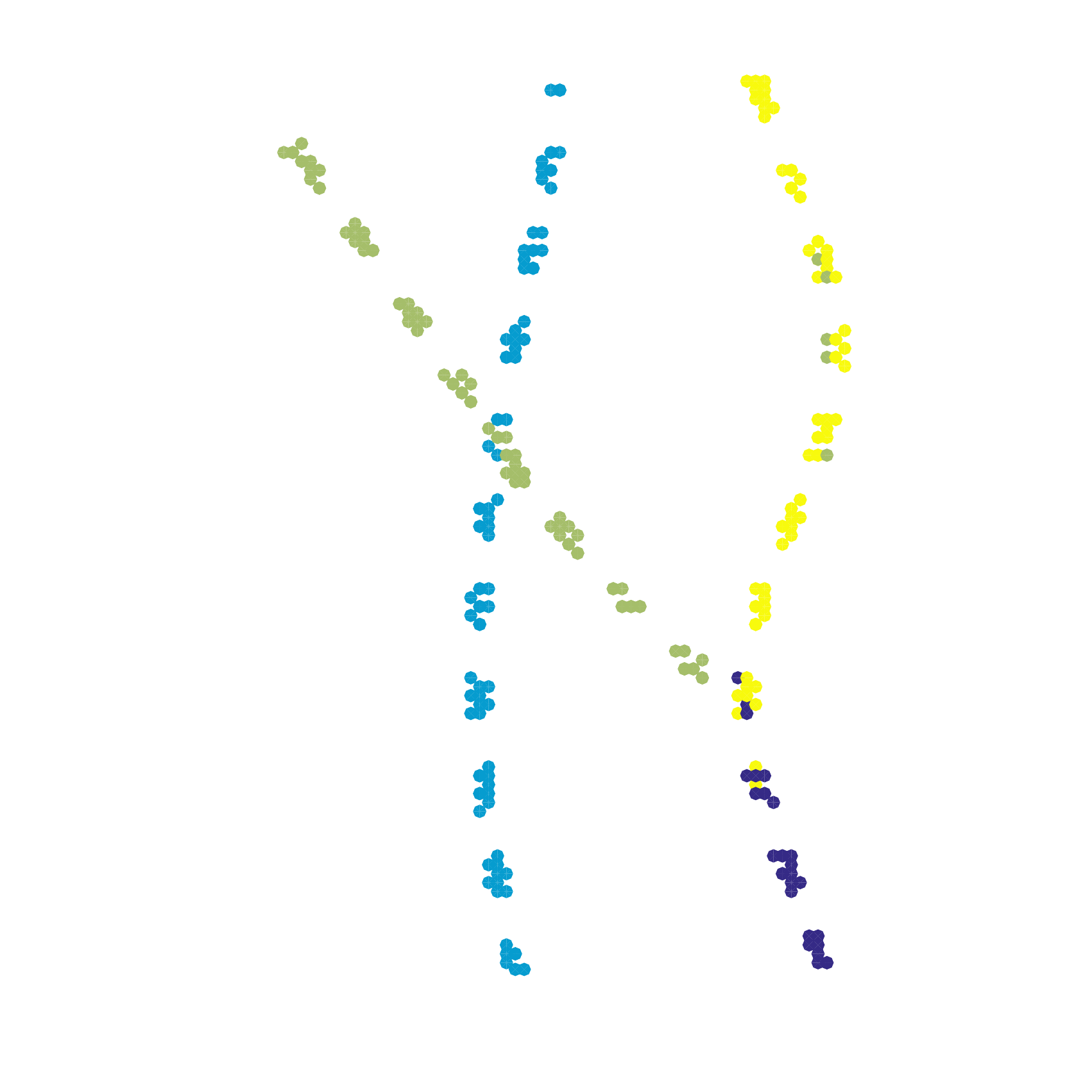} \qquad
 		\includegraphics[width=\sizp,height=\sizp]{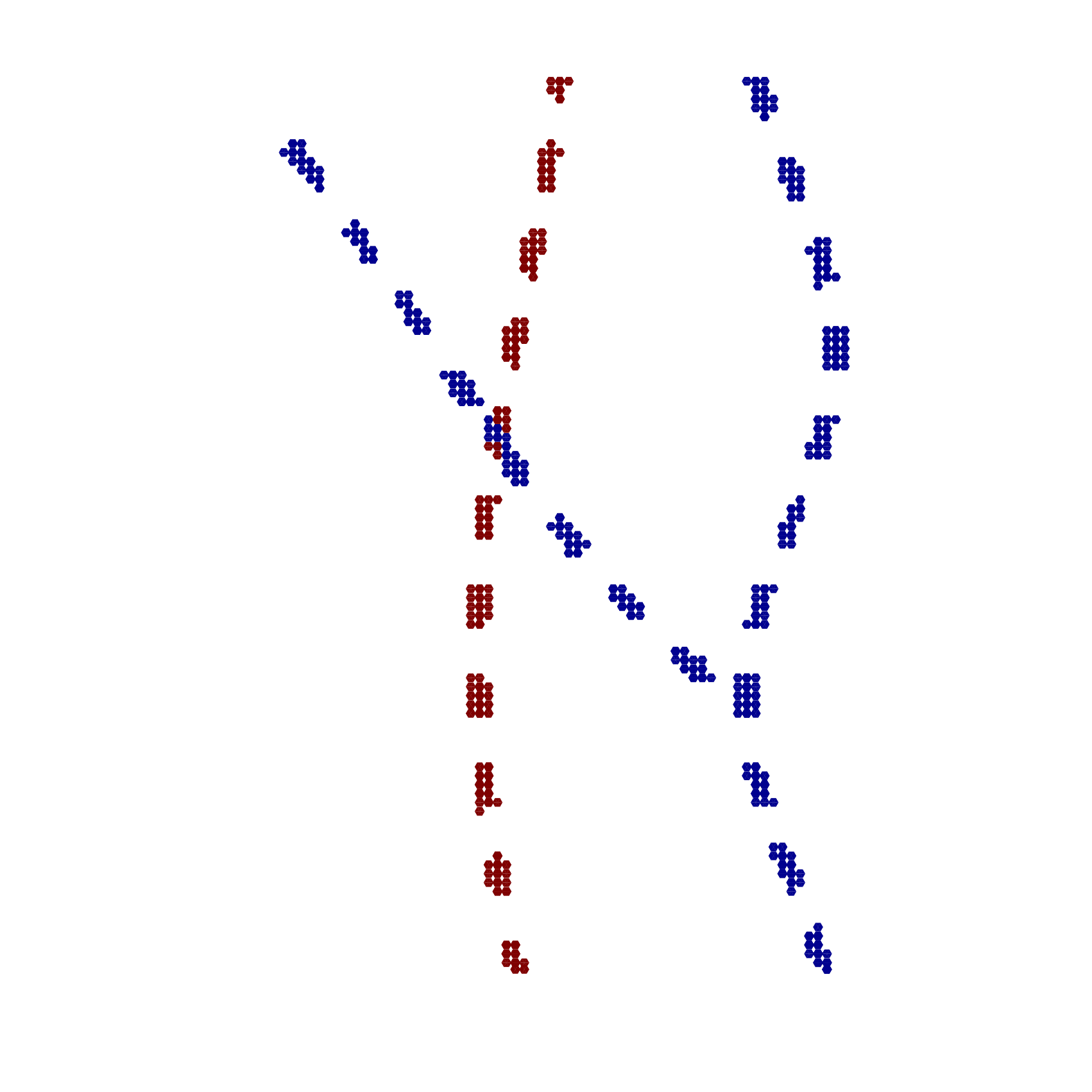} 
 	\end{subfigure}
 	
 	\begin{subfigure}[b]{6.8in}
 		\centering
 		\makebox[0pt][r]{\makebox[15pt]{\raisebox{25pt}{\rotatebox[origin=c]{0}{E}}}}  \qquad
 		\includegraphics[width=\sizp,height=\sizp]{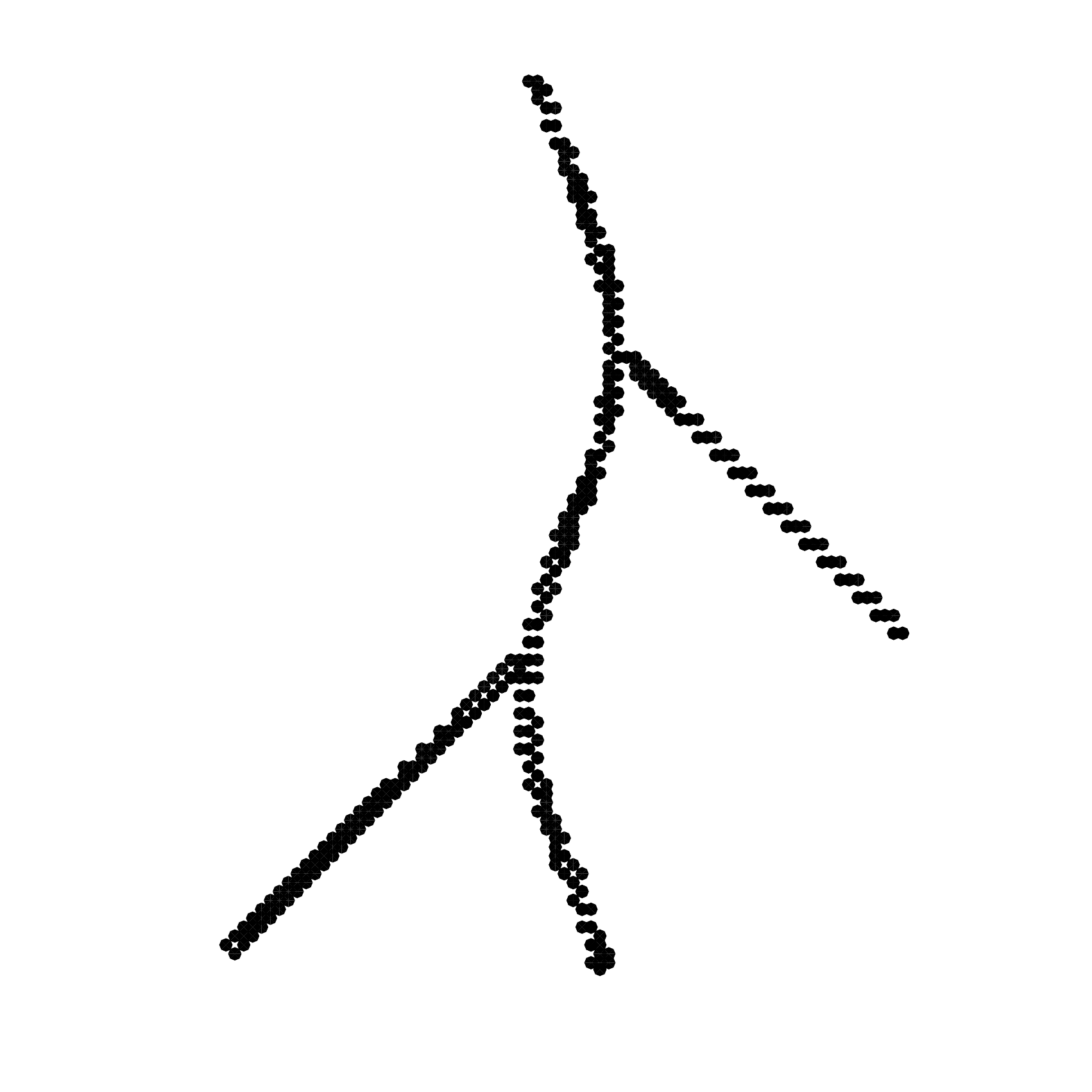} \qquad
 		\includegraphics[width=\sizp,height=\sizp]{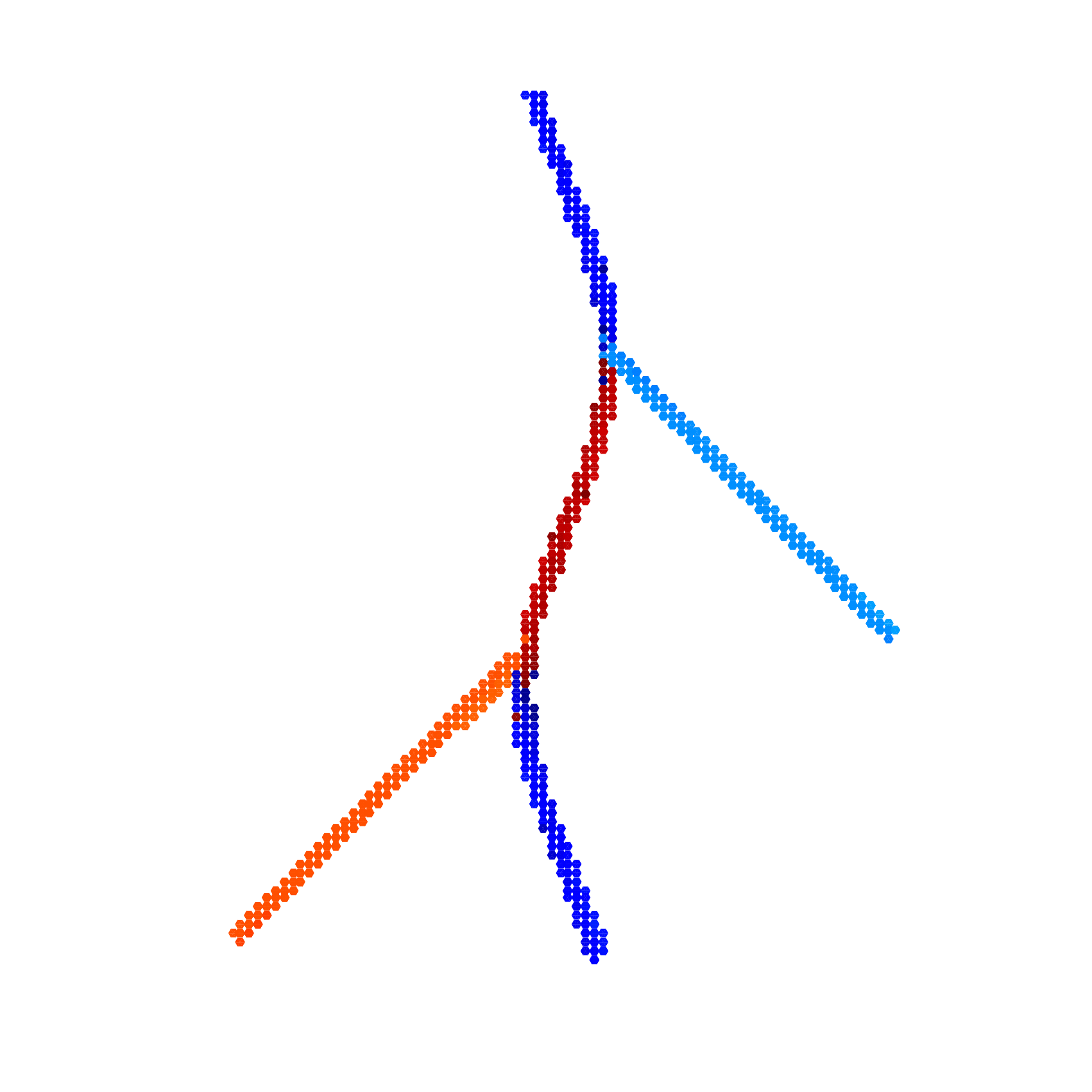} \qquad
 		\includegraphics[width=\sizp,height=\sizp]{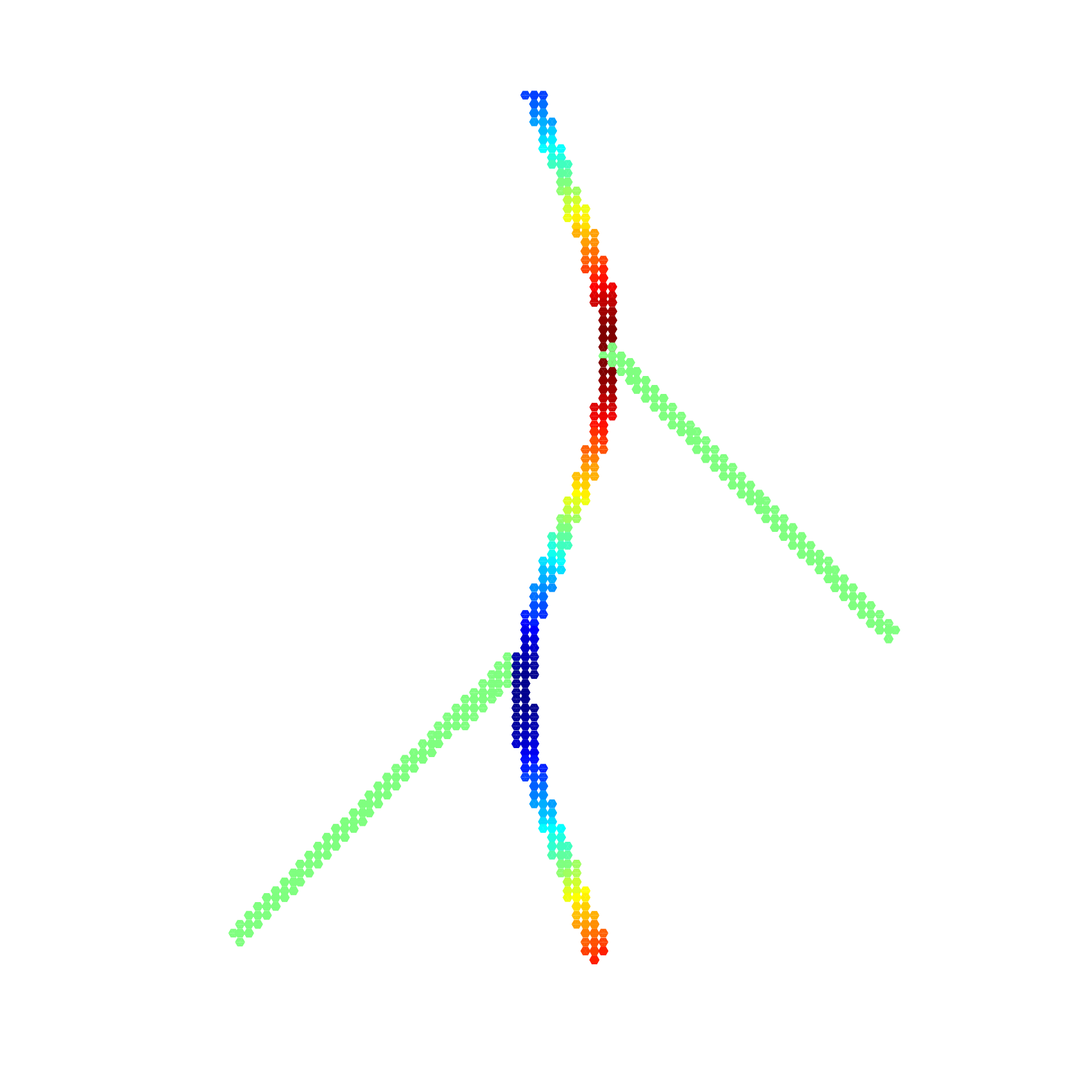}\qquad
 		\includegraphics[width=\sizp,height=\sizp]{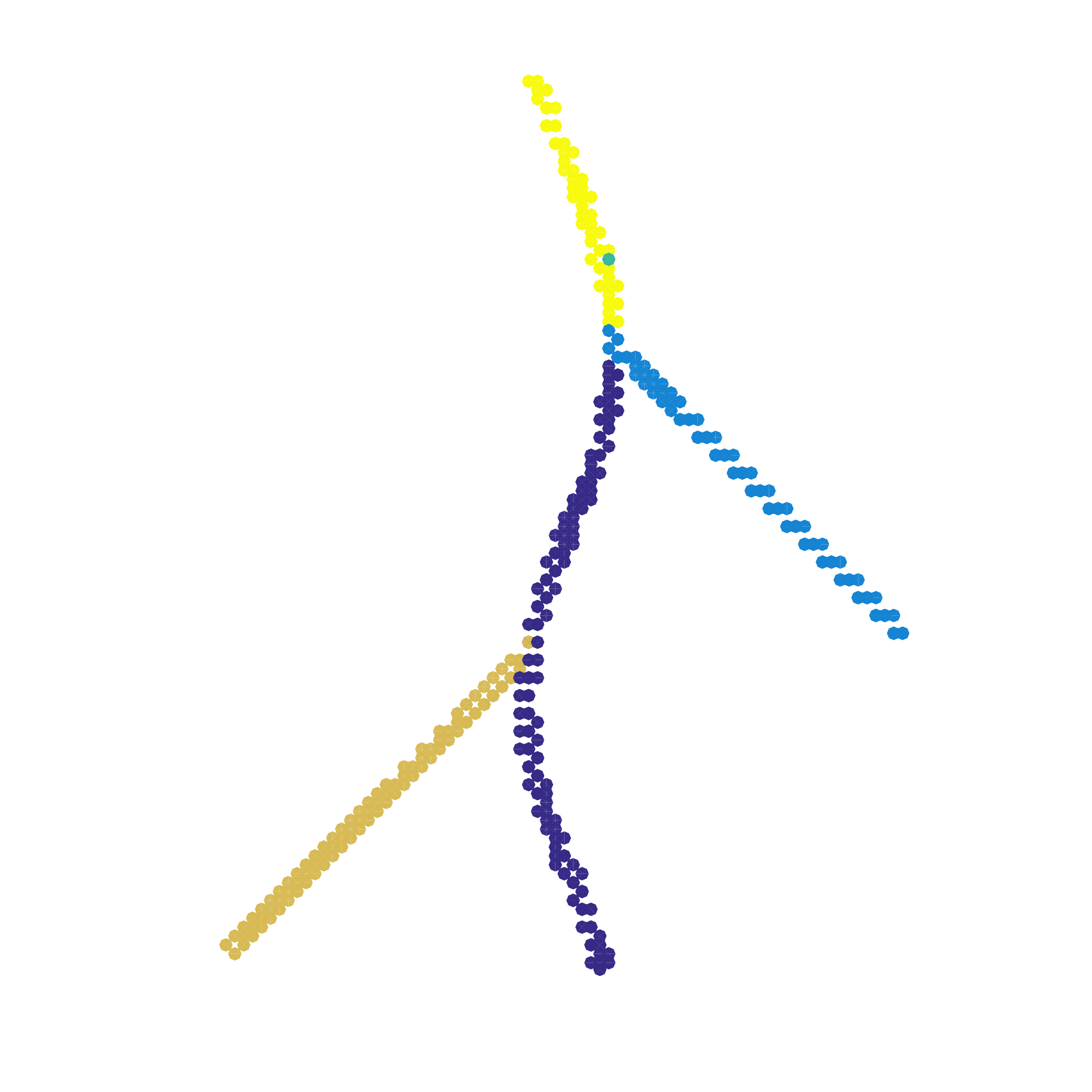} \qquad
 		\includegraphics[width=\sizp,height=\sizp]{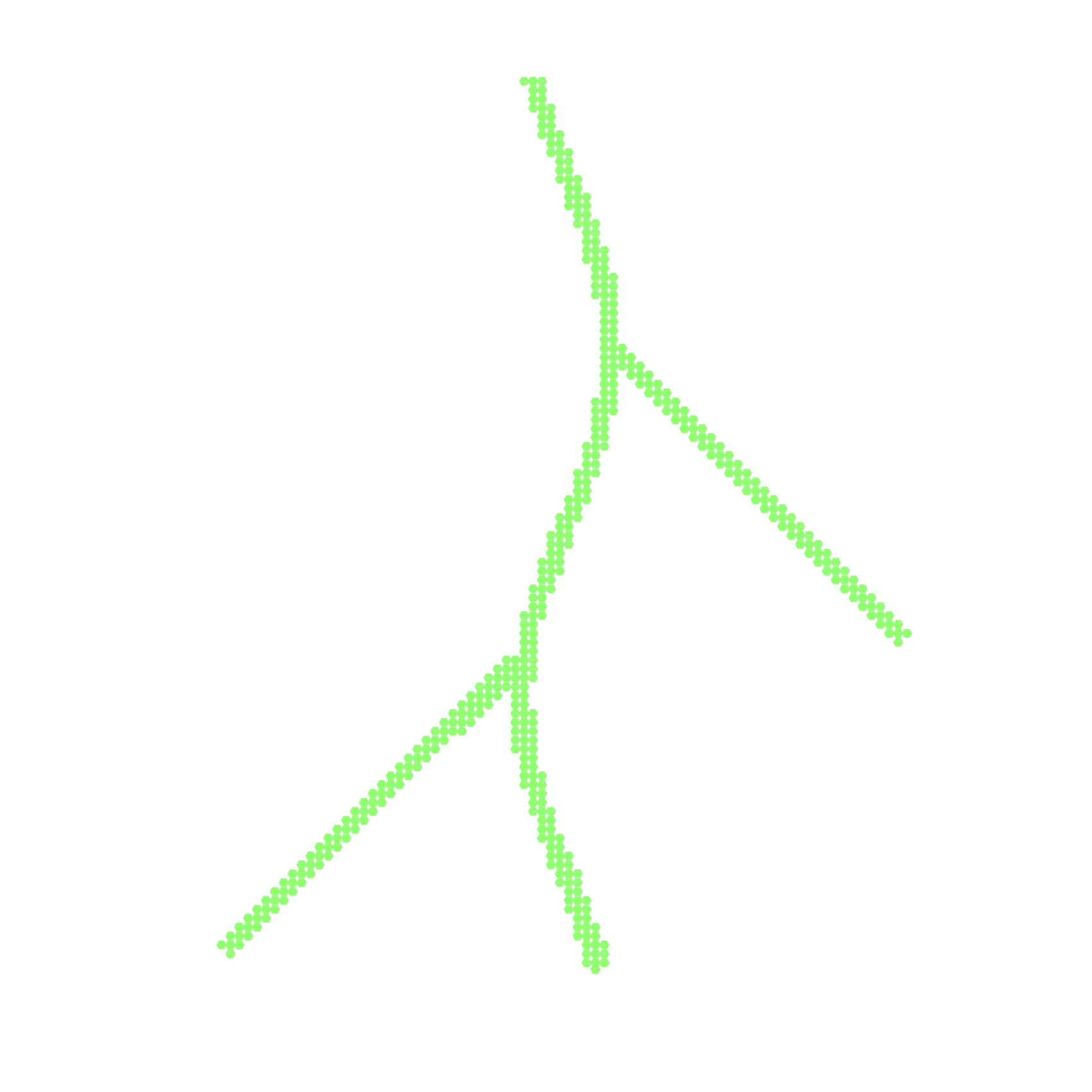} 
 	\end{subfigure}
 	
 	\caption{Samples of phantom images in different categories. From left to right, the images in each category represent:
 		stimulus, the orientation map, the curvature map and the clustering result with the previous [34] and the new kernel. The color of the curvature maps are scaled between the maximum and minimum values of the curvature in each image. }
 	
 \end{figure*}
 
 \newpage
 
 \begin{figure*}
 	\centering
 	
 	\begin{subfigure}[b]{6.8in}
 		\centering
 		\makebox[0pt][r]{\makebox[15pt]{\raisebox{25pt}{\rotatebox[origin=c]{0}{A}}}}  \qquad
 		\includegraphics[width=\siz,height=\siz]{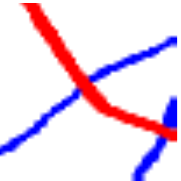} \qquad
 		\includegraphics[width=\siz,height=\siz]{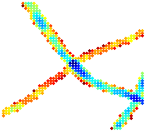}\qquad
 		\includegraphics[width=\siz,height=\siz]{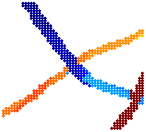} \qquad
 		\includegraphics[width=\siz,height=\siz]{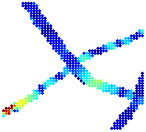} \qquad
 		\includegraphics[width=\siz,height=\siz]{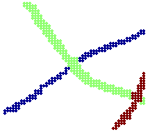} 
 	\end{subfigure}
 	
 	\begin{subfigure}[b]{6.8in}
 		\centering
 		\makebox[0pt][r]{\makebox[15pt]{\raisebox{25pt}{\rotatebox[origin=c]{0}{B}}}}  \qquad
 		\includegraphics[width=\siz,height=\siz]{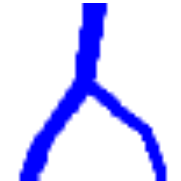} \qquad
 		\includegraphics[width=\siz,height=\siz]{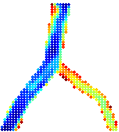}\qquad
 		\includegraphics[width=\siz,height=\siz]{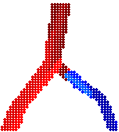} \qquad
 		\includegraphics[width=\siz,height=\siz]{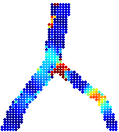} \qquad
 		\includegraphics[width=\siz,height=\siz]{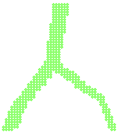} 
 	\end{subfigure}
 	
 	\begin{subfigure}[b]{6.8in}
 		\centering
 		\makebox[0pt][r]{\makebox[15pt]{\raisebox{25pt}{\rotatebox[origin=c]{0}{C}}}}  \qquad
 		\includegraphics[width=\siz,height=\siz]{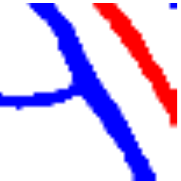} \qquad
 		\includegraphics[width=\siz,height=\siz]{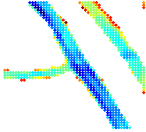}\qquad
 		\includegraphics[width=\siz,height=\siz]{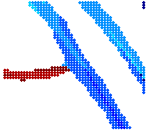} \qquad
 		\includegraphics[width=\siz,height=\siz]{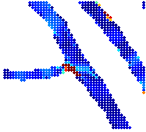} \qquad
 		\includegraphics[width=\siz,height=\siz]{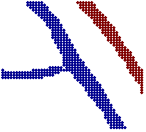} 
 	\end{subfigure}
 	
 	\begin{subfigure}[b]{6.8in}
 		\centering
 		\makebox[0pt][r]{\makebox[15pt]{\raisebox{25pt}{\rotatebox[origin=c]{0}{D}}}}  \qquad
 		\includegraphics[width=\siz,height=\siz]{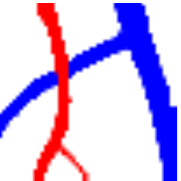} \qquad
 		\includegraphics[width=\siz,height=\siz]{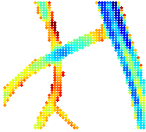} \qquad
 		\includegraphics[width=\siz,height=\siz]{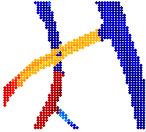}  \qquad
 		\includegraphics[width=\siz,height=\siz]{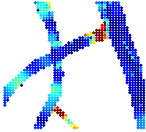} \qquad
 		\includegraphics[width=\siz,height=\siz]{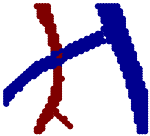} 
 	\end{subfigure}
 	
 	\begin{subfigure}[b]{6.8in}
 		\centering
 		\makebox[0pt][r]{\makebox[15pt]{\raisebox{25pt}{\rotatebox[origin=c]{0}{E}}}}   \qquad
 		\includegraphics[width=\siz,height=\siz]{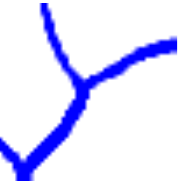} \qquad
 		\includegraphics[width=\siz,height=\siz]{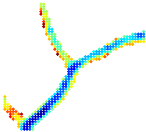} \qquad
 		\includegraphics[width=\siz,height=\siz]{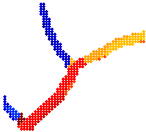}  \qquad
 		\includegraphics[width=\siz,height=\siz]{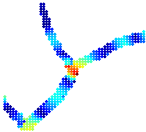}  \qquad
 		\includegraphics[width=\siz,height=\siz]{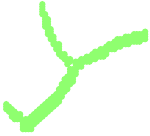} 
 	\end{subfigure}
 	
 	\caption{Samples of retinal patches in different categories. From left to right, the images in each category represent: artery/vein vessel ground truth, intensity, orientation, curvature and clustering results. The color of the curvature maps are scaled between the maximum and minimum values of the curvature in each image patch.}
 	\label{fig:resRetPatch}
 \end{figure*}

